%% file: main.tex
\documentclass[twoside,11pt,table]{article}

\usepackage{jmlr2e}
\usepackage{etoolbox,refcount}
\usepackage{multicol}
\usepackage{colortbl}
\usepackage{pgfplots}
\usepackage{pgfplotstable}
\usepackage{booktabs}
\usepackage{graphicx}
\usepackage{subcaption}
\usepackage[export]{adjustbox}
\usepgfplotslibrary{groupplots}
\pgfplotsset{compat=newest}
\usepackage{multirow}
\usepackage{rotating}
\usepackage{float}
\usepackage{amssymb,amsmath}

\pgfplotstableset{
    /color cells/min/.initial=0,
    /color cells/max/.initial=1000,
    /color cells/textcolor/.initial=,
    color cells/.code={%
        \pgfqkeys{/color cells}{#1}%
        \pgfkeysalso{%
            postproc cell content/.code={%
                \begingroup
                \pgfkeysgetvalue{/pgfplots/table/@preprocessed cell content}\value
\ifx\value\empty
\endgroup
\else
                \pgfmathfloatparsenumber{\value}%
                \pgfmathfloattofixed{\pgfmathresult}%
                \let\value=\pgfmathresult
                \pgfplotscolormapaccess
                    [\pgfkeysvalueof{/color cells/min}:\pgfkeysvalueof{/color cells/max}]%
                    {\value}%
                    {\pgfkeysvalueof{/pgfplots/colormap name}}%
                \pgfkeysgetvalue{/pgfplots/table/@cell content}\typesetvalue
                \pgfkeysgetvalue{/color cells/textcolor}\textcolorvalue
                \toks0=\expandafter{\typesetvalue}%
                \xdef\temp{%
                    \noexpand\pgfkeysalso{%
                        @cell content={%
                            \noexpand\cellcolor[rgb]{\pgfmathresult}%
                            \noexpand\definecolor{mapped color}{rgb}{\pgfmathresult}%
                            \ifx\textcolorvalue\empty
                            \else
                                \noexpand\color{\textcolorvalue}%
                            \fi
                            \the\toks0 %
                        }%
                    }%
                }%
                \endgroup
                \temp
\fi
            }%
        }%
    }
}

\pgfplotsset{
  log x ticks with fixed point/.style={
      xticklabel={
        \pgfkeys{/pgf/fpu=true}
        \pgfmathparse{exp(\tick)}%
        \pgfmathprintnumber[fixed relative, precision=3]{\pgfmathresult}
        \pgfkeys{/pgf/fpu=false}
      }
  },
  log y ticks with fixed point/.style={
      yticklabel={
        \pgfkeys{/pgf/fpu=true}
        \pgfmathparse{exp(\tick)}%
        \pgfmathprintnumber[fixed relative, precision=3]{\pgfmathresult}
        \pgfkeys{/pgf/fpu=false}
      }
  }
}

\newcounter{countitems}
\newcounter{nextitemizecount}
\newcommand{\setupcountitems}{%
  \stepcounter{nextitemizecount}%
  \setcounter{countitems}{0}%
  \preto\item{\stepcounter{countitems}}%
}
\makeatletter
\newcommand{\computecountitems}{%
  \edef\@currentlabel{\number\c@countitems}%
  \label{countitems@\number\numexpr\value{nextitemizecount}-1\relax}%
}
\newcommand{\nextitemizecount}{%
  \getrefnumber{countitems@\number\c@nextitemizecount}%
}
\newcommand{\previtemizecount}{%
  \getrefnumber{countitems@\number\numexpr\value{nextitemizecount}-1\relax}%
}
\makeatother    
\newenvironment{AutoMultiColItemize}{%
\ifnumcomp{\nextitemizecount}{>}{3}{\begin{multicols}{2}}{}%
\setupcountitems\begin{itemize}}%
{\end{itemize}%
\unskip\computecountitems\ifnumcomp{\previtemizecount}{>}{3}{\end{multicols}}{}}

\newcommand{\eg}{\textit{e.g. }}
\newcommand{\ie}{\textit{i.e. }}

\jmlrheading{1}{2000}{1-48}{4/00}{10/00}{Grobelnik and Vanschoren}

\ShortHeadings{Warm-starting DARTS using meta-learning}{Grobelnik and Vanschoren}
\firstpageno{1}

\begin{document}

\title{Warm-starting DARTS using meta-learning}

\author{\name Matej Grobelnik \email matej.grobelnik@student.tue.nl \\
\name Joaquin Vanschoren \email j.vanschoren@tue.nl}

\editor{TBD}

\maketitle

\begin{abstract}%   <- trailing '%' for backward compatibility of .sty file
Neural architecture search (NAS) has shown great promise in the field of automated machine learning (AutoML). NAS has outperformed hand-designed networks and made a significant step forward in the field of automating the design of deep neural networks, thus further reducing the need for human expertise. However, most research is done targeting a single specific task, leaving research of NAS methods over multiple tasks mostly overlooked. Generally, there exist two popular ways to find an architecture for some novel task. Either searching from scratch, which is ineffective by design, or transferring discovered architectures from other tasks, which provides no performance guarantees and is probably not optimal. In this work we present a meta-learning framework to warm-start Differentiable architecture search (DARTS). DARTS is a NAS method that can be initialized with a transferred architecture and is able to quickly adapt to new tasks. A task similarity measure is used to determine which transfer architecture is selected, as transfer architectures found on similar tasks will likely perform better. Additionally, we employ a simple meta-transfer architecture that was learned over multiple tasks. Experiments show that warm-started DARTS is able to find competitive performing architectures while reducing searching cost on average by 60\%.  
\end{abstract}


\section{Introduction}
\input{chapters/00_introduction}

\section{Preliminaries}
\input{chapters/01a_preliminaries}

\section{Related Work}
\input{chapters/01b_related_work}

\section{Methodology}
\input{chapters/02_methodology}

\section{Experiment Design}
\input{chapters/03_experiments}

\section{Results}
\input{chapters/04_results}

\section{Conclusions and Future Work}
\input{chapters/05_conclusions}

% Acknowledgements should go at the end, before appendices and references


\newpage
\bibliography{main}

%\newpage
%\appendix
%\input{chapters/06_appendix}

\end{document}

%% file: chapters/00_introduction.tex
The standard approach of manually creating Artificial Neural Networks (ANNs) for a novel dataset requires substantial resources in terms of human expertise, computational power, and time. To make ANNs more generally accessible, there has been growing interest in automating or partly automating this process. 

Neural architecture search (NAS) plays a central role in the field of automated machine learning and aims to automate the design of neural networks. NAS is not a recent invention but has only recently got increased attention from the research community. This can be attributed to advances in the research that managed to automatically create neural networks that could compete or even outperform other state-of-the-art networks \citep{zoph2017neural,baker2017designing, real2019regularized, elsken2019efficient}. The drawback of many of these methods is their huge computational cost since NAS algorithms are computationally demanding despite their performance. Running NAS for the popular dataset CIFAR-10 \citep{Krizhevsky09learningmultiple} required 2000 GPU days with reinforcement learning \citep{zoph2017learning} and 3150 GPU days with evolutionary algorithms \citep{real2019regularized}. The main problem is that directly searching the best architecture within a discrete space is inefficient given the size of the search space. One-shot methods \citep{liu2019darts, brock2017smash, cai2019proxylessnas} were proposed to reduce search cost as they can combine architectures from the search space by either sharing weights or by continuous relaxation.  

Differentiable architecture search (DARTS) \citep{liu2019darts} is a one-shot method that relaxes the search space to be continuous so that the architecture can be optimized with respect to its validation set performance by gradient descent. The algorithm first starts with a single large hyper-network representing the complete search space, and through gradient descent, it discovers the optimal subgraph. DARTS made an enormous step forward by reducing the required amount of time to create a neural network to just 4 GPU days \citep{liu2019darts}. Nevertheless, DARTS can still be expensive given the large search space. Therefore, there is a need to study how can we use prior knowledge obtained on similar tasks to adapt to some new unseen tasks easily and quickly while reducing the computational complexity of NAS in terms of GPU time, without compromising the architecture accuracy. 

Any type of learning based on experience with prior tasks is known as \textit{meta-learning} \citep{vanschoren2018mlsurvey}. The experience itself is gained by exploiting meta-knowledge extracted in previous learning episodes on single or multiple tasks. However, to successfully transfer prior knowledge from source tasks to a target task there must be some similarity, since unrelated tasks could negatively affect the performance. A task similarity measure is needed to find suitable tasks to warm-start neural architecture search.

Motivated by this intuition, we propose a novel approach to "\textit{warm-start}" DARTS that efficiently uses prior knowledge to find new architectures for previously unseen tasks. In this work, we will present a meta-learning framework \footnote{https://github.com/mgrobelnik/ws-darts} where we introduce architecture transfer to warm-start DARTS. 

%% file: chapters/01a_preliminaries.tex
In this section, we introduce the basic concepts and building blocks of our meta-learning framework. In section \ref{nas}, we present a general overview of neural architecture search (NAS). In sections \ref{darts} and \ref{p-darts}, we dive more deeply into two gradient-based NAS methods: DARTS and P-DARTS. Finally, in section \ref{meta-learning}, we present the concept of meta-learning with an emphasis on the task similarity measure where we present our chosen method: Task2Vec.

\subsection{Neural Architecture Search} \label{nas}

%%%%%%%%%%%%%%%%%%%%%%%%%%%%%%%%%%%%%%
%   NEURAL ARCHITECTURE SEARCH
%%%%%%%%%%%%%%%%%%%%%%%%%%%%%%%%%%%%%%

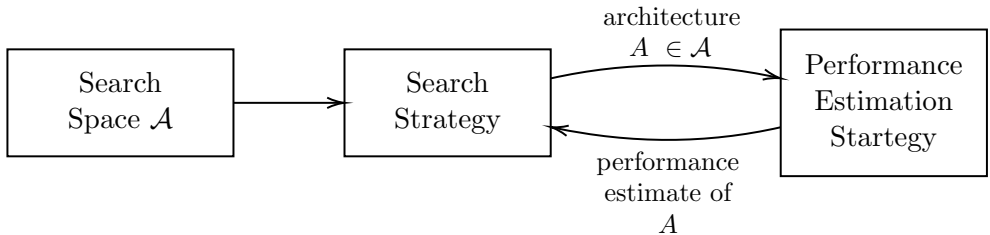
\begin{figure}[ht]
    \centering
    \input{fig/nas}
    \caption{Abstract illustration of Neural Architecture Search methods \citep{elsken2019survey}. A search strategy selects an architecture $A$ from a predefined search space $\mathcal{A}$. The architecture is passed to a performance estimation strategy, \textit{e.g.}, evaluation on a hold-out test set, which returns the estimated performance of $A$ to the search strategy.}
    \label{fig:nas}
\end{figure}

Neural architecture search (NAS) is a technique for automating the design of artificial neural networks and can be seen as a sub-field of AutoML. \cite{elsken2019survey} summarized NAS as consisting of three major components: search space, search strategy, and performance estimation strategy. To find a suitable architecture, a NAS algorithm is run over a predefined search space, following a search strategy that will maximize the performance. The Figure \ref{fig:nas} illustrates the interaction between the components of an abstract NAS method. 

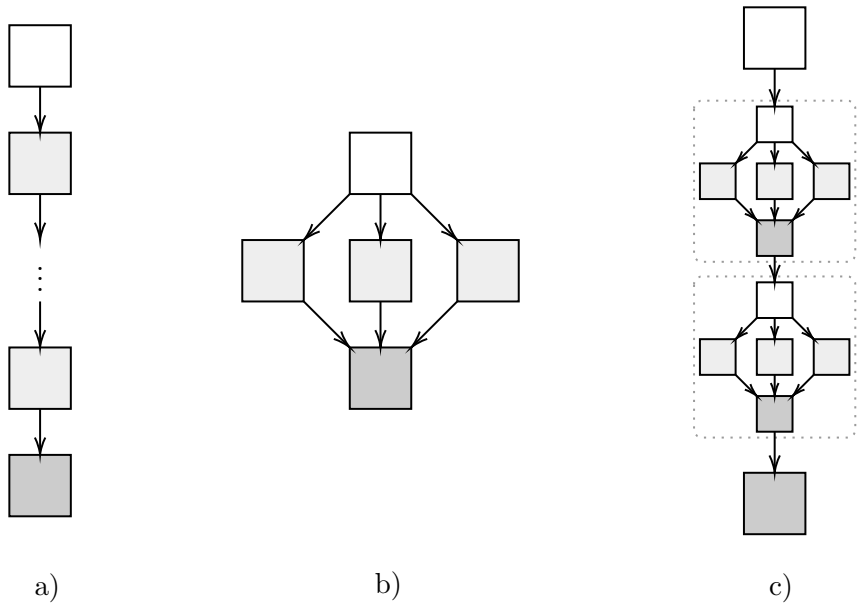
\begin{figure}[h]
    \centering
    \input{fig/search_space}
    \caption{An illustration of different architecture search spaces, where each node represents a layer in a neural network: a) \textit{chain-structured} space, b) \textit{multi-branch} space, and c) \textit{cell} space where cells are repeatedly stacked together to build more complex architectures. }
    \label{fig:search-space}
\end{figure}

The search space defines all architectures that can be represented by our NAS method. It can discover relatively simple \textit{chain-structured} neural networks (Figure \ref{fig:search-space}a) or more complex \textit{multi-branch} networks (Figure \ref{fig:search-space}b). It is also possible to fix the outer structure and let NAS search only for \textit{cell} architectures (Figure \ref{fig:search-space}c) also known as \textit{micro-search} or \textit{cell search}. The size of our search space determines how complex our search will be and can be reduced by defining constraining attributes from prior knowledge of well-suited architectures for the task at hand. However, this carries the risk of introducing human bias, which can prevent finding novel, better performing, architectures.

Next, the search strategy defines the algorithm on how the NAS system should explore the given search space. It plays a crucial role in finding well-performing architectures quickly while avoidinFg getting stuck in a region of sub-optimal architectures. We can choose from a wide variety of strategies to explore the search space:  random search \citep{Bergstra2012random}, Bayesian optimization \citep{pmlr-v28-bergstra13, Golovin2017vizier}, evolutionary methods \citep{miller1989, real2017largescale, real2019regularized, elsken2019efficient}, reinforcement learning \citep{baker2017designing, zoph2017learning, zhong2017practical}, and gradient-based methods \citep{liu2019darts}.

Finally, the performance estimation strategy determines how performance should be estimated. The conventional way of training and validating the architecture on data is computationally expensive and thus limits the number of architectures that can be explored. To address this issue, numerous methods for speeding up performance estimation have been proposed \citep{zoph2017neural, elsken2019efficient, runge2019learning, pmlr-v80-bender18a}.

%%%%%%%%%%%%%%%%%%%%%%%%%%%%%%%%%%%%%%
%   DARTS
%%%%%%%%%%%%%%%%%%%%%%%%%%%%%%%%%%%%%%

\subsection{Differentiable architecture search}\label{darts}

Traditional NAS methods --- \ie{Bayesian optimization, evolutionary methods, or reinforcement learning} --- are currently inefficient as they regard neural architecture search as a black-box optimization problem in a discrete search strategy \citep{ren2021comprehensive}.

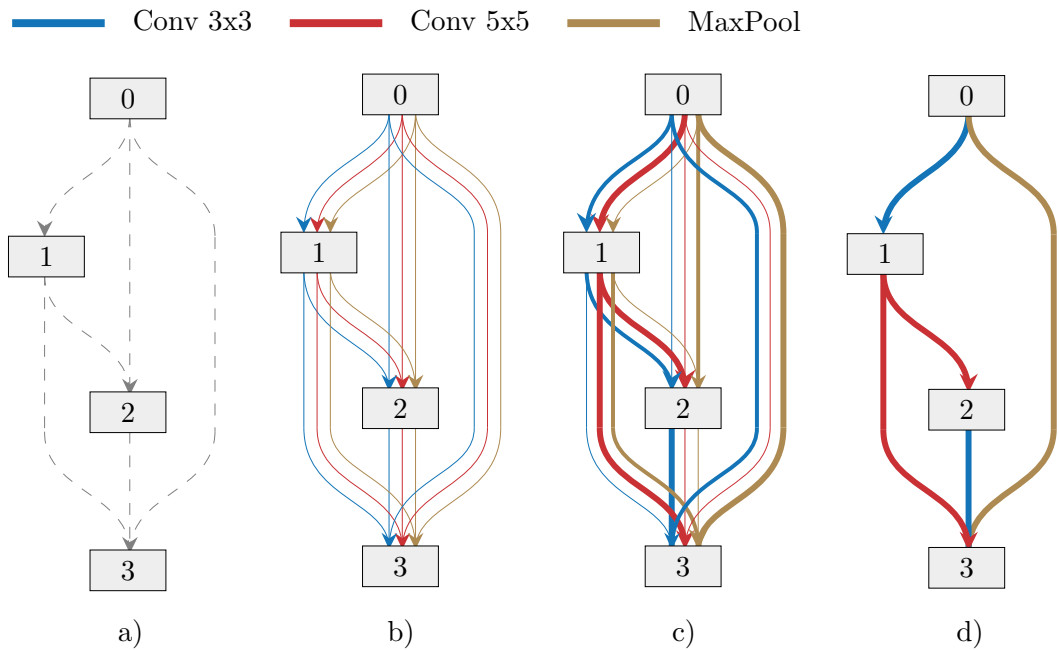
\begin{figure}[h]
    \centering
    \input{fig/darts}
    \caption{An example illustration of DARTS where we have a simple network with an input node (denoted as 0), two hidden nodes (denoted as 1, 2) and one output node (denoted as 4): (a) At the beginning, operations on the edges are unknown. b) Next, instead of a single operation on each edge, we place a mixture of candidate operations. c) Network weights and architecture parameters are jointly optimized. d) The final discrete architecture is obtained by replacing each mixed operation $\bar{o}^{(i,j)}$ with the most likely operation $o^{(i,j)}$ from the learned mixing probabilities.}
    \label{fig:darts}
\end{figure}

Differentiable architecture search (DARTS), proposed by \cite{liu2019darts}, addresses the issue of the discrete search strategy by using a continuous relaxation of the architecture representation which enables direct gradient-based optimization. The algorithm first starts with a single large network representing the complete search space and through gradient descent and bi-level optimization it discovers the optimal sub-graph, as shown in Figure \ref{fig:darts}.

%Differentiable architecture search (DARTS), proposed by \cite{liu2019darts}, addresses the issue of the discrete search strategy by formulating the task in a differential manner.  While the search space of traditional NAS is discrete, DARTS uses a continuous relaxation of the architecture representation which enables direct gradient-based optimization. The algorithm first starts with a single large network representing the complete search space and through gradient descent it discovers the optimal sub-graph, as shown in Figure \ref{fig:darts}.

DARTS uses cells (Figure \ref{fig:search-space}c) as basic building blocks. To form the final architecture, cells are repeatedly stacked together until reaching the desired depth. This approach follows architectures like NASNet \citep{zoph2017learning} and GoogleNet \citep{szegedy2014going}. A cell can be thought of as a topologically ordered sequence of $N$ nodes connected to form a \textit{directed acyclic graph} (DAG). Each cell has two input nodes and a single output node, which is the concatenation result of all intermediate nodes. Each intermediate node $x^{(j)}$ is a latent representation (\eg a feature map in convolutional networks) and each directed edge $e^{(i, j)}$ is associated with some operation $o^{(i,j)}\in\mathcal{O}$ (\eg convolution, pooling) that transforms $x^{(i)}$ into $x^{(j)}$. $\mathcal{O}$ is the set of all candidate operations. Each intermediate node is computed based on its predecessors:

\begin{equation} 
\label{dartsIntermediateNode1}
x^{(j)}=\sum_{i<j}o^{(i,j)}\left( x^{(i)}\right) 
\end{equation}

In continuous relaxation, instead of having a single operation between two nodes, all possible candidate operations are used. To model this in the Figure \ref{fig:darts}, multiple edges between two nodes are kept, each corresponding to a particular operation. DARTS relaxes the categorical choice of a particular operation as a \textit{softmax} function over all possible operations and the task of architecture search is reduced to learning a set of mixed operations $\bar{o}^{(i,j)}(x)$ applied to a feature map $x$:

\begin{equation} 
\label{dartsIntermediateNode2}
\bar{o}^{(i,j)}(x) = \sum_{o\in\mathcal{O}} \frac{\exp(\alpha^{(i,j)}_o)}{\sum_{o'\in\mathcal{O}} \exp(\alpha_{o'}^{(i,j)})} o(x)
\end{equation}

where $\alpha^{(i,j)}_o$ is a vector of dimensions $ |\mathcal{O}|$, containing the weights $w$ of all operations $o$ on the directed edge $e^{(i, j)}$.

After relaxation, DARTS jointly learns the architecture $\alpha$ and the weights $w$ within all the mixed operations. The training and the validation loss are denoted by $\mathcal{L}_\text{train}$ and $\mathcal{L}_\text{val}$ respectively. Both losses are determined by the architecture parameters $\alpha$ and also by the network weights $w$. The goal for architecture search is to find the optimal weights $\alpha^*$ that minimize the validation loss $\mathcal{L}_{val}(w^*, \alpha^*)$, where the weights $w^*$ associated with the architecture are obtained by minimizing the training loss:

\begin{equation} \label{darts3}
w^*=\arg\min_w\mathcal{L}_{\text{train}}(w, \alpha^*)
\end{equation}

This represents a bilevel optimization problem with $\alpha$ as the upper-level variable and $w$ as the lower-level variable:

\begin{equation} 
\label{bi-level optimization}
\begin{aligned}
\min_\alpha\quad & \mathcal{L}_\text{val} (w^*(\alpha), \alpha) \\
\text{s.t.}\quad & w^*(\alpha) = \arg\min_w \mathcal{L}_\text{train} (w, \alpha)
\end{aligned}
\end{equation}

In the the outer loop of a bilevel optimization we are looking for the $\alpha$ that achieves the minimal validation loss. In the inner loop, we are optimizing network weights, $w$, for this particular $\alpha$ by minimizing the training loss. Because the exact evaluation of the architecture gradient can be prohibitively expensive due to the expensive inner optimization, the authors use a simple approximation scheme as follows:

\begin{equation} 
\label{bi-level optimization2}
\begin{aligned}
 & \nabla_{\alpha}\mathcal{L}_{val}(w^*(\alpha),\alpha) \\
\approx & \nabla_{\alpha}\mathcal{L}_{val}(w-\xi\nabla_w\mathcal{L}_\text{train}(w, \alpha),\alpha)
\end{aligned}
\end{equation}

where $w$ denotes the current weights maintained by the algorithm, and $\xi$ is the learning rate for a step of inner optimization. The idea behind this is that the base models are updated with a single training step instead of full stochastic gradient descent to convergence.

At the end of the search, some $\alpha$’s of some edges become much larger than the others. A discrete architecture can be obtained by replacing each mixed operation $\bar{o}^{(i,j)}$ with the most likely operation $o^{(i,j)}$ on the edge $e^{(i,j)}$ while all other operations are discarded:

\begin{equation} 
\label{dartsIntermediateNode3}
{o}^{(i,j)}=\arg\max_{o\in\mathcal{O}} \alpha_{(i,j)}^o
\end{equation}

DARTS limitations have been well documented by the research community \citep{Zela2019Understanding, Hanwen2019DARTS+, Xiangxiang2019FairDARTS, chen2019pdarts, Hundt2019sharpDARTS, pmlr-v119-chen20f}. The most pressing issue is DARTS' instability when the search epochs becomes large as the performance is prone to collapsing and \textit{skip-connections} can become the dominant operator in the architecture. The latter is problematic since the \textit{skip-connect} operator has a weak ability to learn as it is parameter-free. \cite{chen2019pdarts} also found that normal cells discovered with DARTS tend to keep shallow connections, which can be attributed to the fact that shallow networks enjoy faster gradient descent during architecture search. This contradicts previous findings that deeper networks in principle perform better \citep{simonyan2014deep,szegedy2014going}.

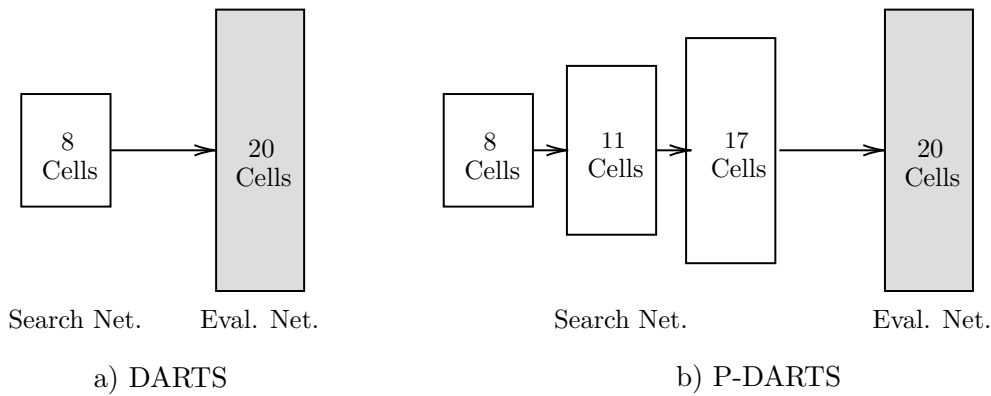
\begin{figure}[H]
    \centering
    \input{fig/darts_vs_pdarts}
    \caption{Difference between DARTS and P-DARTS. While DARTS (left) is searching architecture in shallow settings and evaluating them in deep settings, P-DARTS (right) progressively increase the searching depth to bridge the \textit{depth gap} between search and evaluation.}
    \label{fig:darts-vs-pdarts}
\end{figure}

Moreover, the size of the DARTS' search network is limited by the size of the GPU memory. This forces DARTS to search in a much shallower network and evaluating it in the deeper one (Figure \ref{fig:darts-vs-pdarts}a). Shallow networks tend to behave differently, which means that architectures found during the search may not be the optimal architecture for evaluation \citep{chen2019pdarts}. \cite{chen2019pdarts} named this phenomenon the \textit{depth gap}.

%%%%%%%%%%%%%%%%%%%%%%%%%%%%%%
%
% Progressive DARTS
%
%%%%%%%%%%%%%%%%%%%%%%%%%%%%%%

\subsection{Progressive DARTS} \label{p-darts}

Progressive DARTS (P-DARTS) is a novel and efficient algorithm by \cite{chen2019pdarts} where the search depth is progressively increased until the depth of the search network is close to the depth of the final network (Figure \ref{fig:darts-vs-pdarts}b). P-DARTS solves the problem of searching the architecture in a shallow network and evaluating it in the deeper one. While directly increasing the depth of a searched network may sound tempting, it poses major obstacles, namely GPU memory limitations and a bias towards the \textit{skip-connect} operation. P-DARTS solves these problems by applying search space approximation and search space regularization schemes, which we explain next. Figure \ref{fig:darts-vs-pdarts} illustrates the differences between DARTS and P-DARTS.

\begin{figure}[h]
    \centering
     \includegraphics[width=\linewidth]{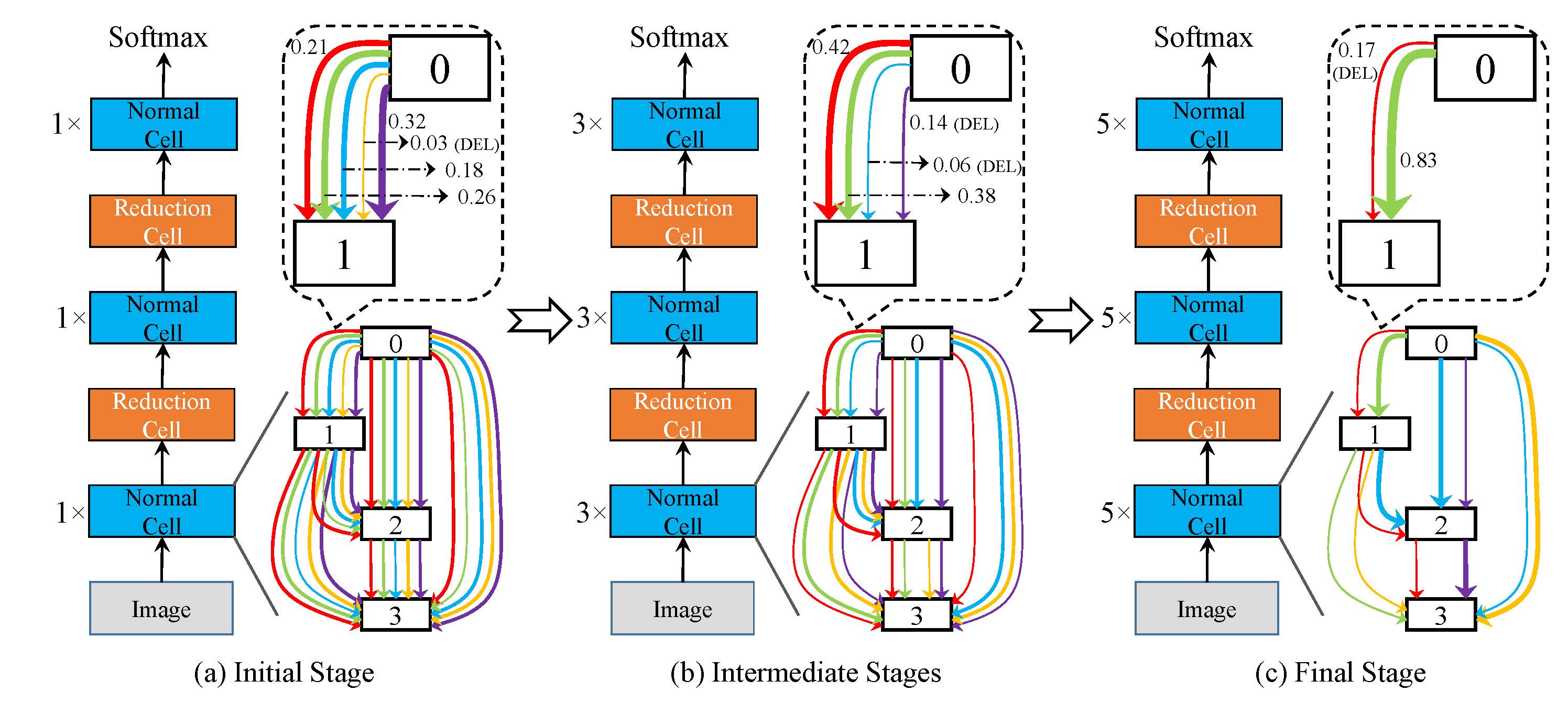}
    \caption{An overview of the whole P-DARTS pipeline \citep{chen2019pdarts}. At the initial stage, we start with a shallow network and a full set of candidate operations. At every subsequent stage, we increase the depth of the search network and at the same time also remove the lowest scoring candidate operations. This process is then repeated until reaching the final stage. The final architecture is obtained on the final stage by replacing each mixed operation with the most likely operation.}
    \label{fig:pdarts}
\end{figure}

%%%%%%%%%%%%%%%%%%%%%%%%%%%%%%
% Search space approximation
%%%%%%%%%%%%%%%%%%%%%%%%%%%%%%

\subsubsection{Search Space Approximation} \label{search-space-approximation}

DARTS' GPU memory usage is proportional to the depth of the searched networks, which is limited by the size of GPU memory. However, if we increase the search depth, while at the same time reducing the search space with respect to the candidate operations at the end of each stage, we can overcome the problem of the GPU limit. 

P-DARTS's search space approximation splits the search process into multiple stages, where for each stage, $\mathfrak{S}_k$, the search network consists of $L_k$ cells and the size of the operations space is $O_k$, \textit{i.e.}, $|\mathcal{O}_{(i,j)}^{k}|=O_k$. We start with an initial stage where the search network is relatively shallow but the operation space is large ($\mathcal{O}_{(i,j)}^{1}=\mathcal{O}$). After each stage, $\mathfrak{S}_{k-1}$, the depth of architecture is increased by stacking more cells, $L_k>L_{k-1}$, and the operation space is approximated by dropping candidate operations with lower weights learned during the previous stage, $\mathfrak{S}_{k-1}$. In other words, operations set for the current step, $\mathcal{O}_{(i,j)}^{k}$, have smaller size than operations set of previous step, $\mathcal{O}_{(i,j)}^{k-1}$. This process is repeated until the desired depth is achieved. The whole pipeline can be observed in Figure \ref{fig:pdarts}. While we can search a network on the optimal depth (\ie{the depth of the evaluation network}), this is often not possible due to the hard limit of a GPU memory. Only at the final stage do we determine the final topology according to the learned architecture parameters $\alpha_K$.

%%%%%%%%%%%%%%%%%%%%%%%%%%%%%%
% Search Space Regularization
%%%%%%%%%%%%%%%%%%%%%%%%%%%%%%

\subsubsection{Search Space Regularization} \label{search-space-regularization}

DARTS tends to be biased towards the \textit{skip-connect} operation because it accelerates for-ward/backward propagation and often leads to a faster way of gradient descent during the search process \citep{chen2019pdarts}. However, the \textit{skip-connect} operation has a relatively weak ability to learn any visual representation. To address this problem, a search space regularization scheme was proposed where operation-level Dropout \citep{Srivastava2014dropout} was added to restrict the number of \textit{skip-connect} operations and prevent the architecture from over-fitting. 

At the beginning of every stage, $\mathfrak{S}_k$, P-DARTS drops all network parameters learned in the previous stage, $\mathfrak{S}_{k-1}$, and trains the architecture from scratch. To avoid architectures with many \textit{skip-connect} operations, P-DARTS employs a search space regularization scheme, which consists of two parts:

\begin{enumerate}

\item \textbf{Operation-level Dropout} is added to partially cut off the straightforward path through \textit{skip-connect} and to force the algorithm to explore other alternatives. The Dropout rate gradually decays during the training process, thus the straightforward path through \textit{skip-connect} is blocked at the beginning and treated equally afterwards when parameters of other operations are well learned. This approach leaves the algorithm to itself to make the final decision.

\item \textbf{Architecture refinement} is employed after the final search step is complete. It controls the number of preserved \textit{skip-connects} to be a constant $M$. This is achieved through an iterative process where we first construct a cell topography using the standard DARTS algorithm. Next, we search for the $M$ \textit{skip-connect} operations with the largest architecture weights, $\alpha$, in the cell topology and set the $\alpha$'s of others to 0. Finally, we redo cell construction with modified architecture parameters. As this might bring up other \textit{skip-connections}, we repeat this process until the desired number $M$ is achieved.

\end{enumerate}

%%%%%%%%%%%%%%%%%%%%%%%%%%%%%%
%
% META-LEARNING
%
%%%%%%%%%%%%%%%%%%%%%%%%%%%%%%

\subsection{Meta-learning} \label{meta-learning}

Meta-learning, or \textit{learning to learn}, is the science of finding relationships between datasets and learning algorithms based on knowledge gained from past experience with prior tasks \citep{vanschoren2018mlsurvey}. In other words, the term meta-learning applies to any type of learning based on prior experience that can improve the design of machine learning pipelines and help learn new tasks much faster \citep{vanschoren2018mlsurvey, Brazdil2003}. \cite{Lemke2013} define meta-learning as a system that must include a learning subsystem, which adapts with experience. The experience itself is gained by exploiting meta-knowledge extracted in previous learning episodes on a single dataset, or from different domains and problems. Meta-learning dates back to the 1990s but has recently gained popularity due to its implementation in hyperparameter and neural network optimization, finding good network architectures, few-shot image recognition, and fast reinforcement learning.

When transferring knowledge gained on the source task to the target tasks, we should ensure that the tasks are related. Transferring knowledge between unrelated tasks could cause the “\textit{negative transfer}” effect \citep{Rosenstein2005}. To determent the distance between tasks, a task similarity measure has to be computed. However, despite the apparent simpleness of the concept, this still poses an open problem. This problem can be illustrated by comparing distinct datasets of pictures containing aircrafts, flowers, and birds that have different cardinality, dimensionality, and label space. Computing the distances of this illustrative problem is far from trivial since it is hard to quantify when two tasks are similar and when they are not.

A common approach is to compare datasets via proxies \citep{kim2017similarity,zamir2018taskonomy,houben2019msc}. For example, one can compare the learning curves of the training process on a pre-specified model as we can assume that similar tasks have similar learning curves \citep{Leite2007}. Another way is to compare the Fisher information metric associated with pre-specified deep neural network parameters \citep{achille2019task2vec}. The latter approach is also used in this work and explained in more detail in the next section. Methods based on proxies often ignore task labels, are architecture dependant and require training a model on each dataset. 

Recently, \cite{alvarezmelis2020geometric} introduced a new task similarity measure that is based on the optimal transport (OT) distances \citep{Villani2008} on both the samples and their labels. OT is an approach to compare probability distributions. The key idea is to compare distributions over feature-label pairs. First, the authors treat the labels as a distribution and use the Wasserstein distance to measure the labels. Then, they wrap the Wasserstein distance over the labels with the Euclidean distance over the samples in the sample space into the ground metric for the OT problem in the outer layer. Compared to the Task2Vec (the method used in this work) this method is model agnostic, does not involve training, and does not ignore image labels even if they are completely
unrelated or disjoint between tasks.

\subsection{Task2Vec}

Task2Vec \citep{achille2019task2vec} tackles the problem of determining relationships between visual tasks. Task2Vec by itself computes fixed-dimensional embeddings of visual classification tasks as vectors in a real vector space. These embeddings can be then used to argue about the nature and the relationship between tasks independently from the number of classes and class label semantics. Embeddings are computed based on estimates of the Fisher information matrix (FIM) associated with the deep neural network parameters. FIM is a Riemannian metric on the space of probability distributions and is a way of measuring the amount of information a particular parameter (weight or feature) contains about the joint distribution $p_w(x,y)=\hat{p}(x)p_w(y|x)$ \citep{book,achille2019task2vec}. 

Since the FIMs are not comparable if computed on different networks, the authors introduced an approach of using a single pre-trained \textit{probe network} as a feature extractor. The architecture and weights of a probe network are fixed to provide a fixed-dimensional representation of the task and only the classifier layer is re-trained. The intuition behind this is that the FIM provides information about the sensitivity of the task preference to small disturbances of parameters in the probe network. To embed task using Fisher information we train a classifier with the task loss on features from a probe network. Next, the gradients of a probe network with respect to the task loss are computed. Finally, fixed dimensional task embedding is obtained by using statistics of the probe parameter gradients.

When predicting a label $y$ for an image $x$ in a deep network, not all network weights, $w$, are equally useful. The importance of weight for the task can be quantified by considering a perturbation $w'=w+\delta w$ of the weights, and measuring the average Kullbach-Leibler (KL) divergence between the original output distribution $p_w(y|x)$ and the perturbed one $p_{w'}(y|x)$. The second-order approximation of this is:

\begin{equation} 
\label{FIM}
E_{x \sim \hat{p}}KL(p_{w'}(y|x)||p_w(y|x))+\delta w \cdot F \delta w + o(\delta w^2)
\end{equation}

where F is the Fisher information matrix:

\begin{equation} 
\label{FIM2}
F = E_{x \sim dataset} E_{y \sim p_w(y|x)} [\nabla_w log p_w(y|x) \nabla_w log p_w(y|x)^t]
\end{equation}

where $p_w(y|x)$ is the output probability vector of the network and $w$ are the weights of the network.

However, if the rich probe networks are used (\ie{networks based on CNN}), the full FIM becomes too large to be useful. To mitigate this issue, \cite{achille2019task2vec} introduced two approximations:
\begin{enumerate}
\item Only the diagonal entries are considered as we assume that correlations between different filters in the probe network are not important.
\item Robust FIM estimation is used where we estimate $\Lambda$ of a Gaussian perturbation:

\begin{equation} 
\label{FIM3}
L(\hat{w}:\Lambda)=\mathbb{E}_{w\sim \mathcal{N}(\hat{w} \Lambda)}[H_{p_{w,\hat{p}}}p(y|x)] + \beta KL(\mathcal{N}(0,\Lambda)||\mathcal{N}(0,\lambda^2I))
\end{equation}

where optimal $\Lambda$ satisfies:

\begin{equation} 
\label{FIM4}
\dfrac{\beta}{2 N} \sim F+ \dfrac{\beta \lambda^2}{2N}I
\end{equation}

Therefore, $\dfrac{\beta}{2 N} \sim F+o(1)$ can be used as an an estimator of the FIM $F$ .
\end{enumerate}

Task2Vec embeddings only represent the space of tasks. Choosing the metric to compute similarity measure in that space depends on the meta-task we are considering. A robust distance computation presented by the authors and used in this work is the cosine distance between normalized embeddings:

\begin{equation} 
\label{symT2Vdistance}
d_\text{sym}(F_a,F_b)=d_\text{cos}\left(\frac{F_a}{F_a+F_b},\frac{F_b}{F_a+F_b}\right),
\end{equation}
 
where $ d_{cos} $ is the cosine distance, $F_a$ and $F_b$ are embeddings (\ie{the diagonal of the FIM computed on the same probe network}) of tasks $ t_a $ and $ t_b $, and the division is element-wise.

%% file: fig/nas.tex
\tikzset{every picture/.style={line width=0.75pt}} %set default line width to 0.75pt        
    \begin{tikzpicture}[x=0.75pt,y=0.75pt,yscale=-1,xscale=1]
        %uncomment if require: \path (0,129); %set diagram left start at 0, and has height of 129
        
        % Text Node
        \draw (297,3) node [anchor=north west][inner sep=0.75pt]  [font=\small] [align=left] {\begin{minipage}[lt]{50.660000000000004pt}\setlength\topsep{0pt}
        \begin{center}
        architecture\\$\displaystyle A\ \in \mathcal{A}$
        \end{center}
        \end{minipage}};
        % Text Node
        \draw (291,83) node [anchor=north west][inner sep=0.75pt]  [font=\small] [align=left] {\begin{minipage}[lt]{57.95864400000001pt}\setlength\topsep{0pt}
        \begin{center}
        performance\\estimate of $\displaystyle A$
        \end{center}
        \end{minipage}};
        % Text Node
        \draw    (-2,33) -- (112,33) -- (112,87) -- (-2,87) -- cycle  ;
        \draw (55,60) node   [align=left] {\begin{minipage}[lt]{74.80000000000001pt}\setlength\topsep{0pt}
        \begin{center}
        Search \\Space $\displaystyle \mathcal{A}$
        \end{center}
        \end{minipage}};
        % Text Node
        \draw    (168,33) -- (272,33) -- (272,87) -- (168,87) -- cycle  ;
        \draw (220,60) node   [align=left] {\begin{minipage}[lt]{68pt}\setlength\topsep{0pt}
        \begin{center}
        Search \\Strategy
        \end{center}
        \end{minipage}};
        % Text Node
        \draw    (388,23) -- (492,23) -- (492,97) -- (388,97) -- cycle  ;
        \draw (440,60) node   [align=left] {\begin{minipage}[lt]{68pt}\setlength\topsep{0pt}
        \begin{center}
        Performance \\Estimation\\Startegy
        \end{center}
        \end{minipage}};
        % Connection
        \draw    (112,60) -- (166,60) ;
        \draw [shift={(168,60)}, rotate = 180] [color={rgb, 255:red, 0; green, 0; blue, 0 }  ][line width=0.75]    (8.74,-2.63) .. controls (5.56,-1.12) and (2.65,-0.24) .. (0,0) .. controls (2.65,0.24) and (5.56,1.12) .. (8.74,2.63)   ;
        % Connection
        \draw    (272,47.67) .. controls (310.37,39.12) and (348.46,38.98) .. (386.27,47.25) ;
        \draw [shift={(388,47.64)}, rotate = 192.71] [color={rgb, 255:red, 0; green, 0; blue, 0 }  ][line width=0.75]    (8.74,-2.63) .. controls (5.56,-1.12) and (2.65,-0.24) .. (0,0) .. controls (2.65,0.24) and (5.56,1.12) .. (8.74,2.63)   ;
        % Connection
        \draw    (388,72.32) .. controls (350.86,80.74) and (312.78,80.83) .. (273.78,72.6) ;
        \draw [shift={(272,72.22)}, rotate = 372.26] [color={rgb, 255:red, 0; green, 0; blue, 0 }  ][line width=0.75]    (8.74,-2.63) .. controls (5.56,-1.12) and (2.65,-0.24) .. (0,0) .. controls (2.65,0.24) and (5.56,1.12) .. (8.74,2.63)   ;
    \end{tikzpicture}

%% file: fig/search_space.tex
\tikzset{every picture/.style={line width=0.75pt}} %set default line width to 0.75pt        

\begin{tikzpicture}[x=0.75pt,y=0.75pt,yscale=-1,xscale=1]
%uncomment if require: \path (0,414); %set diagram left start at 0, and has height of 414

%Shape: Rectangle [id:dp29455613720717166] 
\draw  [color={rgb, 255:red, 155; green, 155; blue, 155 }  ,draw opacity=1 ][dash pattern={on 0.84pt off 2.51pt}][line width=0.75]  (343.32,48.17) .. controls (343.32,45.96) and (345.11,44.17) .. (347.32,44.17) -- (420.64,44.17) .. controls (422.85,44.17) and (424.64,45.96) .. (424.64,48.17) -- (424.64,121.49) .. controls (424.64,123.7) and (422.85,125.49) .. (420.64,125.49) -- (347.32,125.49) .. controls (345.11,125.49) and (343.32,123.7) .. (343.32,121.49) -- cycle ;
%Shape: Rectangle [id:dp8317376648145836] 
\draw  [color={rgb, 255:red, 155; green, 155; blue, 155 }  ,draw opacity=1 ][dash pattern={on 0.84pt off 2.51pt}][line width=0.75]  (343.32,136.89) .. controls (343.32,134.68) and (345.11,132.89) .. (347.32,132.89) -- (420.64,132.89) .. controls (422.85,132.89) and (424.64,134.68) .. (424.64,136.89) -- (424.64,210.21) .. controls (424.64,212.42) and (422.85,214.21) .. (420.64,214.21) -- (347.32,214.21) .. controls (345.11,214.21) and (343.32,212.42) .. (343.32,210.21) -- cycle ;

% Text Node
\draw  [fill={rgb, 255:red, 255; green, 255; blue, 255 }  ,fill opacity=1 ]  (-2,6.08) -- (29,6.08) -- (29,37.08) -- (-2,37.08) -- cycle  ;
\draw (13.5,21.58) node   [align=left] {\begin{minipage}[lt]{18.439356000000004pt}\setlength\topsep{0pt}

\end{minipage}};
% Text Node
\draw  [fill={rgb, 255:red, 238; green, 238; blue, 238 }  ,fill opacity=1 ]  (-2,60.3) -- (29,60.3) -- (29,91.3) -- (-2,91.3) -- cycle  ;
\draw (13.5,75.8) node   [align=left] {\begin{minipage}[lt]{18.439356000000004pt}\setlength\topsep{0pt}

\end{minipage}};
% Text Node
\draw  [fill={rgb, 255:red, 238; green, 238; blue, 238 }  ,fill opacity=1 ]  (-2,168.73) -- (29,168.73) -- (29,199.73) -- (-2,199.73) -- cycle  ;
\draw (13.5,184.23) node   [align=left] {\begin{minipage}[lt]{18.439356000000004pt}\setlength\topsep{0pt}

\end{minipage}};
% Text Node
\draw (13.5,130.01) node   [align=left] {\begin{minipage}[lt]{18.439356000000004pt}\setlength\topsep{0pt}
\begin{center}
$\displaystyle \vdots $
\end{center}

\end{minipage}};
% Text Node
\draw  [fill={rgb, 255:red, 204; green, 204; blue, 204 }  ,fill opacity=1 ]  (-2,222.95) -- (29,222.95) -- (29,253.95) -- (-2,253.95) -- cycle  ;
\draw (13.5,238.45) node   [align=left] {\begin{minipage}[lt]{18.439356000000004pt}\setlength\topsep{0pt}

\end{minipage}};
% Text Node
\draw  [fill={rgb, 255:red, 238; green, 238; blue, 238 }  ,fill opacity=1 ]  (115.47,114.51) -- (146.47,114.51) -- (146.47,145.51) -- (115.47,145.51) -- cycle  ;
\draw (130.97,130.01) node   [align=left] {\begin{minipage}[lt]{18.439356000000004pt}\setlength\topsep{0pt}

\end{minipage}};
% Text Node
\draw  [fill={rgb, 255:red, 255; green, 255; blue, 255 }  ,fill opacity=1 ]  (169.69,60.3) -- (200.69,60.3) -- (200.69,91.3) -- (169.69,91.3) -- cycle  ;
\draw (185.19,75.8) node   [align=left] {\begin{minipage}[lt]{18.439356000000004pt}\setlength\topsep{0pt}

\end{minipage}};
% Text Node
\draw  [fill={rgb, 255:red, 238; green, 238; blue, 238 }  ,fill opacity=1 ]  (169.69,114.51) -- (200.69,114.51) -- (200.69,145.51) -- (169.69,145.51) -- cycle  ;
\draw (185.19,130.01) node   [align=left] {\begin{minipage}[lt]{18.439356000000004pt}\setlength\topsep{0pt}

\end{minipage}};
% Text Node
\draw  [fill={rgb, 255:red, 204; green, 204; blue, 204 }  ,fill opacity=1 ]  (169.69,168.73) -- (200.69,168.73) -- (200.69,199.73) -- (169.69,199.73) -- cycle  ;
\draw (185.19,184.23) node   [align=left] {\begin{minipage}[lt]{18.439356000000004pt}\setlength\topsep{0pt}

\end{minipage}};
% Text Node
\draw  [fill={rgb, 255:red, 238; green, 238; blue, 238 }  ,fill opacity=1 ]  (223.9,114.51) -- (254.9,114.51) -- (254.9,145.51) -- (223.9,145.51) -- cycle  ;
\draw (239.4,130.01) node   [align=left] {\begin{minipage}[lt]{18.439356000000004pt}\setlength\topsep{0pt}

\end{minipage}};
% Text Node
\draw (9.16,281) node [anchor=north west][inner sep=0.75pt]   [align=left] {a)};
% Text Node
\draw (180.85,280.1) node [anchor=north west][inner sep=0.75pt]   [align=left] {b)};
% Text Node
\draw (379.69,281) node [anchor=north west][inner sep=0.75pt]   [align=left] {c)};
% Text Node
\draw  [fill={rgb, 255:red, 255; green, 255; blue, 255 }  ,fill opacity=1 ]  (368.48,-2.96) -- (399.48,-2.96) -- (399.48,28.04) -- (368.48,28.04) -- cycle  ;
\draw (383.98,12.54) node   [align=left] {\begin{minipage}[lt]{18.439356000000004pt}\setlength\topsep{0pt}

\end{minipage}};
% Text Node
\draw  [fill={rgb, 255:red, 238; green, 238; blue, 238 }  ,fill opacity=1 ][line width=0.75]   (346.28,75.83) -- (364.28,75.83) -- (364.28,93.83) -- (346.28,93.83) -- cycle  ;
\draw (355.28,84.83) node   [align=left] {\begin{minipage}[lt]{9.759036144578195pt}\setlength\topsep{0pt}

\end{minipage}};
% Text Node
\draw  [fill={rgb, 255:red, 255; green, 255; blue, 255 }  ,fill opacity=1 ][line width=0.75]   (374.98,47.13) -- (392.98,47.13) -- (392.98,65.13) -- (374.98,65.13) -- cycle  ;
\draw (383.98,56.13) node   [align=left] {\begin{minipage}[lt]{9.75903614457835pt}\setlength\topsep{0pt}

\end{minipage}};
% Text Node
\draw  [fill={rgb, 255:red, 238; green, 238; blue, 238 }  ,fill opacity=1 ][line width=0.75]   (374.98,75.83) -- (392.98,75.83) -- (392.98,93.83) -- (374.98,93.83) -- cycle  ;
\draw (383.98,84.83) node   [align=left] {\begin{minipage}[lt]{9.75903614457835pt}\setlength\topsep{0pt}

\end{minipage}};
% Text Node
\draw  [fill={rgb, 255:red, 204; green, 204; blue, 204 }  ,fill opacity=1 ][line width=0.75]   (374.98,104.53) -- (392.98,104.53) -- (392.98,122.53) -- (374.98,122.53) -- cycle  ;
\draw (383.98,113.53) node   [align=left] {\begin{minipage}[lt]{9.75903614457835pt}\setlength\topsep{0pt}

\end{minipage}};
% Text Node
\draw  [fill={rgb, 255:red, 238; green, 238; blue, 238 }  ,fill opacity=1 ][line width=0.75]   (403.68,75.83) -- (421.68,75.83) -- (421.68,93.83) -- (403.68,93.83) -- cycle  ;
\draw (412.68,84.83) node   [align=left] {\begin{minipage}[lt]{9.75903614457835pt}\setlength\topsep{0pt}

\end{minipage}};
% Text Node
\draw  [fill={rgb, 255:red, 238; green, 238; blue, 238 }  ,fill opacity=1 ][line width=0.75]   (346.28,164.55) -- (364.28,164.55) -- (364.28,182.55) -- (346.28,182.55) -- cycle  ;
\draw (355.28,173.55) node   [align=left] {\begin{minipage}[lt]{9.759036144578195pt}\setlength\topsep{0pt}

\end{minipage}};
% Text Node
\draw  [fill={rgb, 255:red, 255; green, 255; blue, 255 }  ,fill opacity=1 ][line width=0.75]   (374.98,135.85) -- (392.98,135.85) -- (392.98,153.85) -- (374.98,153.85) -- cycle  ;
\draw (383.98,144.85) node   [align=left] {\begin{minipage}[lt]{9.75903614457835pt}\setlength\topsep{0pt}

\end{minipage}};
% Text Node
\draw  [fill={rgb, 255:red, 238; green, 238; blue, 238 }  ,fill opacity=1 ][line width=0.75]   (374.98,164.55) -- (392.98,164.55) -- (392.98,182.55) -- (374.98,182.55) -- cycle  ;
\draw (383.98,173.55) node   [align=left] {\begin{minipage}[lt]{9.75903614457835pt}\setlength\topsep{0pt}

\end{minipage}};
% Text Node
\draw  [fill={rgb, 255:red, 204; green, 204; blue, 204 }  ,fill opacity=1 ][line width=0.75]   (374.98,193.25) -- (392.98,193.25) -- (392.98,211.25) -- (374.98,211.25) -- cycle  ;
\draw (383.98,202.25) node   [align=left] {\begin{minipage}[lt]{9.75903614457835pt}\setlength\topsep{0pt}

\end{minipage}};
% Text Node
\draw  [fill={rgb, 255:red, 238; green, 238; blue, 238 }  ,fill opacity=1 ][line width=0.75]   (403.68,164.55) -- (421.68,164.55) -- (421.68,182.55) -- (403.68,182.55) -- cycle  ;
\draw (412.68,173.55) node   [align=left] {\begin{minipage}[lt]{9.75903614457835pt}\setlength\topsep{0pt}

\end{minipage}};
% Text Node
\draw  [fill={rgb, 255:red, 204; green, 204; blue, 204 }  ,fill opacity=1 ]  (368.48,231.98) -- (399.48,231.98) -- (399.48,262.98) -- (368.48,262.98) -- cycle  ;
\draw (383.98,247.48) node   [align=left] {\begin{minipage}[lt]{18.439356000000004pt}\setlength\topsep{0pt}

\end{minipage}};
% Connection
\draw    (13.5,37.08) -- (13.5,58.3) ;
\draw [shift={(13.5,60.3)}, rotate = 270] [color={rgb, 255:red, 0; green, 0; blue, 0 }  ][line width=0.75]    (8.74,-2.63) .. controls (5.56,-1.12) and (2.65,-0.24) .. (0,0) .. controls (2.65,0.24) and (5.56,1.12) .. (8.74,2.63)   ;
% Connection
\draw    (13.5,91.3) -- (13.5,112.51) ;
\draw [shift={(13.5,114.51)}, rotate = 270] [color={rgb, 255:red, 0; green, 0; blue, 0 }  ][line width=0.75]    (8.74,-2.63) .. controls (5.56,-1.12) and (2.65,-0.24) .. (0,0) .. controls (2.65,0.24) and (5.56,1.12) .. (8.74,2.63)   ;
% Connection
\draw    (13.5,145.51) -- (13.5,166.73) ;
\draw [shift={(13.5,168.73)}, rotate = 270] [color={rgb, 255:red, 0; green, 0; blue, 0 }  ][line width=0.75]    (8.74,-2.63) .. controls (5.56,-1.12) and (2.65,-0.24) .. (0,0) .. controls (2.65,0.24) and (5.56,1.12) .. (8.74,2.63)   ;
% Connection
\draw    (13.5,199.73) -- (13.5,220.95) ;
\draw [shift={(13.5,222.95)}, rotate = 270] [color={rgb, 255:red, 0; green, 0; blue, 0 }  ][line width=0.75]    (8.74,-2.63) .. controls (5.56,-1.12) and (2.65,-0.24) .. (0,0) .. controls (2.65,0.24) and (5.56,1.12) .. (8.74,2.63)   ;
% Connection
\draw    (169.69,91.3) -- (147.88,113.1) ;
\draw [shift={(146.47,114.51)}, rotate = 315] [color={rgb, 255:red, 0; green, 0; blue, 0 }  ][line width=0.75]    (8.74,-2.63) .. controls (5.56,-1.12) and (2.65,-0.24) .. (0,0) .. controls (2.65,0.24) and (5.56,1.12) .. (8.74,2.63)   ;
% Connection
\draw    (185.19,91.3) -- (185.19,112.51) ;
\draw [shift={(185.19,114.51)}, rotate = 270] [color={rgb, 255:red, 0; green, 0; blue, 0 }  ][line width=0.75]    (8.74,-2.63) .. controls (5.56,-1.12) and (2.65,-0.24) .. (0,0) .. controls (2.65,0.24) and (5.56,1.12) .. (8.74,2.63)   ;
% Connection
\draw    (200.69,91.3) -- (222.49,113.1) ;
\draw [shift={(223.9,114.51)}, rotate = 225] [color={rgb, 255:red, 0; green, 0; blue, 0 }  ][line width=0.75]    (8.74,-2.63) .. controls (5.56,-1.12) and (2.65,-0.24) .. (0,0) .. controls (2.65,0.24) and (5.56,1.12) .. (8.74,2.63)   ;
% Connection
\draw    (185.19,145.51) -- (185.19,166.73) ;
\draw [shift={(185.19,168.73)}, rotate = 270] [color={rgb, 255:red, 0; green, 0; blue, 0 }  ][line width=0.75]    (8.74,-2.63) .. controls (5.56,-1.12) and (2.65,-0.24) .. (0,0) .. controls (2.65,0.24) and (5.56,1.12) .. (8.74,2.63)   ;
% Connection
\draw    (223.9,145.51) -- (202.1,167.31) ;
\draw [shift={(200.69,168.73)}, rotate = 315] [color={rgb, 255:red, 0; green, 0; blue, 0 }  ][line width=0.75]    (8.74,-2.63) .. controls (5.56,-1.12) and (2.65,-0.24) .. (0,0) .. controls (2.65,0.24) and (5.56,1.12) .. (8.74,2.63)   ;
% Connection
\draw    (146.47,145.51) -- (168.27,167.31) ;
\draw [shift={(169.69,168.73)}, rotate = 225] [color={rgb, 255:red, 0; green, 0; blue, 0 }  ][line width=0.75]    (8.74,-2.63) .. controls (5.56,-1.12) and (2.65,-0.24) .. (0,0) .. controls (2.65,0.24) and (5.56,1.12) .. (8.74,2.63)   ;
% Connection
\draw [line width=0.75]    (374.98,65.13) -- (365.69,74.42) ;
\draw [shift={(364.28,75.83)}, rotate = 315] [color={rgb, 255:red, 0; green, 0; blue, 0 }  ][line width=0.75]    (6.56,-1.97) .. controls (4.17,-0.84) and (1.99,-0.18) .. (0,0) .. controls (1.99,0.18) and (4.17,0.84) .. (6.56,1.97)   ;
% Connection
\draw [line width=0.75]    (383.98,65.13) -- (383.98,73.83) ;
\draw [shift={(383.98,75.83)}, rotate = 270] [color={rgb, 255:red, 0; green, 0; blue, 0 }  ][line width=0.75]    (6.56,-1.97) .. controls (4.17,-0.84) and (1.99,-0.18) .. (0,0) .. controls (1.99,0.18) and (4.17,0.84) .. (6.56,1.97)   ;
% Connection
\draw [line width=0.75]    (392.98,65.13) -- (402.27,74.42) ;
\draw [shift={(403.68,75.83)}, rotate = 225] [color={rgb, 255:red, 0; green, 0; blue, 0 }  ][line width=0.75]    (6.56,-1.97) .. controls (4.17,-0.84) and (1.99,-0.18) .. (0,0) .. controls (1.99,0.18) and (4.17,0.84) .. (6.56,1.97)   ;
% Connection
\draw [line width=0.75]    (383.98,93.83) -- (383.98,102.53) ;
\draw [shift={(383.98,104.53)}, rotate = 270] [color={rgb, 255:red, 0; green, 0; blue, 0 }  ][line width=0.75]    (6.56,-1.97) .. controls (4.17,-0.84) and (1.99,-0.18) .. (0,0) .. controls (1.99,0.18) and (4.17,0.84) .. (6.56,1.97)   ;
% Connection
\draw [line width=0.75]    (403.68,93.83) -- (394.4,103.12) ;
\draw [shift={(392.98,104.53)}, rotate = 315] [color={rgb, 255:red, 0; green, 0; blue, 0 }  ][line width=0.75]    (6.56,-1.97) .. controls (4.17,-0.84) and (1.99,-0.18) .. (0,0) .. controls (1.99,0.18) and (4.17,0.84) .. (6.56,1.97)   ;
% Connection
\draw [line width=0.75]    (364.28,93.83) -- (373.57,103.12) ;
\draw [shift={(374.98,104.53)}, rotate = 225] [color={rgb, 255:red, 0; green, 0; blue, 0 }  ][line width=0.75]    (6.56,-1.97) .. controls (4.17,-0.84) and (1.99,-0.18) .. (0,0) .. controls (1.99,0.18) and (4.17,0.84) .. (6.56,1.97)   ;
% Connection
\draw [line width=0.75]    (374.98,153.85) -- (365.69,163.14) ;
\draw [shift={(364.28,164.55)}, rotate = 315] [color={rgb, 255:red, 0; green, 0; blue, 0 }  ][line width=0.75]    (6.56,-1.97) .. controls (4.17,-0.84) and (1.99,-0.18) .. (0,0) .. controls (1.99,0.18) and (4.17,0.84) .. (6.56,1.97)   ;
% Connection
\draw [line width=0.75]    (383.98,153.85) -- (383.98,162.55) ;
\draw [shift={(383.98,164.55)}, rotate = 270] [color={rgb, 255:red, 0; green, 0; blue, 0 }  ][line width=0.75]    (6.56,-1.97) .. controls (4.17,-0.84) and (1.99,-0.18) .. (0,0) .. controls (1.99,0.18) and (4.17,0.84) .. (6.56,1.97)   ;
% Connection
\draw [line width=0.75]    (392.98,153.85) -- (402.27,163.14) ;
\draw [shift={(403.68,164.55)}, rotate = 225] [color={rgb, 255:red, 0; green, 0; blue, 0 }  ][line width=0.75]    (6.56,-1.97) .. controls (4.17,-0.84) and (1.99,-0.18) .. (0,0) .. controls (1.99,0.18) and (4.17,0.84) .. (6.56,1.97)   ;
% Connection
\draw [line width=0.75]    (383.98,182.55) -- (383.98,191.25) ;
\draw [shift={(383.98,193.25)}, rotate = 270] [color={rgb, 255:red, 0; green, 0; blue, 0 }  ][line width=0.75]    (6.56,-1.97) .. controls (4.17,-0.84) and (1.99,-0.18) .. (0,0) .. controls (1.99,0.18) and (4.17,0.84) .. (6.56,1.97)   ;
% Connection
\draw [line width=0.75]    (403.68,182.55) -- (394.4,191.84) ;
\draw [shift={(392.98,193.25)}, rotate = 315] [color={rgb, 255:red, 0; green, 0; blue, 0 }  ][line width=0.75]    (6.56,-1.97) .. controls (4.17,-0.84) and (1.99,-0.18) .. (0,0) .. controls (1.99,0.18) and (4.17,0.84) .. (6.56,1.97)   ;
% Connection
\draw [line width=0.75]    (364.28,182.55) -- (373.57,191.84) ;
\draw [shift={(374.98,193.25)}, rotate = 225] [color={rgb, 255:red, 0; green, 0; blue, 0 }  ][line width=0.75]    (6.56,-1.97) .. controls (4.17,-0.84) and (1.99,-0.18) .. (0,0) .. controls (1.99,0.18) and (4.17,0.84) .. (6.56,1.97)   ;
% Connection
\draw    (383.98,122.53) -- (383.98,133.85) ;
\draw [shift={(383.98,135.85)}, rotate = 270] [color={rgb, 255:red, 0; green, 0; blue, 0 }  ][line width=0.75]    (8.74,-2.63) .. controls (5.56,-1.12) and (2.65,-0.24) .. (0,0) .. controls (2.65,0.24) and (5.56,1.12) .. (8.74,2.63)   ;
% Connection
\draw    (383.98,28.04) -- (383.98,45.13) ;
\draw [shift={(383.98,47.13)}, rotate = 270] [color={rgb, 255:red, 0; green, 0; blue, 0 }  ][line width=0.75]    (8.74,-2.63) .. controls (5.56,-1.12) and (2.65,-0.24) .. (0,0) .. controls (2.65,0.24) and (5.56,1.12) .. (8.74,2.63)   ;
% Connection
\draw    (383.98,211.25) -- (383.98,229.98) ;
\draw [shift={(383.98,231.98)}, rotate = 270] [color={rgb, 255:red, 0; green, 0; blue, 0 }  ][line width=0.75]    (8.74,-2.63) .. controls (5.56,-1.12) and (2.65,-0.24) .. (0,0) .. controls (2.65,0.24) and (5.56,1.12) .. (8.74,2.63)   ;

\end{tikzpicture}

%% file: fig/darts.tex
\tikzset{every picture/.style={line width=0.1pt}} %set default line width to 0.75pt        

\begin{tikzpicture}[x=0.7pt,y=0.7pt,yscale=-1,xscale=1]
%uncomment if require: \path (0,363); %set diagram left start at 0, and has height of 363

%Curve Lines [id:da7504417814568738] 
\draw [color={rgb, 255:red, 201; green, 49; blue, 53 }  ,draw opacity=1 ]   (210.8,57.74) .. controls (210.45,89.76) and (167.86,90.72) .. (164.92,120.65) ;
\draw [shift={(164.76,123.52)}, rotate = 270.74] [fill={rgb, 255:red, 201; green, 49; blue, 53 }  ,fill opacity=1 ][line width=0.08]  [draw opacity=0] (8.93,-4.29) -- (0,0) -- (8.93,4.29) -- (5.93,0) -- cycle    ;
%Curve Lines [id:da8572905087116345] 
\draw [color={rgb, 255:red, 19; green, 117; blue, 183 }  ,draw opacity=1 ]   (203.65,57.74) .. controls (203.94,83.57) and (159.83,84.21) .. (157.69,120.64) ;
\draw [shift={(157.61,123.52)}, rotate = 269.89] [fill={rgb, 255:red, 19; green, 117; blue, 183 }  ,fill opacity=1 ][line width=0.08]  [draw opacity=0] (8.93,-4.29) -- (0,0) -- (8.93,4.29) -- (5.93,0) -- cycle    ;
%Curve Lines [id:da16276081262191666] 
\draw [color={rgb, 255:red, 173; green, 138; blue, 81 }  ,draw opacity=1 ]   (217.96,57.74) .. controls (218.07,96) and (175.37,97.16) .. (172.11,120.85) ;
\draw [shift={(171.92,123.52)}, rotate = 270.65] [fill={rgb, 255:red, 173; green, 138; blue, 81 }  ,fill opacity=1 ][line width=0.08]  [draw opacity=0] (8.93,-4.29) -- (0,0) -- (8.93,4.29) -- (5.93,0) -- cycle    ;
%Straight Lines [id:da3865411487324436] 
\draw [color={rgb, 255:red, 201; green, 49; blue, 53 }  ,draw opacity=1 ]   (210.8,57.74) -- (210.8,209.02) ;
%Straight Lines [id:da499758515843279] 
\draw [color={rgb, 255:red, 19; green, 117; blue, 183 }  ,draw opacity=1 ]   (203.65,57.74) -- (203.65,209.02) ;
%Straight Lines [id:da9394423626337692] 
\draw [color={rgb, 255:red, 173; green, 138; blue, 81 }  ,draw opacity=1 ]   (217.96,57.74) -- (217.96,209.02) ;
%Straight Lines [id:da7815212486331992] 
\draw [color={rgb, 255:red, 201; green, 49; blue, 53 }  ,draw opacity=1 ]   (210.8,228.75) -- (210.8,294.52) ;
%Straight Lines [id:da5663091834951124] 
\draw [color={rgb, 255:red, 19; green, 117; blue, 183 }  ,draw opacity=1 ]   (203.65,228.75) -- (203.65,294.52) ;
%Straight Lines [id:da14277800777045346] 
\draw [color={rgb, 255:red, 173; green, 138; blue, 81 }  ,draw opacity=1 ]   (217.96,228.75) -- (217.96,294.52) ;
%Curve Lines [id:da32396733785209986] 
\draw [color={rgb, 255:red, 201; green, 49; blue, 53 }  ,draw opacity=1 ]   (164.76,143.25) .. controls (164.41,175.26) and (208.34,176.07) .. (210.71,206.14) ;
\draw [shift={(210.8,209.02)}, rotate = 270.52] [fill={rgb, 255:red, 201; green, 49; blue, 53 }  ,fill opacity=1 ][line width=0.08]  [draw opacity=0] (8.93,-4.29) -- (0,0) -- (8.93,4.29) -- (5.93,0) -- cycle    ;
%Curve Lines [id:da937125369756485] 
\draw [color={rgb, 255:red, 19; green, 117; blue, 183 }  ,draw opacity=1 ]   (157.61,143.25) .. controls (157.72,180.92) and (200.95,182.77) .. (203.53,206.36) ;
\draw [shift={(203.65,209.02)}, rotate = 271.07] [fill={rgb, 255:red, 19; green, 117; blue, 183 }  ,fill opacity=1 ][line width=0.08]  [draw opacity=0] (8.93,-4.29) -- (0,0) -- (8.93,4.29) -- (5.93,0) -- cycle    ;
%Curve Lines [id:da6110281639584693] 
\draw [color={rgb, 255:red, 173; green, 138; blue, 81 }  ,draw opacity=1 ]   (171.92,143.25) .. controls (172.21,169.04) and (215.81,169.68) .. (217.88,206.14) ;
\draw [shift={(217.96,209.02)}, rotate = 270.15999999999997] [fill={rgb, 255:red, 173; green, 138; blue, 81 }  ,fill opacity=1 ][line width=0.08]  [draw opacity=0] (8.93,-4.29) -- (0,0) -- (8.93,4.29) -- (5.93,0) -- cycle    ;
%Curve Lines [id:da284618027973214] 
\draw [color={rgb, 255:red, 201; green, 49; blue, 53 }  ,draw opacity=1 ]   (164.76,228.75) .. controls (164.41,260.77) and (208.34,261.57) .. (210.71,291.65) ;
\draw [shift={(210.8,294.52)}, rotate = 270.52] [fill={rgb, 255:red, 201; green, 49; blue, 53 }  ,fill opacity=1 ][line width=0.08]  [draw opacity=0] (8.93,-4.29) -- (0,0) -- (8.93,4.29) -- (5.93,0) -- cycle    ;
%Curve Lines [id:da4393558348582036] 
\draw [color={rgb, 255:red, 19; green, 117; blue, 183 }  ,draw opacity=1 ]   (157.61,228.75) .. controls (157.72,266.43) and (200.95,268.27) .. (203.53,291.87) ;
\draw [shift={(203.65,294.52)}, rotate = 271.07] [fill={rgb, 255:red, 19; green, 117; blue, 183 }  ,fill opacity=1 ][line width=0.08]  [draw opacity=0] (8.93,-4.29) -- (0,0) -- (8.93,4.29) -- (5.93,0) -- cycle    ;
%Curve Lines [id:da2255961452300932] 
\draw [color={rgb, 255:red, 173; green, 138; blue, 81 }  ,draw opacity=1 ]   (171.92,228.75) .. controls (172.21,254.54) and (215.81,255.18) .. (217.88,291.65) ;
\draw [shift={(217.96,294.52)}, rotate = 270.15999999999997] [fill={rgb, 255:red, 173; green, 138; blue, 81 }  ,fill opacity=1 ][line width=0.08]  [draw opacity=0] (8.93,-4.29) -- (0,0) -- (8.93,4.29) -- (5.93,0) -- cycle    ;
%Curve Lines [id:da9200736108418311] 
\draw [color={rgb, 255:red, 201; green, 49; blue, 53 }  ,draw opacity=1 ]   (210.8,57.74) .. controls (210.44,90.75) and (257.15,90.58) .. (256.85,123.52) ;
%Curve Lines [id:da3641967549980042] 
\draw [color={rgb, 255:red, 19; green, 117; blue, 183 }  ,draw opacity=1 ]   (203.65,57.74) .. controls (203.76,96.78) and (250.18,97.35) .. (249.69,123.52) ;
%Curve Lines [id:da3046795879484804] 
\draw [color={rgb, 255:red, 173; green, 138; blue, 81 }  ,draw opacity=1 ]   (217.96,57.74) .. controls (218.26,84.19) and (264.11,84.19) .. (264,123.52) ;
%Straight Lines [id:da7583148938830677] 
\draw [color={rgb, 255:red, 201; green, 49; blue, 53 }  ,draw opacity=1 ]   (164.76,143.25) -- (164.76,228.75) ;
%Straight Lines [id:da3195747254728495] 
\draw [color={rgb, 255:red, 19; green, 117; blue, 183 }  ,draw opacity=1 ]   (157.61,143.25) -- (157.61,228.75) ;
%Straight Lines [id:da18184924093643895] 
\draw [color={rgb, 255:red, 173; green, 138; blue, 81 }  ,draw opacity=1 ]   (171.92,143.25) -- (171.92,228.75) ;
%Curve Lines [id:da5151292738950305] 
\draw [color={rgb, 255:red, 201; green, 49; blue, 53 }  ,draw opacity=1 ]   (256.85,228.75) .. controls (256.48,261.76) and (211.11,261.59) .. (210.8,294.52) ;
%Curve Lines [id:da31863518610694996] 
\draw [color={rgb, 255:red, 173; green, 138; blue, 81 }  ,draw opacity=1 ]   (264,228.75) .. controls (264.11,267.79) and (218.45,268.36) .. (217.96,294.52) ;
%Curve Lines [id:da2843550876711676] 
\draw [color={rgb, 255:red, 19; green, 117; blue, 183 }  ,draw opacity=1 ]   (249.69,228.75) .. controls (249.94,251.03) and (217.18,254.54) .. (206.78,278.6) .. controls (204.83,283.1) and (203.67,288.32) .. (203.65,294.52) ;
%Straight Lines [id:da12830980928281377] 
\draw [color={rgb, 255:red, 201; green, 49; blue, 53 }  ,draw opacity=1 ]   (256.85,123.52) -- (256.85,228.75) ;
%Straight Lines [id:da014388777602539538] 
\draw [color={rgb, 255:red, 19; green, 117; blue, 183 }  ,draw opacity=1 ]   (249.69,123.52) -- (249.69,228.75) ;
%Straight Lines [id:da1950176729593509] 
\draw [color={rgb, 255:red, 173; green, 138; blue, 81 }  ,draw opacity=1 ]   (264,123.52) -- (264,228.75) ;
%Curve Lines [id:da8459963503859792] 
\draw [color={rgb, 255:red, 128; green, 128; blue, 128 }  ,draw opacity=1 ] [dash pattern={on 4.5pt off 4.5pt}]  (63.55,59.92) .. controls (63.2,91.94) and (20.61,92.9) .. (17.67,122.83) ;
\draw [shift={(17.51,125.7)}, rotate = 270.74] [fill={rgb, 255:red, 128; green, 128; blue, 128 }  ,fill opacity=1 ][line width=0.08]  [draw opacity=0] (8.93,-4.29) -- (0,0) -- (8.93,4.29) -- (5.93,0) -- cycle    ;
%Straight Lines [id:da0813963859111877] 
\draw [color={rgb, 255:red, 128; green, 128; blue, 128 }  ,draw opacity=1 ] [dash pattern={on 4.5pt off 4.5pt}]  (63.55,59.92) -- (63.55,211.2) ;
%Straight Lines [id:da4814910979578162] 
\draw [color={rgb, 255:red, 128; green, 128; blue, 128 }  ,draw opacity=1 ] [dash pattern={on 4.5pt off 4.5pt}]  (63.55,230.93) -- (63.55,296.7) ;
%Curve Lines [id:da027637390822671137] 
\draw [color={rgb, 255:red, 128; green, 128; blue, 128 }  ,draw opacity=1 ] [dash pattern={on 4.5pt off 4.5pt}]  (17.51,145.43) .. controls (17.15,177.44) and (61.09,178.25) .. (63.46,208.32) ;
\draw [shift={(63.55,211.2)}, rotate = 270.52] [fill={rgb, 255:red, 128; green, 128; blue, 128 }  ,fill opacity=1 ][line width=0.08]  [draw opacity=0] (8.93,-4.29) -- (0,0) -- (8.93,4.29) -- (5.93,0) -- cycle    ;
%Curve Lines [id:da4694268582640567] 
\draw [color={rgb, 255:red, 128; green, 128; blue, 128 }  ,draw opacity=1 ] [dash pattern={on 4.5pt off 4.5pt}]  (17.51,230.93) .. controls (17.15,262.95) and (61.09,263.75) .. (63.46,293.83) ;
\draw [shift={(63.55,296.7)}, rotate = 270.52] [fill={rgb, 255:red, 128; green, 128; blue, 128 }  ,fill opacity=1 ][line width=0.08]  [draw opacity=0] (8.93,-4.29) -- (0,0) -- (8.93,4.29) -- (5.93,0) -- cycle    ;
%Curve Lines [id:da3939906267128892] 
\draw [color={rgb, 255:red, 128; green, 128; blue, 128 }  ,draw opacity=1 ] [dash pattern={on 4.5pt off 4.5pt}]  (63.55,59.92) .. controls (63.19,92.93) and (109.9,92.76) .. (109.59,125.7) ;
%Straight Lines [id:da319758157494207] 
\draw [color={rgb, 255:red, 128; green, 128; blue, 128 }  ,draw opacity=1 ] [dash pattern={on 4.5pt off 4.5pt}]  (17.51,145.43) -- (17.51,230.93) ;
%Curve Lines [id:da8864590370998892] 
\draw [color={rgb, 255:red, 128; green, 128; blue, 128 }  ,draw opacity=1 ] [dash pattern={on 4.5pt off 4.5pt}]  (109.59,230.93) .. controls (109.23,263.94) and (63.85,263.77) .. (63.55,296.7) ;
%Straight Lines [id:da46960313946938914] 
\draw [color={rgb, 255:red, 128; green, 128; blue, 128 }  ,draw opacity=1 ] [dash pattern={on 4.5pt off 4.5pt}]  (109.59,125.7) -- (109.59,230.93) ;

%Curve Lines [id:da3216043599927687] 
\draw [color={rgb, 255:red, 19; green, 117; blue, 183 }  ,draw opacity=1 ][line width=2.25]    (516.79,58.55) .. controls (516.44,89.91) and (475.59,91.47) .. (471.14,119.65) ;
\draw [shift={(470.75,124.33)}, rotate = 270.74] [fill={rgb, 255:red, 19; green, 117; blue, 183 }  ,fill opacity=1 ][line width=0.08]  [draw opacity=0] (10.36,-4.98) -- (0,0) -- (10.36,4.98) -- (6.88,0) -- cycle    ;
%Curve Lines [id:da3308199265541284] 
\draw [color={rgb, 255:red, 201; green, 49; blue, 53 }  ,draw opacity=1 ][line width=2.25]    (470.75,144.06) .. controls (470.4,175.41) and (512.54,176.83) .. (516.5,205.13) ;
\draw [shift={(516.79,209.83)}, rotate = 270.52] [fill={rgb, 255:red, 201; green, 49; blue, 53 }  ,fill opacity=1 ][line width=0.08]  [draw opacity=0] (10.36,-4.98) -- (0,0) -- (10.36,4.98) -- (6.88,0) -- cycle    ;
%Curve Lines [id:da03581487614109158] 
\draw [color={rgb, 255:red, 173; green, 138; blue, 81 }  ,draw opacity=1 ][line width=2.25]    (516.79,58.55) .. controls (516.43,91.56) and (563.14,91.39) .. (562.84,124.33) ;
%Straight Lines [id:da8980064767416698] 
\draw [color={rgb, 255:red, 201; green, 49; blue, 53 }  ,draw opacity=1 ][line width=2.25]    (470.75,144.06) -- (470.75,229.56) ;
%Curve Lines [id:da8795005248199839] 
\draw [color={rgb, 255:red, 173; green, 138; blue, 81 }  ,draw opacity=1 ][line width=2.25]    (562.84,229.56) .. controls (562.47,262.57) and (517.1,262.4) .. (516.8,295.33) ;
%Straight Lines [id:da4181801192700343] 
\draw [color={rgb, 255:red, 173; green, 138; blue, 81 }  ,draw opacity=1 ][line width=2.25]    (562.84,124.33) -- (562.84,229.56) ;
%Straight Lines [id:da8042855703109516] 
\draw [color={rgb, 255:red, 19; green, 117; blue, 183 }  ,draw opacity=1 ][line width=2.25]    (516.8,229.56) -- (516.8,295.33) ;
%Straight Lines [id:da4788165717554871] 
\draw [color={rgb, 255:red, 19; green, 117; blue, 183 }  ,draw opacity=1 ][line width=3]    (0,10) -- (50,10) ;

%Straight Lines [id:da40654779418211395] 
\draw [color={rgb, 255:red, 201; green, 49; blue, 53 }  ,draw opacity=1 ][line width=3]    (150,10) -- (200,10) ;

%Straight Lines [id:da41500664475359816] 
\draw [color={rgb, 255:red, 173; green, 138; blue, 81 }  ,draw opacity=1 ][line width=3]    (300,10) -- (350,10) ;

%Curve Lines [id:da6313952649618807] 
\draw [color={rgb, 255:red, 201; green, 49; blue, 53 }  ,draw opacity=1 ][line width=2.25]    (363.55,57.74) .. controls (363.2,89.1) and (322.35,90.67) .. (317.9,118.84) ;
\draw [shift={(317.51,123.52)}, rotate = 270.74] [fill={rgb, 255:red, 201; green, 49; blue, 53 }  ,fill opacity=1 ][line width=0.08]  [draw opacity=0] (10.36,-4.98) -- (0,0) -- (10.36,4.98) -- (6.88,0) -- cycle    ;
%Curve Lines [id:da4058519977174274] 
\draw [color={rgb, 255:red, 19; green, 117; blue, 183 }  ,draw opacity=1 ][line width=1.5]    (356.4,57.74) .. controls (356.69,83.3) and (313.48,84.2) .. (310.52,119.53) ;
\draw [shift={(310.36,123.52)}, rotate = 269.89] [fill={rgb, 255:red, 19; green, 117; blue, 183 }  ,fill opacity=1 ][line width=0.08]  [draw opacity=0] (11.07,-5.32) -- (0,0) -- (11.07,5.32) -- (7.35,0) -- cycle    ;
%Curve Lines [id:da15644361799496564] 
\draw [color={rgb, 255:red, 173; green, 138; blue, 81 }  ,draw opacity=1 ]   (370.71,57.74) .. controls (370.82,96) and (328.12,97.16) .. (324.86,120.85) ;
\draw [shift={(324.67,123.52)}, rotate = 270.65] [fill={rgb, 255:red, 173; green, 138; blue, 81 }  ,fill opacity=1 ][line width=0.08]  [draw opacity=0] (8.93,-4.29) -- (0,0) -- (8.93,4.29) -- (5.93,0) -- cycle    ;
%Straight Lines [id:da4307154139954579] 
\draw [color={rgb, 255:red, 201; green, 49; blue, 53 }  ,draw opacity=1 ]   (363.55,57.74) -- (363.55,209.02) ;
%Straight Lines [id:da20269216915512822] 
\draw [color={rgb, 255:red, 19; green, 117; blue, 183 }  ,draw opacity=1 ]   (356.4,57.74) -- (356.4,209.02) ;
%Straight Lines [id:da5411648472941404] 
\draw [color={rgb, 255:red, 173; green, 138; blue, 81 }  ,draw opacity=1 ][line width=1.5]    (370.71,57.74) -- (370.71,209.02) ;
%Straight Lines [id:da18157506936839585] 
\draw [color={rgb, 255:red, 201; green, 49; blue, 53 }  ,draw opacity=1 ]   (363.55,228.75) -- (363.55,294.52) ;
%Straight Lines [id:da37185302948426857] 
\draw [color={rgb, 255:red, 19; green, 117; blue, 183 }  ,draw opacity=1 ][line width=2.25]    (356.4,228.75) -- (356.4,294.52) ;
%Straight Lines [id:da8612388825529651] 
\draw [color={rgb, 255:red, 173; green, 138; blue, 81 }  ,draw opacity=1 ]   (370.71,228.75) -- (370.71,294.52) ;
%Curve Lines [id:da2250432210588056] 
\draw [color={rgb, 255:red, 201; green, 49; blue, 53 }  ,draw opacity=1 ][line width=2.25]    (317.51,143.25) .. controls (317.16,174.6) and (359.3,176.02) .. (363.26,204.32) ;
\draw [shift={(363.55,209.02)}, rotate = 270.52] [fill={rgb, 255:red, 201; green, 49; blue, 53 }  ,fill opacity=1 ][line width=0.08]  [draw opacity=0] (11.79,-5.66) -- (0,0) -- (11.79,5.66) -- (7.83,0) -- cycle    ;
%Curve Lines [id:da48893143179059917] 
\draw [color={rgb, 255:red, 19; green, 117; blue, 183 }  ,draw opacity=1 ][line width=1.5]    (310.36,143.25) .. controls (310.47,180.34) and (352.36,182.7) .. (356.13,205.28) ;
\draw [shift={(356.4,209.02)}, rotate = 271.07] [fill={rgb, 255:red, 19; green, 117; blue, 183 }  ,fill opacity=1 ][line width=0.08]  [draw opacity=0] (11.07,-5.32) -- (0,0) -- (11.07,5.32) -- (7.35,0) -- cycle    ;
%Curve Lines [id:da8730206543212923] 
\draw [color={rgb, 255:red, 173; green, 138; blue, 81 }  ,draw opacity=1 ]   (324.67,143.25) .. controls (324.96,169.04) and (368.56,169.68) .. (370.63,206.14) ;
\draw [shift={(370.71,209.02)}, rotate = 270.15999999999997] [fill={rgb, 255:red, 173; green, 138; blue, 81 }  ,fill opacity=1 ][line width=0.08]  [draw opacity=0] (8.93,-4.29) -- (0,0) -- (8.93,4.29) -- (5.93,0) -- cycle    ;
%Curve Lines [id:da5122222760865696] 
\draw [color={rgb, 255:red, 201; green, 49; blue, 53 }  ,draw opacity=1 ][line width=2.25]    (317.51,228.75) .. controls (317.16,260.11) and (359.3,261.52) .. (363.26,289.82) ;
\draw [shift={(363.55,294.52)}, rotate = 270.52] [fill={rgb, 255:red, 201; green, 49; blue, 53 }  ,fill opacity=1 ][line width=0.08]  [draw opacity=0] (10.36,-4.98) -- (0,0) -- (10.36,4.98) -- (6.88,0) -- cycle    ;
%Curve Lines [id:da5560534176438023] 
\draw [color={rgb, 255:red, 19; green, 117; blue, 183 }  ,draw opacity=1 ]   (310.36,228.75) .. controls (310.47,266.43) and (353.7,268.27) .. (356.28,291.87) ;
\draw [shift={(356.4,294.52)}, rotate = 271.07] [fill={rgb, 255:red, 19; green, 117; blue, 183 }  ,fill opacity=1 ][line width=0.08]  [draw opacity=0] (8.93,-4.29) -- (0,0) -- (8.93,4.29) -- (5.93,0) -- cycle    ;
%Curve Lines [id:da16318752485276955] 
\draw [color={rgb, 255:red, 173; green, 138; blue, 81 }  ,draw opacity=1 ][line width=1.5]    (324.67,228.75) .. controls (324.96,254.28) and (367.67,255.17) .. (370.56,290.54) ;
\draw [shift={(370.71,294.52)}, rotate = 270.15999999999997] [fill={rgb, 255:red, 173; green, 138; blue, 81 }  ,fill opacity=1 ][line width=0.08]  [draw opacity=0] (11.07,-5.32) -- (0,0) -- (11.07,5.32) -- (7.35,0) -- cycle    ;
%Curve Lines [id:da750161894814765] 
\draw [color={rgb, 255:red, 201; green, 49; blue, 53 }  ,draw opacity=1 ]   (363.55,57.74) .. controls (363.19,90.75) and (409.9,90.58) .. (409.59,123.52) ;
%Curve Lines [id:da903385876924106] 
\draw [color={rgb, 255:red, 19; green, 117; blue, 183 }  ,draw opacity=1 ][line width=1.5]    (356.4,57.74) .. controls (356.51,96.78) and (402.93,97.35) .. (402.44,123.52) ;
%Curve Lines [id:da259079691759043] 
\draw [color={rgb, 255:red, 173; green, 138; blue, 81 }  ,draw opacity=1 ][line width=2.25]    (370.71,57.74) .. controls (371.01,84.19) and (416.86,84.19) .. (416.75,123.52) ;
%Straight Lines [id:da36433058333595125] 
\draw [color={rgb, 255:red, 201; green, 49; blue, 53 }  ,draw opacity=1 ][line width=2.25]    (317.51,143.25) -- (317.51,228.75) ;
%Straight Lines [id:da200544569302313] 
\draw [color={rgb, 255:red, 19; green, 117; blue, 183 }  ,draw opacity=1 ]   (310.36,143.25) -- (310.36,228.75) ;
%Straight Lines [id:da269880087422386] 
\draw [color={rgb, 255:red, 173; green, 138; blue, 81 }  ,draw opacity=1 ][line width=1.5]    (324.67,143.25) -- (324.67,228.75) ;
%Curve Lines [id:da7315771577680499] 
\draw [color={rgb, 255:red, 201; green, 49; blue, 53 }  ,draw opacity=1 ]   (409.59,228.75) .. controls (409.23,261.76) and (363.85,261.59) .. (363.55,294.52) ;
%Curve Lines [id:da6763038278572308] 
\draw [color={rgb, 255:red, 173; green, 138; blue, 81 }  ,draw opacity=1 ][line width=2.25]    (416.75,228.75) .. controls (416.86,267.79) and (371.2,268.36) .. (370.71,294.52) ;
%Curve Lines [id:da9469967856757917] 
\draw [color={rgb, 255:red, 19; green, 117; blue, 183 }  ,draw opacity=1 ][line width=1.5]    (402.44,228.75) .. controls (402.69,251.03) and (369.93,254.54) .. (359.53,278.6) .. controls (357.58,283.1) and (356.42,288.32) .. (356.4,294.52) ;
%Straight Lines [id:da4486412692251923] 
\draw [color={rgb, 255:red, 201; green, 49; blue, 53 }  ,draw opacity=1 ]   (409.59,123.52) -- (409.59,228.75) ;
%Straight Lines [id:da8245570969828782] 
\draw [color={rgb, 255:red, 19; green, 117; blue, 183 }  ,draw opacity=1 ][line width=1.5]    (402.44,123.52) -- (402.44,228.75) ;
%Straight Lines [id:da14060830640344013] 
\draw [color={rgb, 255:red, 173; green, 138; blue, 81 }  ,draw opacity=1 ][line width=2.25]    (416.75,123.52) -- (416.75,228.75) ;
%Curve Lines [id:da19147896727954006] 
\draw [color={rgb, 255:red, 201; green, 49; blue, 53 }  ,draw opacity=1 ][line width=2.25]    (470.75,229.56) .. controls (470.4,260.92) and (512.54,262.33) .. (516.5,290.63) ;
\draw [shift={(516.79,295.33)}, rotate = 270.52] [fill={rgb, 255:red, 201; green, 49; blue, 53 }  ,fill opacity=1 ][line width=0.08]  [draw opacity=0] (10.36,-4.98) -- (0,0) -- (10.36,4.98) -- (6.88,0) -- cycle    ;

% Text Node
\draw  [fill={rgb, 255:red, 238; green, 238; blue, 238 }  ,fill opacity=1 ]  (495.31,38.34) -- (536.31,38.34) -- (536.31,60.34) -- (495.31,60.34) -- cycle  ;
\draw (515.81,49.34) node   [align=left] {\begin{minipage}[lt]{25.4933357671259pt}\setlength\topsep{0pt}
\begin{center}
0
\end{center}

\end{minipage}};
% Text Node
\draw  [fill={rgb, 255:red, 238; green, 238; blue, 238 }  ,fill opacity=1 ]  (451.24,123.85) -- (492.24,123.85) -- (492.24,145.85) -- (451.24,145.85) -- cycle  ;
\draw (471.74,134.85) node   [align=left] {\begin{minipage}[lt]{25.49333576712586pt}\setlength\topsep{0pt}
\begin{center}
1
\end{center}

\end{minipage}};
% Text Node
\draw  [fill={rgb, 255:red, 238; green, 238; blue, 238 }  ,fill opacity=1 ]  (495.31,208.04) -- (536.31,208.04) -- (536.31,230.04) -- (495.31,230.04) -- cycle  ;
\draw (515.81,219.04) node   [align=left] {\begin{minipage}[lt]{25.4933357671259pt}\setlength\topsep{0pt}
\begin{center}
2\\
\end{center}

\end{minipage}};
% Text Node
\draw  [fill={rgb, 255:red, 238; green, 238; blue, 238 }  ,fill opacity=1 ]  (495.31,293.54) -- (536.31,293.54) -- (536.31,315.54) -- (495.31,315.54) -- cycle  ;
\draw (515.81,304.54) node   [align=left] {\begin{minipage}[lt]{25.4933357671259pt}\setlength\topsep{0pt}
\begin{center}
3\\
\end{center}

\end{minipage}};
% Text Node
\draw  [fill={rgb, 255:red, 238; green, 238; blue, 238 }  ,fill opacity=1 ]  (42.07,39.72) -- (83.07,39.72) -- (83.07,61.72) -- (42.07,61.72) -- cycle  ;
\draw (62.57,50.72) node   [align=left] {\begin{minipage}[lt]{25.49333576712591pt}\setlength\topsep{0pt}
\begin{center}
0
\end{center}

\end{minipage}};
% Text Node
\draw  [fill={rgb, 255:red, 238; green, 238; blue, 238 }  ,fill opacity=1 ]  (-2,125.22) -- (39,125.22) -- (39,147.22) -- (-2,147.22) -- cycle  ;
\draw (18.5,136.22) node   [align=left] {\begin{minipage}[lt]{25.49333576712589pt}\setlength\topsep{0pt}
\begin{center}
1
\end{center}

\end{minipage}};
% Text Node
\draw  [fill={rgb, 255:red, 238; green, 238; blue, 238 }  ,fill opacity=1 ]  (42.07,209.41) -- (83.07,209.41) -- (83.07,231.41) -- (42.07,231.41) -- cycle  ;
\draw (62.57,220.41) node   [align=left] {\begin{minipage}[lt]{25.49333576712591pt}\setlength\topsep{0pt}
\begin{center}
2\\
\end{center}

\end{minipage}};
% Text Node
\draw  [fill={rgb, 255:red, 238; green, 238; blue, 238 }  ,fill opacity=1 ]  (42.07,294.91) -- (83.07,294.91) -- (83.07,316.91) -- (42.07,316.91) -- cycle  ;
\draw (62.57,305.91) node   [align=left] {\begin{minipage}[lt]{25.49333576712591pt}\setlength\topsep{0pt}
\begin{center}
3\\
\end{center}

\end{minipage}};
% Text Node
\draw  [fill={rgb, 255:red, 238; green, 238; blue, 238 }  ,fill opacity=1 ]  (189.32,37.54) -- (230.32,37.54) -- (230.32,59.54) -- (189.32,59.54) -- cycle  ;
\draw (209.82,48.54) node   [align=left] {\begin{minipage}[lt]{25.493335767125938pt}\setlength\topsep{0pt}
\begin{center}
0
\end{center}

\end{minipage}};
% Text Node
\draw  [fill={rgb, 255:red, 238; green, 238; blue, 238 }  ,fill opacity=1 ]  (145.25,123.04) -- (186.25,123.04) -- (186.25,145.04) -- (145.25,145.04) -- cycle  ;
\draw (165.75,134.04) node   [align=left] {\begin{minipage}[lt]{25.493335767125938pt}\setlength\topsep{0pt}
\begin{center}
1
\end{center}

\end{minipage}};
% Text Node
\draw  [fill={rgb, 255:red, 238; green, 238; blue, 238 }  ,fill opacity=1 ]  (189.32,207.23) -- (230.32,207.23) -- (230.32,229.23) -- (189.32,229.23) -- cycle  ;
\draw (209.82,218.23) node   [align=left] {\begin{minipage}[lt]{25.493335767125938pt}\setlength\topsep{0pt}
\begin{center}
2\\
\end{center}

\end{minipage}};
% Text Node
\draw  [fill={rgb, 255:red, 238; green, 238; blue, 238 }  ,fill opacity=1 ]  (189.32,292.73) -- (230.32,292.73) -- (230.32,314.73) -- (189.32,314.73) -- cycle  ;
\draw (209.82,303.73) node   [align=left] {\begin{minipage}[lt]{25.493335767125938pt}\setlength\topsep{0pt}
\begin{center}
3\\
\end{center}

\end{minipage}};
% Text Node
\draw (55.92,331) node [anchor=north west][inner sep=0.75pt]   [align=left] {a)};
% Text Node
\draw (201.6,331) node [anchor=north west][inner sep=0.75pt]   [align=left] {b)};
% Text Node
\draw (355.48,331) node [anchor=north west][inner sep=0.75pt]   [align=left] {c)};
% Text Node
\draw (509.16,331) node [anchor=north west][inner sep=0.75pt]   [align=left] {d)};
% Text Node
\draw (64,2) node [anchor=north west][inner sep=0.75pt]   [align=left] {Conv 3x3};
% Text Node
\draw (214,2) node [anchor=north west][inner sep=0.75pt]   [align=left] {Conv 5x5};
% Text Node
\draw (364,2) node [anchor=north west][inner sep=0.75pt]   [align=left] {MaxPool};
% Text Node
\draw  [fill={rgb, 255:red, 238; green, 238; blue, 238 }  ,fill opacity=1 ]  (342.07,37.54) -- (383.07,37.54) -- (383.07,59.54) -- (342.07,59.54) -- cycle  ;
\draw (362.57,48.54) node   [align=left] {\begin{minipage}[lt]{25.493335767125938pt}\setlength\topsep{0pt}
\begin{center}
0
\end{center}

\end{minipage}};
% Text Node
\draw  [fill={rgb, 255:red, 238; green, 238; blue, 238 }  ,fill opacity=1 ]  (298,123.04) -- (339,123.04) -- (339,145.04) -- (298,145.04) -- cycle  ;
\draw (318.5,134.04) node   [align=left] {\begin{minipage}[lt]{25.493335767125938pt}\setlength\topsep{0pt}
\begin{center}
1
\end{center}

\end{minipage}};
% Text Node
\draw  [fill={rgb, 255:red, 238; green, 238; blue, 238 }  ,fill opacity=1 ]  (342.07,207.23) -- (383.07,207.23) -- (383.07,229.23) -- (342.07,229.23) -- cycle  ;
\draw (362.57,218.23) node   [align=left] {\begin{minipage}[lt]{25.493335767125938pt}\setlength\topsep{0pt}
\begin{center}
2\\
\end{center}

\end{minipage}};
% Text Node
\draw  [fill={rgb, 255:red, 238; green, 238; blue, 238 }  ,fill opacity=1 ]  (342.07,292.73) -- (383.07,292.73) -- (383.07,314.73) -- (342.07,314.73) -- cycle  ;
\draw (362.57,303.73) node   [align=left] {\begin{minipage}[lt]{25.493335767125938pt}\setlength\topsep{0pt}
\begin{center}
3\\
\end{center}

\end{minipage}};

\end{tikzpicture}

%% file: fig/darts_vs_pdarts.tex
\tikzset{every picture/.style={line width=0.75pt}} %set default line width to 0.75pt        

\begin{tikzpicture}[x=0.75pt,y=0.75pt,yscale=-0.8,xscale=1]

%Shape: Rectangle [id:dp7756819472532681] 
\draw   (34.7,73.26) -- (79.71,73.26) -- (79.71,144.28) -- (34.7,144.28) -- cycle ;

%Shape: Rectangle [id:dp19409265827711064] 
\draw  [fill={rgb, 255:red, 221; green, 221; blue, 221 }  ,fill opacity=1 ] (132.97,20) -- (177.36,20) -- (177.36,197.54) -- (132.97,197.54) -- cycle ;

%Straight Lines [id:da7772867690925238] 
\draw    (79.71,108.77) -- (130.97,108.77) ;
\draw [shift={(132.97,108.77)}, rotate = 180] [color={rgb, 255:red, 0; green, 0; blue, 0 }  ][line width=0.75]    (10.93,-3.29) .. controls (6.95,-1.4) and (3.31,-0.3) .. (0,0) .. controls (3.31,0.3) and (6.95,1.4) .. (10.93,3.29)   ;
%Shape: Rectangle [id:dp07339942168625258] 
\draw   (247.75,73.26) -- (292.75,73.26) -- (292.75,144.28) -- (247.75,144.28) -- cycle ;

%Shape: Rectangle [id:dp1272363232771706] 
\draw  [fill={rgb, 255:red, 221; green, 221; blue, 221 }  ,fill opacity=1 ] (470.29,20) -- (514.67,20) -- (514.67,197.54) -- (470.29,197.54) -- cycle ;

%Straight Lines [id:da35951016975361905] 
\draw    (417.03,108.77) -- (468.29,108.77) ;
\draw [shift={(470.29,108.77)}, rotate = 180] [color={rgb, 255:red, 0; green, 0; blue, 0 }  ][line width=0.75]    (10.93,-3.29) .. controls (6.95,-1.4) and (3.31,-0.3) .. (0,0) .. controls (3.31,0.3) and (6.95,1.4) .. (10.93,3.29)   ;
%Shape: Rectangle [id:dp17423621609080686] 
\draw   (309.89,55.51) -- (354.89,55.51) -- (354.89,162.03) -- (309.89,162.03) -- cycle ;
%Shape: Rectangle [id:dp2101838416619941] 
\draw   (370,37.97) -- (415.01,37.97) -- (415.01,180) -- (370,180) -- cycle ;
%Straight Lines [id:da5379311887597568] 
\draw    (354.89,108.77) -- (370.64,108.77) ;
\draw [shift={(372.64,108.77)}, rotate = 180] [color={rgb, 255:red, 0; green, 0; blue, 0 }  ][line width=0.75]    (10.93,-3.29) .. controls (6.95,-1.4) and (3.31,-0.3) .. (0,0) .. controls (3.31,0.3) and (6.95,1.4) .. (10.93,3.29)   ;
%Straight Lines [id:da7831344714512023] 
\draw    (292.75,108.77) -- (308.51,108.77) ;
\draw [shift={(310.51,108.77)}, rotate = 180] [color={rgb, 255:red, 0; green, 0; blue, 0 }  ][line width=0.75]    (10.93,-3.29) .. controls (6.95,-1.4) and (3.31,-0.3) .. (0,0) .. controls (3.31,0.3) and (6.95,1.4) .. (10.93,3.29)   ;

% Text Node
\draw (43.63,95.82) node [anchor=north west][inner sep=0.75pt]  [font=\footnotesize] [align=left] {\begin{minipage}[lt]{19.584000000000003pt}\setlength\topsep{0pt}
\begin{center}
{\small 8}\\{\small Cells}
\end{center}

\end{minipage}};
% Text Node
\draw (141.28,100.26) node [anchor=north west][inner sep=0.75pt]  [font=\footnotesize] [align=left] {\begin{minipage}[lt]{19.584000000000003pt}\setlength\topsep{0pt}
\begin{center}
{\small \textcolor[rgb]{0,0,0}{20}}\\{\small \textcolor[rgb]{0,0,0}{Cells}}
\end{center}

\end{minipage}};
% Text Node
\draw (478.59,100.26) node [anchor=north west][inner sep=0.75pt]  [font=\footnotesize] [align=left] {\begin{minipage}[lt]{19.584000000000003pt}\setlength\topsep{0pt}
\begin{center}
{\small \textcolor[rgb]{0,0,0}{20}}\\{\small \textcolor[rgb]{0,0,0}{Cells}}
\end{center}

\end{minipage}};
% Text Node
\draw (256.67,95.82) node [anchor=north west][inner sep=0.75pt]  [font=\footnotesize] [align=left] {\begin{minipage}[lt]{19.584000000000003pt}\setlength\topsep{0pt}
\begin{center}
{\small 8}\\{\small Cells}
\end{center}

\end{minipage}};
% Text Node
\draw (318.81,95.88) node [anchor=north west][inner sep=0.75pt]  [font=\footnotesize] [align=left] {\begin{minipage}[lt]{19.584000000000003pt}\setlength\topsep{0pt}
\begin{center}
11\\{\small Cells}
\end{center}

\end{minipage}};
% Text Node
\draw (380.95,95.88) node [anchor=north west][inner sep=0.75pt]  [font=\footnotesize] [align=left] {\begin{minipage}[lt]{19.584000000000003pt}\setlength\topsep{0pt}
\begin{center}
17\\{\small Cells}
\end{center}

\end{minipage}};
% Text Node
\draw (26.79,207.35) node [anchor=north west][inner sep=0.75pt]  [font=\small] [align=left] {Search Net.};
% Text Node
\draw (123.39,207.35) node [anchor=north west][inner sep=0.75pt]  [font=\small] [align=left] {Eval. Net.};
% Text Node
\draw (301.97,207.35) node [anchor=north west][inner sep=0.75pt]  [font=\small] [align=left] {Search Net.};
% Text Node
\draw (462.48,207.35) node [anchor=north west][inner sep=0.75pt]  [font=\small] [align=left] {Eval. Net.};
% Text Node
\draw (69.57,242.74) node [anchor=north west][inner sep=0.75pt]   [align=left] {a) DARTS};
% Text Node
\draw (363.43,241.85) node [anchor=north west][inner sep=0.75pt]   [align=left] {b) P-DARTS};

\end{tikzpicture}

%% file: chapters/01b_related_work.tex
 To obtain more flexible architectures and further automate NAS, there has been recent work on combining meta-learning with NAS. This approach requires fewer resources than searching architectures from scratch and provides performance improvements compared to the simple architecture transfer. 
 
 \cite{wong2019transfer} proposed Transfer Neural AutoML, where a reinforcement learning algorithm learns over multiple tasks in parallel and then transfers the search strategy to a new task. Authors transfer the parameters of a pre-trained controller and add a new randomly initialized embedding for the new task. Their approach is able to significantly reduce convergence time by speeding up the hyper-parameters and architecture search on a new task.

\cite{fang2019eatnas} proposed an elastic architecture transfer mechanism (EAT-NAS), where a \textit{seed architecture} is first searched on a small-scale task and later used as a seed for the search on a large-scale task. EAT-NAS relies on an evolutionary algorithm and its population-based search process. The best architecture in the population of small-scale task search is selected as the seed architecture. In the second step, the seed architecture is used to initialize a search population for the large-scale search by obtaining the new architectures by adding perturbations to the seed architecture. This approach enables faster evolution of searched architectures on the target dataset.

While the above approaches successfully presented mechanisms that allow relatively computationally cheap adaptation of transfer knowledge for a new task, it is based on the reinforcement learning algorithm and the evolutionary algorithm respectively. In the field of gradient descent-based NAS, \cite{Lian2020TowardsFA} tackled the challenge of NAS for multiple tasks in a few-shot and supervised environment by using DARTS as the NAS method. The authors propose Transferable Neural Architecture Search (T-NAS) based on MAML \citep{maml} and DARTS, where T-NAS learns a meta-architecture that can be adapted to a new task through only a few gradient steps. T-NAS achieved state-of-the-art performance while greatly reducing required search costs. 

\cite{elsken2020metalearning} proposed the MetaNAS approach for few-shot learning, where an arbitrary gradient-based  NAS  method is fully integrated with arbitrary model-agnostic meta-learning algorithms. The authors demonstrated their approach by a combination of DARTS and REPTILE \citep{nichol2018firstorder} to obtain state-of-the-art results. Furthermore, the authors also proposed an extension that reduces the performance drop incurred during hard-pruning. Compared to this work, where we used P-DARTS to tackle the DARTS' performance drop and Task2Vec as a backbone of our meta-framework.

The research presented above is primarily concentrated on a few-shot learning environment. \cite{houben2019msc} presented a similar framework to ours to warm-start DARTS, where DARTS is warm-started based on a task similarity measure. However, his concept to warm-start DARTS, while only theoretically presented, is different from our work as DARTS would be used to find additional layers on top of a transferred architecture --- it should not be confused with the transfer of seed architectures --- instead of warm starting the search process. For a task similarity measure author presented a novel approach that uses FaceNet \cite{schroff2015facenet}, a facial recognition algorithm.

%% file: chapters/02_methodology.tex
\begin{figure}[h]
    \centering
     \includegraphics[width=\linewidth]{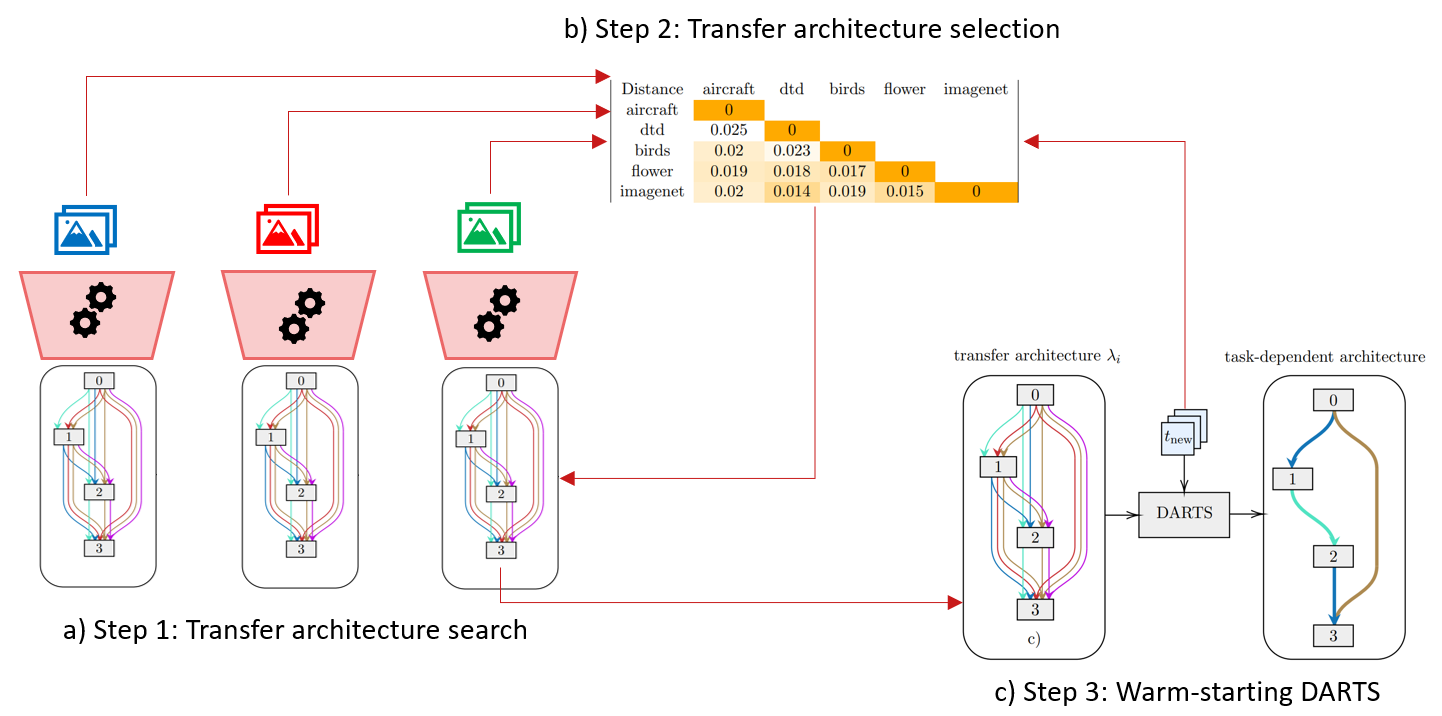}
    \caption{An overview of the whole pipeline, where step 1 and partially step 2 are done offline: a) In the first step, we generate transfer architectures by \textit{transfer architecture search} on a range of different tasks. Ideally, we want to create transfer architectures from a wide range of different tasks as we do not know about the novel task at this point. b) In step two, we compute a task similarity matrix to get the information about relationships between proxy tasks (done offline) and our novel tasks, $t_{new}$ (done live). c) Finally, we select the transfer architecture found on the most similar task and use it to warm-start DARTS algorithm on our $t_{new}$.}
    \label{fig:pipeline-v2}
\end{figure}

Our goal is to learn a \textit{transferable architecture} that can be quickly and easily adapted to a novel task $ t_\text{new}$. We propose a meta-learning framework that leverages P-DARTS \citep{chen2019pdarts} as our backbone component to compute a transfer architecture that can be used to warm-start DARTS for $t_\text{new}$. The proposed framework is made up of three components: \textit{transfer architecture search}, \textit{transfer architecture selection}, and \textit{warm-starting DARTS} that can be observed in Figure \ref{fig:pipeline-v2}.

\subsection{Transfer Architecture Search}

Transfer Architecture Search (TAS) is used to find transferable architectures from a set of predefined tasks. TAS follows the approach of P-DARTS \citep{chen2019pdarts}, where the search space approximation scheme (section \ref{search-space-approximation}) reduces the size of the operation space, $O_k$, while increasing the number of stacked cells, $L_k $, in a multi-stage search process. At the initial step, $\mathfrak{S}_1$, the algorithm starts with a relatively shallow network with a single large DAG representing the complete search space. After each stage, $\mathfrak{S}_{k-1}$, the depth of the network is increased by stacking more cells together and the operation space is approximated by dropping candidate operations with lower architecture parameters, $\alpha_{k-1}$, learned during the previous stage.  In the final stage, our approach diverges from the approach of P-DARTS. While P-DARTS determines the final topology on the last stage according to the standard DARTS algorithm, we discard this step as we are only interested in the discovered DAG together with learned network weights $\hat{w}$ and learned architecture parameters $\hat{\alpha}$. If the TAS is done over a single task $t_i$, we call the discovered DAG \textit{transfer architecture}, $\lambda_i$. To obtain the \textit{trained transfer architecture}, $\hat{\lambda_i}$, corresponding $\hat{w}$ and $\hat{\alpha}$ are jointly transferred with $\lambda_i$. 

Our scheme can also be used in a meta-learning environment. If the TAS is employed over multiple tasks, $t_i \in \mathcal{T}_\text{meta}$, it learns only one architecture by sharing network weights, $w_k$, and architecture parameters, $\alpha_k$, on every stage, $\mathfrak{S}_k$. We define this architecture as \textit{meta-transfer architecture}, $\lambda_\text{meta}$.

\begin{figure}[h]
    \centering
    \input{fig/tas}
    \caption{Illustration of a single step of TAS: a) Architecture discovered in the previous stage, $\mathfrak{S}_{k-1}$, is used as an input. If the current stage is the initial stage, $\mathfrak{S}_{1}$, a single large DAG representing the complete search space is used. b) Architecture parameters, $\alpha_k$, are learned following the approach described in P-DARTS \citep{liu2019darts}. c) Candidate operations with the lowest architecture, $\alpha_k$, for this step are dropped. An important aspect of TAS is that we never discover a single discrete architecture. We can imagine it as a DARTS or P-DARTS without the discretization of the architecture. }
    \label{fig:tas-illustration}
\end{figure}
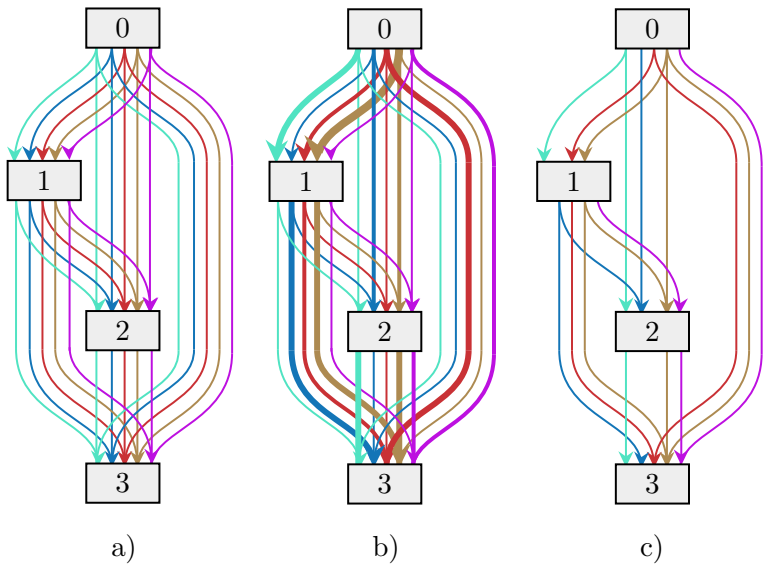

\subsection{Transfer architecture selection} 

Selecting the right transfer architecture for $t_\text{new}$ poses a similar challenge as defining restrictive search space for NAS. This carries the risk of bias, which can prevent NAS from finding a novel, better performing, architecture. In essence, the transfer architecture is defining the search space of P-DARTS. To address this challenge, we added a task similarity component to our framework, which follows the intuition that NAS will find similar architectures on similar tasks and by reducing the search space we are less likely to negatively affect performance.

For every $t_i \in \mathcal{T}$ we compute a transfer architecture, $\lambda_i \in \Lambda$, using TAS. To select the most suitable architecture for $t_{new}$, as mentioned before, a task similarity measure is employed. Let $t^*_i$ denote the most similar task to $t_\text{new}$ and its transfer architecture, $\lambda^*_i$, and learned transfer architecture, $\hat{\lambda}^*_i$.

The task similarity measure can be chosen freely as the framework does not impose any special restrictions or requirements. However, for the purpose of this work, we used Task2Vec by \cite{achille2019task2vec} to determine similarity. Other methods to determine task similarity \citep{kim2017similarity,zamir2018taskonomy,alvarezmelis2020geometric} were also studied. In the end, Task2Vec was selected as it allows us to compute similarity distances between tasks independently from the number of classes and class label semantics. While it may not be the most optimal method, the answer to the question if the selected method would yield to the same conclusion is outside of the scope of this work. Further research would be needed to evaluate the performance of different task similarity methods.

In the case of using a meta-transfer architecture, we want to find a search space that would generalize well over the new task $t_{new}$. The datasets that represent the meta-dataset can be selected by clustering $k$ most similar datasets together. The TA obtained from the closest cluster to the $t_new$ would be used to warm-start DARTS. However, in this work, we are using datasets from the meta-test group. Please refer to the Sections \ref{search-space} and \ref{sec:e2} for more information.

\subsection{Warm-started DARTS}  
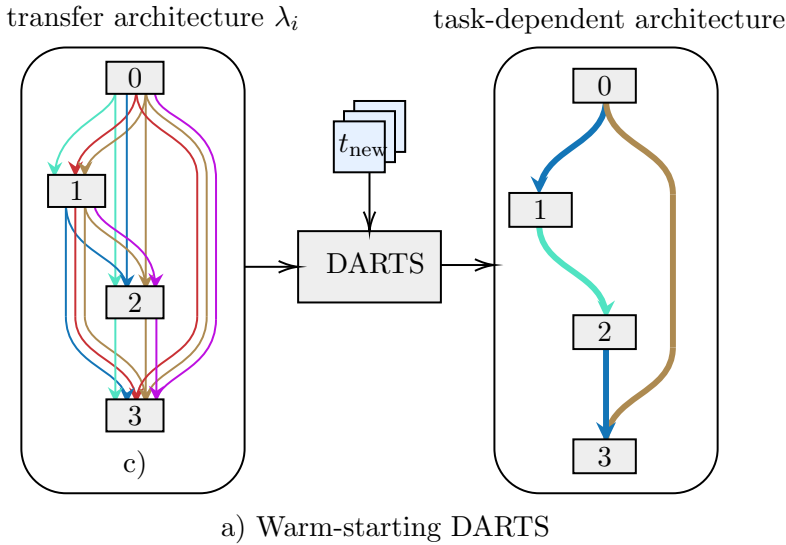
\begin{figure}[ht]
    \centering
    \input{fig/ws-darts}
    \caption{DARTS algorithm is used to compute a task-dependent architecture from a transfer architecture for some novel task. Transfer architecture, $\lambda_i$ is first used as a seed architecture to reduce the search space for novel task, $t_new$. Next, DARTS is run on $t_new$ to compute the final architecture. }
    \label{fig:pipeline}
\end{figure}

Warm-started DARTS (WS-DARTS) uses a transfer architecture, $\lambda^*_i$ (alternatively learned, $\hat{\lambda}^*_i$, or meta, $\lambda_\text{meta}$, transfer architecture), as a seed to start the search for $t_\text{new}$. Instead of running the DARTS algorithm from scratch using the initial operation space, a transfer architecture, $\lambda^*_i$, is used to define the cell search space and thus reducing the computational complexity. The whole pipeline can be observed in the Figure \ref{fig:pipeline}.

In the case of a trained transfer architecture, $\hat{\lambda}_i$, we warm-start the search not only with $\lambda^*_i$ but by additionally transferring weight initialization. This is achieved by removing the last layer in our model and freshly re-initializing it in the new task. Training is then started with the transferred weights, where no layers are frozen --- allowing for so-called end-to-end tuning. The main advantage of this approach allows for higher training and testing accuracy from the start, which could help to find better architectures. Furthermore, because we are also transferring the learned $\alpha$'s, DARTS will start the search process already directed to a better performing architecture. Intuitively, this could help tasks containing smaller amounts of images to find better architectures. 

Since P-DARTS is built on top of DARTS, we take advantage of this fact and use P-DARTS to do warm-starting. Since we discover the final architecture in a single, final, step, we in essence run a DARTS. However, the search space regularization scheme (section \ref{search-space-regularization}) is only partially used, where operation-level Dropout \citep{Srivastava2014dropout} is added to our DARTS search. With the exception of operation-level Dropout, the search follows the same approach as the authors of DARTS \citep{liu2019darts}.

%% file: fig/tas.tex
\tikzset{every picture/.style={line width=0.75pt}} %set default line width to 0.75pt        

\begin{tikzpicture}[x=0.75pt,y=0.75pt,yscale=-0.9,xscale=0.9]
%uncomment if require: \path (0,363); %set diagram left start at 0, and has height of 363

%Straight Lines [id:da1646741341996304] 
\draw [color={rgb, 255:red, 80; green, 227; blue, 194 }  ,draw opacity=1 ]   (47.75,19.19) -- (47.75,170.46) ;
%Curve Lines [id:da5724254698111918] 
\draw [color={rgb, 255:red, 201; green, 49; blue, 53 }  ,draw opacity=1 ]   (63.55,18.21) .. controls (63.2,50.22) and (20.61,51.18) .. (17.67,81.12) ;
\draw [shift={(17.51,83.98)}, rotate = 270.74] [fill={rgb, 255:red, 201; green, 49; blue, 53 }  ,fill opacity=1 ][line width=0.08]  [draw opacity=0] (8.93,-4.29) -- (0,0) -- (8.93,4.29) -- (5.93,0) -- cycle    ;
%Curve Lines [id:da25261550577595016] 
\draw [color={rgb, 255:red, 19; green, 117; blue, 183 }  ,draw opacity=1 ]   (56.4,18.21) .. controls (56.69,44.03) and (12.58,44.68) .. (10.44,81.11) ;
\draw [shift={(10.36,83.98)}, rotate = 269.89] [fill={rgb, 255:red, 19; green, 117; blue, 183 }  ,fill opacity=1 ][line width=0.08]  [draw opacity=0] (8.93,-4.29) -- (0,0) -- (8.93,4.29) -- (5.93,0) -- cycle    ;
%Curve Lines [id:da7996813815685391] 
\draw [color={rgb, 255:red, 173; green, 138; blue, 81 }  ,draw opacity=1 ]   (70.71,18.21) .. controls (70.82,56.46) and (28.12,57.63) .. (24.86,81.31) ;
\draw [shift={(24.67,83.98)}, rotate = 270.65] [fill={rgb, 255:red, 173; green, 138; blue, 81 }  ,fill opacity=1 ][line width=0.08]  [draw opacity=0] (8.93,-4.29) -- (0,0) -- (8.93,4.29) -- (5.93,0) -- cycle    ;
%Straight Lines [id:da09471848366401414] 
\draw [color={rgb, 255:red, 201; green, 49; blue, 53 }  ,draw opacity=1 ]   (63.55,18.21) -- (63.55,169.48) ;
%Straight Lines [id:da7126287112468628] 
\draw [color={rgb, 255:red, 19; green, 117; blue, 183 }  ,draw opacity=1 ]   (56.4,18.21) -- (56.4,169.48) ;
%Straight Lines [id:da8406331336295593] 
\draw [color={rgb, 255:red, 173; green, 138; blue, 81 }  ,draw opacity=1 ]   (70.71,17.21) -- (70.71,168.48) ;
%Straight Lines [id:da03790983856161434] 
\draw [color={rgb, 255:red, 201; green, 49; blue, 53 }  ,draw opacity=1 ]   (63.55,189.22) -- (63.55,254.99) ;
%Straight Lines [id:da025111173404144704] 
\draw [color={rgb, 255:red, 19; green, 117; blue, 183 }  ,draw opacity=1 ]   (56.4,189.22) -- (56.4,254.99) ;
%Straight Lines [id:da1449738292194649] 
\draw [color={rgb, 255:red, 173; green, 138; blue, 81 }  ,draw opacity=1 ]   (70.71,189.22) -- (70.71,254.99) ;
%Curve Lines [id:da2940742389430263] 
\draw [color={rgb, 255:red, 201; green, 49; blue, 53 }  ,draw opacity=1 ]   (17.51,103.71) .. controls (17.15,135.73) and (61.09,136.53) .. (63.46,166.61) ;
\draw [shift={(63.55,169.48)}, rotate = 270.52] [fill={rgb, 255:red, 201; green, 49; blue, 53 }  ,fill opacity=1 ][line width=0.08]  [draw opacity=0] (8.93,-4.29) -- (0,0) -- (8.93,4.29) -- (5.93,0) -- cycle    ;
%Curve Lines [id:da7754960612845854] 
\draw [color={rgb, 255:red, 19; green, 117; blue, 183 }  ,draw opacity=1 ]   (10.36,103.71) .. controls (10.47,141.39) and (53.7,143.23) .. (56.28,166.83) ;
\draw [shift={(56.4,169.48)}, rotate = 271.07] [fill={rgb, 255:red, 19; green, 117; blue, 183 }  ,fill opacity=1 ][line width=0.08]  [draw opacity=0] (8.93,-4.29) -- (0,0) -- (8.93,4.29) -- (5.93,0) -- cycle    ;
%Curve Lines [id:da0019346765664887133] 
\draw [color={rgb, 255:red, 173; green, 138; blue, 81 }  ,draw opacity=1 ]   (24.67,103.71) .. controls (24.96,129.5) and (68.56,130.15) .. (70.63,166.61) ;
\draw [shift={(70.71,169.48)}, rotate = 270.15999999999997] [fill={rgb, 255:red, 173; green, 138; blue, 81 }  ,fill opacity=1 ][line width=0.08]  [draw opacity=0] (8.93,-4.29) -- (0,0) -- (8.93,4.29) -- (5.93,0) -- cycle    ;
%Curve Lines [id:da7660907431977544] 
\draw [color={rgb, 255:red, 201; green, 49; blue, 53 }  ,draw opacity=1 ]   (17.51,189.22) .. controls (17.15,221.23) and (61.09,222.04) .. (63.46,252.11) ;
\draw [shift={(63.55,254.99)}, rotate = 270.52] [fill={rgb, 255:red, 201; green, 49; blue, 53 }  ,fill opacity=1 ][line width=0.08]  [draw opacity=0] (8.93,-4.29) -- (0,0) -- (8.93,4.29) -- (5.93,0) -- cycle    ;
%Curve Lines [id:da020812277807150226] 
\draw [color={rgb, 255:red, 19; green, 117; blue, 183 }  ,draw opacity=1 ]   (10.36,189.22) .. controls (10.47,226.89) and (53.7,228.73) .. (56.28,252.33) ;
\draw [shift={(56.4,254.99)}, rotate = 271.07] [fill={rgb, 255:red, 19; green, 117; blue, 183 }  ,fill opacity=1 ][line width=0.08]  [draw opacity=0] (8.93,-4.29) -- (0,0) -- (8.93,4.29) -- (5.93,0) -- cycle    ;
%Curve Lines [id:da31794065785526515] 
\draw [color={rgb, 255:red, 173; green, 138; blue, 81 }  ,draw opacity=1 ]   (24.67,189.22) .. controls (24.96,215) and (68.56,215.65) .. (70.63,252.11) ;
\draw [shift={(70.71,254.99)}, rotate = 270.15999999999997] [fill={rgb, 255:red, 173; green, 138; blue, 81 }  ,fill opacity=1 ][line width=0.08]  [draw opacity=0] (8.93,-4.29) -- (0,0) -- (8.93,4.29) -- (5.93,0) -- cycle    ;
%Curve Lines [id:da2654148030518617] 
\draw [color={rgb, 255:red, 201; green, 49; blue, 53 }  ,draw opacity=1 ]   (63.55,18.21) .. controls (63.19,51.21) and (109.9,51.05) .. (109.59,83.98) ;
%Curve Lines [id:da4864234152586566] 
\draw [color={rgb, 255:red, 19; green, 117; blue, 183 }  ,draw opacity=1 ]   (56.4,18.21) .. controls (56.51,57.25) and (102.93,57.81) .. (102.44,83.98) ;
%Curve Lines [id:da4604609213021307] 
\draw [color={rgb, 255:red, 173; green, 138; blue, 81 }  ,draw opacity=1 ]   (70.71,18.21) .. controls (71.01,44.66) and (116.86,44.66) .. (116.75,83.98) ;
%Straight Lines [id:da5799419036271227] 
\draw [color={rgb, 255:red, 201; green, 49; blue, 53 }  ,draw opacity=1 ]   (17.51,103.71) -- (17.51,189.22) ;
%Straight Lines [id:da29777387358783725] 
\draw [color={rgb, 255:red, 19; green, 117; blue, 183 }  ,draw opacity=1 ]   (10.36,103.71) -- (10.36,189.22) ;
%Straight Lines [id:da01748450510750643] 
\draw [color={rgb, 255:red, 173; green, 138; blue, 81 }  ,draw opacity=1 ]   (24.67,103.71) -- (24.67,189.22) ;
%Curve Lines [id:da020833919656461752] 
\draw [color={rgb, 255:red, 201; green, 49; blue, 53 }  ,draw opacity=1 ]   (109.59,189.22) .. controls (109.23,222.22) and (63.85,222.06) .. (63.55,254.99) ;
%Curve Lines [id:da6578719769984881] 
\draw [color={rgb, 255:red, 173; green, 138; blue, 81 }  ,draw opacity=1 ]   (116.75,189.22) .. controls (116.86,228.26) and (71.2,228.82) .. (70.71,254.99) ;
%Curve Lines [id:da9830407922764246] 
\draw [color={rgb, 255:red, 19; green, 117; blue, 183 }  ,draw opacity=1 ]   (102.44,189.22) .. controls (102.69,211.5) and (69.93,215.01) .. (59.53,239.06) .. controls (57.58,243.57) and (56.42,248.79) .. (56.4,254.99) ;
%Straight Lines [id:da09599617650138148] 
\draw [color={rgb, 255:red, 201; green, 49; blue, 53 }  ,draw opacity=1 ]   (109.59,83.98) -- (109.59,189.22) ;
%Straight Lines [id:da0011168214791976716] 
\draw [color={rgb, 255:red, 19; green, 117; blue, 183 }  ,draw opacity=1 ]   (102.44,83.98) -- (102.44,189.22) ;
%Straight Lines [id:da6082260335764262] 
\draw [color={rgb, 255:red, 173; green, 138; blue, 81 }  ,draw opacity=1 ]   (116.75,83.98) -- (116.75,189.22) ;
%Curve Lines [id:da5248041547888255] 
\draw [color={rgb, 255:red, 189; green, 16; blue, 224 }  ,draw opacity=1 ]   (77.75,20.46) .. controls (78.25,40.41) and (124.25,40.41) .. (123.79,86.24) ;
%Curve Lines [id:da8499885227661569] 
\draw [color={rgb, 255:red, 189; green, 16; blue, 224 }  ,draw opacity=1 ]   (123.79,191.47) .. controls (123.25,231.91) and (77.75,233.41) .. (77.75,257.24) ;
%Straight Lines [id:da40643437504109836] 
\draw [color={rgb, 255:red, 189; green, 16; blue, 224 }  ,draw opacity=1 ]   (123.79,86.24) -- (123.79,191.47) ;
%Curve Lines [id:da5485340227435875] 
\draw [color={rgb, 255:red, 80; green, 227; blue, 194 }  ,draw opacity=1 ]   (47.75,18.68) .. controls (47.86,57.72) and (94.28,58.29) .. (93.79,84.46) ;
%Curve Lines [id:da8095319539638977] 
\draw [color={rgb, 255:red, 80; green, 227; blue, 194 }  ,draw opacity=1 ]   (93.79,189.69) .. controls (94.04,211.97) and (61.28,215.48) .. (50.88,239.54) .. controls (48.93,244.04) and (47.77,249.27) .. (47.75,255.46) ;
%Straight Lines [id:da3692132830340126] 
\draw [color={rgb, 255:red, 80; green, 227; blue, 194 }  ,draw opacity=1 ]   (93.79,84.46) -- (93.79,189.69) ;
%Straight Lines [id:da124244014839625] 
\draw [color={rgb, 255:red, 189; green, 16; blue, 224 }  ,draw opacity=1 ]   (77.75,17.46) -- (77.75,165.74) ;
\draw [shift={(77.75,168.74)}, rotate = 270] [fill={rgb, 255:red, 189; green, 16; blue, 224 }  ,fill opacity=1 ][line width=0.08]  [draw opacity=0] (8.93,-4.29) -- (0,0) -- (8.93,4.29) -- (5.93,0) -- cycle    ;
%Curve Lines [id:da4699834272831027] 
\draw [color={rgb, 255:red, 189; green, 16; blue, 224 }  ,draw opacity=1 ]   (31.71,102.97) .. controls (31.85,125.83) and (77.48,128.3) .. (77.81,165.79) ;
\draw [shift={(77.75,168.74)}, rotate = 272.98] [fill={rgb, 255:red, 189; green, 16; blue, 224 }  ,fill opacity=1 ][line width=0.08]  [draw opacity=0] (8.93,-4.29) -- (0,0) -- (8.93,4.29) -- (5.93,0) -- cycle    ;
%Curve Lines [id:da4361598919223372] 
\draw [color={rgb, 255:red, 80; green, 227; blue, 194 }  ,draw opacity=1 ]   (2.7,103.46) .. controls (2.81,140.36) and (44.23,150.12) .. (48.41,166.33) ;
\draw [shift={(48.75,169.24)}, rotate = 271.43] [fill={rgb, 255:red, 80; green, 227; blue, 194 }  ,fill opacity=1 ][line width=0.08]  [draw opacity=0] (8.93,-4.29) -- (0,0) -- (8.93,4.29) -- (5.93,0) -- cycle    ;
%Curve Lines [id:da3898203805473015] 
\draw [color={rgb, 255:red, 189; green, 16; blue, 224 }  ,draw opacity=1 ]   (78.71,17.21) .. controls (78.37,58.04) and (36.38,64.3) .. (32.87,80.11) ;
\draw [shift={(32.67,82.98)}, rotate = 265.83] [fill={rgb, 255:red, 189; green, 16; blue, 224 }  ,fill opacity=1 ][line width=0.08]  [draw opacity=0] (8.93,-4.29) -- (0,0) -- (8.93,4.29) -- (5.93,0) -- cycle    ;
%Curve Lines [id:da47414646759635193] 
\draw [color={rgb, 255:red, 80; green, 227; blue, 194 }  ,draw opacity=1 ]   (47.75,18.68) .. controls (48.33,37.43) and (5.1,44.08) .. (1.88,81.52) ;
\draw [shift={(1.71,84.46)}, rotate = 271.63] [fill={rgb, 255:red, 80; green, 227; blue, 194 }  ,fill opacity=1 ][line width=0.08]  [draw opacity=0] (8.93,-4.29) -- (0,0) -- (8.93,4.29) -- (5.93,0) -- cycle    ;
%Curve Lines [id:da0006857368192482172] 
\draw [color={rgb, 255:red, 80; green, 227; blue, 194 }  ,draw opacity=1 ]   (2.7,188.97) .. controls (2.81,226.64) and (46.04,228.49) .. (48.63,252.08) ;
\draw [shift={(48.75,254.74)}, rotate = 271.07] [fill={rgb, 255:red, 80; green, 227; blue, 194 }  ,fill opacity=1 ][line width=0.08]  [draw opacity=0] (8.93,-4.29) -- (0,0) -- (8.93,4.29) -- (5.93,0) -- cycle    ;
%Straight Lines [id:da2354673539484492] 
\draw [color={rgb, 255:red, 80; green, 227; blue, 194 }  ,draw opacity=1 ]   (2.7,103.46) -- (2.7,188.97) ;
%Curve Lines [id:da4150522355056352] 
\draw [color={rgb, 255:red, 189; green, 16; blue, 224 }  ,draw opacity=1 ]   (32.71,187.97) .. controls (33,213.76) and (76.6,214.4) .. (78.67,250.86) ;
\draw [shift={(78.75,253.74)}, rotate = 270.15999999999997] [fill={rgb, 255:red, 189; green, 16; blue, 224 }  ,fill opacity=1 ][line width=0.08]  [draw opacity=0] (8.93,-4.29) -- (0,0) -- (8.93,4.29) -- (5.93,0) -- cycle    ;
%Straight Lines [id:da4379024234792195] 
\draw [color={rgb, 255:red, 189; green, 16; blue, 224 }  ,draw opacity=1 ]   (32.71,104.46) -- (32.71,189.97) ;
%Straight Lines [id:da7851217722405489] 
\draw [color={rgb, 255:red, 189; green, 16; blue, 224 }  ,draw opacity=1 ]   (78.75,187.97) -- (78.75,253.74) ;
%Straight Lines [id:da9091013173556771] 
\draw [color={rgb, 255:red, 80; green, 227; blue, 194 }  ,draw opacity=1 ]   (47.75,188.97) -- (47.75,254.74) ;
%Straight Lines [id:da4933938329635057] 
\draw [color={rgb, 255:red, 80; green, 227; blue, 194 }  ,draw opacity=1 ]   (194.5,19.65) -- (194.5,170.93) ;
%Curve Lines [id:da6695129321968274] 
\draw [color={rgb, 255:red, 201; green, 49; blue, 53 }  ,draw opacity=1 ][line width=1.5]    (210.3,18.67) .. controls (209.95,50.36) and (168.23,51.62) .. (164.52,80.67) ;
\draw [shift={(164.26,84.44)}, rotate = 270.74] [fill={rgb, 255:red, 201; green, 49; blue, 53 }  ,fill opacity=1 ][line width=0.08]  [draw opacity=0] (11.07,-5.32) -- (0,0) -- (11.07,5.32) -- (7.35,0) -- cycle    ;
%Curve Lines [id:da7060842919584417] 
\draw [color={rgb, 255:red, 19; green, 117; blue, 183 }  ,draw opacity=1 ]   (203.15,18.67) .. controls (203.44,44.5) and (159.32,45.14) .. (157.19,81.57) ;
\draw [shift={(157.11,84.44)}, rotate = 269.89] [fill={rgb, 255:red, 19; green, 117; blue, 183 }  ,fill opacity=1 ][line width=0.08]  [draw opacity=0] (8.93,-4.29) -- (0,0) -- (8.93,4.29) -- (5.93,0) -- cycle    ;
%Curve Lines [id:da5503504405157855] 
\draw [color={rgb, 255:red, 173; green, 138; blue, 81 }  ,draw opacity=1 ][line width=3]    (217.46,18.67) .. controls (217.56,55.14) and (178.76,57.9) .. (172.32,78.6) ;
\draw [shift={(171.42,84.44)}, rotate = 270.65] [fill={rgb, 255:red, 173; green, 138; blue, 81 }  ,fill opacity=1 ][line width=0.08]  [draw opacity=0] (15.36,-7.38) -- (0,0) -- (15.36,7.38) -- (10.2,0) -- cycle    ;
%Straight Lines [id:da4920766164109307] 
\draw [color={rgb, 255:red, 201; green, 49; blue, 53 }  ,draw opacity=1 ]   (210.3,18.67) -- (210.3,169.95) ;
%Straight Lines [id:da8285075865010575] 
\draw [color={rgb, 255:red, 19; green, 117; blue, 183 }  ,draw opacity=1 ][line width=1.5]    (203.15,18.67) -- (203.15,169.95) ;
%Straight Lines [id:da10786683825413979] 
\draw [color={rgb, 255:red, 173; green, 138; blue, 81 }  ,draw opacity=1 ][line width=1.5]    (217.46,17.67) -- (217.46,168.95) ;
%Straight Lines [id:da15975457376626223] 
\draw [color={rgb, 255:red, 201; green, 49; blue, 53 }  ,draw opacity=1 ]   (210.3,189.68) -- (210.3,255.45) ;
%Straight Lines [id:da6758785677601716] 
\draw [color={rgb, 255:red, 19; green, 117; blue, 183 }  ,draw opacity=1 ]   (203.15,189.68) -- (203.15,255.45) ;
%Straight Lines [id:da019305210087180447] 
\draw [color={rgb, 255:red, 173; green, 138; blue, 81 }  ,draw opacity=1 ][line width=2.25]    (217.46,189.68) -- (217.46,255.45) ;
%Curve Lines [id:da9567022066792327] 
\draw [color={rgb, 255:red, 201; green, 49; blue, 53 }  ,draw opacity=1 ]   (164.26,104.18) .. controls (163.9,136.19) and (207.84,137) .. (210.2,167.07) ;
\draw [shift={(210.3,169.95)}, rotate = 270.52] [fill={rgb, 255:red, 201; green, 49; blue, 53 }  ,fill opacity=1 ][line width=0.08]  [draw opacity=0] (8.93,-4.29) -- (0,0) -- (8.93,4.29) -- (5.93,0) -- cycle    ;
%Curve Lines [id:da26750654057297796] 
\draw [color={rgb, 255:red, 19; green, 117; blue, 183 }  ,draw opacity=1 ]   (157.11,104.18) .. controls (157.22,141.85) and (200.44,143.69) .. (203.03,167.29) ;
\draw [shift={(203.15,169.95)}, rotate = 271.07] [fill={rgb, 255:red, 19; green, 117; blue, 183 }  ,fill opacity=1 ][line width=0.08]  [draw opacity=0] (8.93,-4.29) -- (0,0) -- (8.93,4.29) -- (5.93,0) -- cycle    ;
%Curve Lines [id:da09187723868344211] 
\draw [color={rgb, 255:red, 173; green, 138; blue, 81 }  ,draw opacity=1 ]   (171.42,104.18) .. controls (171.71,129.97) and (215.31,130.61) .. (217.38,167.07) ;
\draw [shift={(217.46,169.95)}, rotate = 270.15999999999997] [fill={rgb, 255:red, 173; green, 138; blue, 81 }  ,fill opacity=1 ][line width=0.08]  [draw opacity=0] (8.93,-4.29) -- (0,0) -- (8.93,4.29) -- (5.93,0) -- cycle    ;
%Curve Lines [id:da30609093164788725] 
\draw [color={rgb, 255:red, 201; green, 49; blue, 53 }  ,draw opacity=1 ][line width=1.5]    (164.26,189.68) .. controls (163.91,221.36) and (206.94,222.48) .. (210.12,251.66) ;
\draw [shift={(210.3,255.45)}, rotate = 270.52] [fill={rgb, 255:red, 201; green, 49; blue, 53 }  ,fill opacity=1 ][line width=0.08]  [draw opacity=0] (11.07,-5.32) -- (0,0) -- (11.07,5.32) -- (7.35,0) -- cycle    ;
%Curve Lines [id:da08080099476079272] 
\draw [color={rgb, 255:red, 19; green, 117; blue, 183 }  ,draw opacity=1 ][line width=2.25]    (157.11,189.68) .. controls (157.21,226.18) and (197.8,229.05) .. (202.68,250.66) ;
\draw [shift={(203.15,255.45)}, rotate = 271.07] [fill={rgb, 255:red, 19; green, 117; blue, 183 }  ,fill opacity=1 ][line width=0.08]  [draw opacity=0] (13.22,-6.35) -- (0,0) -- (13.22,6.35) -- (8.78,0) -- cycle    ;
%Curve Lines [id:da7280687659103448] 
\draw [color={rgb, 255:red, 173; green, 138; blue, 81 }  ,draw opacity=1 ][line width=2.25]    (171.42,189.68) .. controls (171.71,215.07) and (213.98,216.09) .. (217.26,250.92) ;
\draw [shift={(217.46,255.45)}, rotate = 270.15999999999997] [fill={rgb, 255:red, 173; green, 138; blue, 81 }  ,fill opacity=1 ][line width=0.08]  [draw opacity=0] (13.22,-6.35) -- (0,0) -- (13.22,6.35) -- (8.78,0) -- cycle    ;
%Curve Lines [id:da22713156627277264] 
\draw [color={rgb, 255:red, 201; green, 49; blue, 53 }  ,draw opacity=1 ][line width=2.25]    (210.3,18.67) .. controls (209.93,51.68) and (256.64,51.51) .. (256.34,84.44) ;
%Curve Lines [id:da9717280116988131] 
\draw [color={rgb, 255:red, 19; green, 117; blue, 183 }  ,draw opacity=1 ]   (203.15,18.67) .. controls (203.26,57.71) and (249.68,58.28) .. (249.19,84.44) ;
%Curve Lines [id:da08760659560406092] 
\draw [color={rgb, 255:red, 173; green, 138; blue, 81 }  ,draw opacity=1 ]   (217.46,18.67) .. controls (217.76,45.12) and (263.61,45.12) .. (263.5,84.44) ;
%Straight Lines [id:da7406345517047208] 
\draw [color={rgb, 255:red, 201; green, 49; blue, 53 }  ,draw opacity=1 ][line width=1.5]    (164.26,104.18) -- (164.26,189.68) ;
%Straight Lines [id:da04845592348160377] 
\draw [color={rgb, 255:red, 19; green, 117; blue, 183 }  ,draw opacity=1 ][line width=2.25]    (157.11,104.18) -- (157.11,189.68) ;
%Straight Lines [id:da49320740759284165] 
\draw [color={rgb, 255:red, 173; green, 138; blue, 81 }  ,draw opacity=1 ][line width=2.25]    (171.42,104.18) -- (171.42,189.68) ;
%Curve Lines [id:da7014827346423899] 
\draw [color={rgb, 255:red, 201; green, 49; blue, 53 }  ,draw opacity=1 ][line width=2.25]    (256.34,189.68) .. controls (255.98,222.69) and (210.6,222.52) .. (210.3,255.45) ;
%Curve Lines [id:da10092129898891888] 
\draw [color={rgb, 255:red, 173; green, 138; blue, 81 }  ,draw opacity=1 ]   (263.5,189.68) .. controls (263.61,228.72) and (217.95,229.28) .. (217.46,255.45) ;
%Curve Lines [id:da47593295360443344] 
\draw [color={rgb, 255:red, 19; green, 117; blue, 183 }  ,draw opacity=1 ]   (249.19,189.68) .. controls (249.44,211.96) and (216.68,215.47) .. (206.28,239.53) .. controls (204.33,244.03) and (203.17,249.25) .. (203.15,255.45) ;
%Straight Lines [id:da10708036576092861] 
\draw [color={rgb, 255:red, 201; green, 49; blue, 53 }  ,draw opacity=1 ][line width=2.25]    (256.34,84.44) -- (256.34,189.68) ;
%Straight Lines [id:da6509932398237541] 
\draw [color={rgb, 255:red, 19; green, 117; blue, 183 }  ,draw opacity=1 ]   (249.19,84.44) -- (249.19,189.68) ;
%Straight Lines [id:da5868879780095602] 
\draw [color={rgb, 255:red, 173; green, 138; blue, 81 }  ,draw opacity=1 ]   (263.5,84.44) -- (263.5,189.68) ;
%Curve Lines [id:da9800298837677484] 
\draw [color={rgb, 255:red, 189; green, 16; blue, 224 }  ,draw opacity=1 ][line width=1.5]    (224.5,20.93) .. controls (225,40.88) and (271,40.88) .. (270.54,86.7) ;
%Curve Lines [id:da21101589096385664] 
\draw [color={rgb, 255:red, 189; green, 16; blue, 224 }  ,draw opacity=1 ][line width=1.5]    (270.54,191.94) .. controls (270,232.38) and (224.5,233.88) .. (224.5,257.71) ;
%Straight Lines [id:da6397271378669] 
\draw [color={rgb, 255:red, 189; green, 16; blue, 224 }  ,draw opacity=1 ][line width=1.5]    (270.54,86.7) -- (270.54,191.94) ;
%Curve Lines [id:da2789987887778428] 
\draw [color={rgb, 255:red, 80; green, 227; blue, 194 }  ,draw opacity=1 ]   (194.5,19.15) .. controls (194.61,58.19) and (241.03,58.75) .. (240.54,84.92) ;
%Curve Lines [id:da8230925337689956] 
\draw [color={rgb, 255:red, 80; green, 227; blue, 194 }  ,draw opacity=1 ]   (240.54,190.16) .. controls (240.79,212.44) and (208.03,215.95) .. (197.63,240.01) .. controls (195.68,244.51) and (194.52,249.73) .. (194.5,255.93) ;
%Straight Lines [id:da9420850577996617] 
\draw [color={rgb, 255:red, 80; green, 227; blue, 194 }  ,draw opacity=1 ]   (240.54,84.92) -- (240.54,190.16) ;
%Straight Lines [id:da41711871675790346] 
\draw [color={rgb, 255:red, 189; green, 16; blue, 224 }  ,draw opacity=1 ]   (224.5,17.93) -- (224.5,166.21) ;
\draw [shift={(224.5,169.21)}, rotate = 270] [fill={rgb, 255:red, 189; green, 16; blue, 224 }  ,fill opacity=1 ][line width=0.08]  [draw opacity=0] (8.93,-4.29) -- (0,0) -- (8.93,4.29) -- (5.93,0) -- cycle    ;
%Curve Lines [id:da22786627986903674] 
\draw [color={rgb, 255:red, 189; green, 16; blue, 224 }  ,draw opacity=1 ]   (178.46,103.43) .. controls (178.6,126.29) and (224.23,128.77) .. (224.56,166.25) ;
\draw [shift={(224.5,169.21)}, rotate = 272.98] [fill={rgb, 255:red, 189; green, 16; blue, 224 }  ,fill opacity=1 ][line width=0.08]  [draw opacity=0] (8.93,-4.29) -- (0,0) -- (8.93,4.29) -- (5.93,0) -- cycle    ;
%Curve Lines [id:da8377025557370331] 
\draw [color={rgb, 255:red, 80; green, 227; blue, 194 }  ,draw opacity=1 ]   (149.45,103.93) .. controls (149.56,140.82) and (190.98,150.59) .. (195.16,166.8) ;
\draw [shift={(195.49,169.7)}, rotate = 271.43] [fill={rgb, 255:red, 80; green, 227; blue, 194 }  ,fill opacity=1 ][line width=0.08]  [draw opacity=0] (8.93,-4.29) -- (0,0) -- (8.93,4.29) -- (5.93,0) -- cycle    ;
%Curve Lines [id:da8908767283823245] 
\draw [color={rgb, 255:red, 189; green, 16; blue, 224 }  ,draw opacity=1 ]   (225.46,17.67) .. controls (225.12,58.5) and (183.13,64.77) .. (179.62,80.58) ;
\draw [shift={(179.42,83.44)}, rotate = 265.83] [fill={rgb, 255:red, 189; green, 16; blue, 224 }  ,fill opacity=1 ][line width=0.08]  [draw opacity=0] (8.93,-4.29) -- (0,0) -- (8.93,4.29) -- (5.93,0) -- cycle    ;
%Curve Lines [id:da6640692814346654] 
\draw [color={rgb, 255:red, 80; green, 227; blue, 194 }  ,draw opacity=1 ][line width=2.25]    (194.5,19.15) .. controls (195.07,37.61) and (153.16,44.34) .. (148.8,80.27) ;
\draw [shift={(148.46,84.92)}, rotate = 271.63] [fill={rgb, 255:red, 80; green, 227; blue, 194 }  ,fill opacity=1 ][line width=0.08]  [draw opacity=0] (13.22,-6.35) -- (0,0) -- (13.22,6.35) -- (8.78,0) -- cycle    ;
%Curve Lines [id:da09899654680810566] 
\draw [color={rgb, 255:red, 80; green, 227; blue, 194 }  ,draw opacity=1 ]   (149.45,189.43) .. controls (149.56,227.11) and (192.79,228.95) .. (195.38,252.55) ;
\draw [shift={(195.49,255.21)}, rotate = 271.07] [fill={rgb, 255:red, 80; green, 227; blue, 194 }  ,fill opacity=1 ][line width=0.08]  [draw opacity=0] (8.93,-4.29) -- (0,0) -- (8.93,4.29) -- (5.93,0) -- cycle    ;
%Straight Lines [id:da2569057625361113] 
\draw [color={rgb, 255:red, 80; green, 227; blue, 194 }  ,draw opacity=1 ]   (149.45,103.93) -- (149.45,189.43) ;
%Curve Lines [id:da39349381829885044] 
\draw [color={rgb, 255:red, 189; green, 16; blue, 224 }  ,draw opacity=1 ]   (179.46,188.43) .. controls (179.75,214.22) and (223.35,214.87) .. (225.42,251.33) ;
\draw [shift={(225.5,254.21)}, rotate = 270.15999999999997] [fill={rgb, 255:red, 189; green, 16; blue, 224 }  ,fill opacity=1 ][line width=0.08]  [draw opacity=0] (8.93,-4.29) -- (0,0) -- (8.93,4.29) -- (5.93,0) -- cycle    ;
%Straight Lines [id:da8704985399152049] 
\draw [color={rgb, 255:red, 189; green, 16; blue, 224 }  ,draw opacity=1 ]   (179.46,104.93) -- (179.46,190.43) ;
%Straight Lines [id:da8088088398839426] 
\draw [color={rgb, 255:red, 189; green, 16; blue, 224 }  ,draw opacity=1 ][line width=1.5]    (225.5,188.43) -- (225.5,254.21) ;
%Straight Lines [id:da8943471414595314] 
\draw [color={rgb, 255:red, 80; green, 227; blue, 194 }  ,draw opacity=1 ][line width=2.25]    (194.49,189.43) -- (194.49,255.21) ;
%Straight Lines [id:da27089610427615307] 
\draw [color={rgb, 255:red, 80; green, 227; blue, 194 }  ,draw opacity=1 ]   (344.5,19.65) -- (344.5,165.93) ;
\draw [shift={(344.5,168.93)}, rotate = 270] [fill={rgb, 255:red, 80; green, 227; blue, 194 }  ,fill opacity=1 ][line width=0.08]  [draw opacity=0] (8.93,-4.29) -- (0,0) -- (8.93,4.29) -- (5.93,0) -- cycle    ;
%Curve Lines [id:da1339298591511877] 
\draw [color={rgb, 255:red, 201; green, 49; blue, 53 }  ,draw opacity=1 ]   (360.3,18.67) .. controls (359.94,50.69) and (317.36,51.65) .. (314.42,81.58) ;
\draw [shift={(314.26,84.44)}, rotate = 270.74] [fill={rgb, 255:red, 201; green, 49; blue, 53 }  ,fill opacity=1 ][line width=0.08]  [draw opacity=0] (8.93,-4.29) -- (0,0) -- (8.93,4.29) -- (5.93,0) -- cycle    ;
%Curve Lines [id:da6657442174496553] 
\draw [color={rgb, 255:red, 173; green, 138; blue, 81 }  ,draw opacity=1 ]   (367.46,18.67) .. controls (367.57,56.93) and (324.87,58.09) .. (321.61,81.78) ;
\draw [shift={(321.42,84.44)}, rotate = 270.65] [fill={rgb, 255:red, 173; green, 138; blue, 81 }  ,fill opacity=1 ][line width=0.08]  [draw opacity=0] (8.93,-4.29) -- (0,0) -- (8.93,4.29) -- (5.93,0) -- cycle    ;
%Straight Lines [id:da8902140105478372] 
\draw [color={rgb, 255:red, 19; green, 117; blue, 183 }  ,draw opacity=1 ]   (353.15,18.67) -- (353.15,169.95) ;
%Straight Lines [id:da2862380684653816] 
\draw [color={rgb, 255:red, 173; green, 138; blue, 81 }  ,draw opacity=1 ]   (367.46,17.67) -- (367.46,168.95) ;
%Straight Lines [id:da6927489358455875] 
\draw [color={rgb, 255:red, 173; green, 138; blue, 81 }  ,draw opacity=1 ]   (367.46,189.68) -- (367.46,255.45) ;
%Curve Lines [id:da19221172443119938] 
\draw [color={rgb, 255:red, 19; green, 117; blue, 183 }  ,draw opacity=1 ]   (307.11,104.18) .. controls (307.22,141.85) and (350.44,143.69) .. (353.03,167.29) ;
\draw [shift={(353.15,169.95)}, rotate = 271.07] [fill={rgb, 255:red, 19; green, 117; blue, 183 }  ,fill opacity=1 ][line width=0.08]  [draw opacity=0] (8.93,-4.29) -- (0,0) -- (8.93,4.29) -- (5.93,0) -- cycle    ;
%Curve Lines [id:da4028601160953502] 
\draw [color={rgb, 255:red, 173; green, 138; blue, 81 }  ,draw opacity=1 ]   (321.42,104.18) .. controls (321.71,129.97) and (365.31,130.61) .. (367.38,167.07) ;
\draw [shift={(367.46,169.95)}, rotate = 270.15999999999997] [fill={rgb, 255:red, 173; green, 138; blue, 81 }  ,fill opacity=1 ][line width=0.08]  [draw opacity=0] (8.93,-4.29) -- (0,0) -- (8.93,4.29) -- (5.93,0) -- cycle    ;
%Curve Lines [id:da519023542239662] 
\draw [color={rgb, 255:red, 201; green, 49; blue, 53 }  ,draw opacity=1 ]   (314.26,189.68) .. controls (313.9,221.7) and (357.84,222.5) .. (360.2,252.58) ;
\draw [shift={(360.3,255.45)}, rotate = 270.52] [fill={rgb, 255:red, 201; green, 49; blue, 53 }  ,fill opacity=1 ][line width=0.08]  [draw opacity=0] (8.93,-4.29) -- (0,0) -- (8.93,4.29) -- (5.93,0) -- cycle    ;
%Curve Lines [id:da471166575922469] 
\draw [color={rgb, 255:red, 19; green, 117; blue, 183 }  ,draw opacity=1 ]   (307.11,189.68) .. controls (307.22,227.35) and (350.44,229.2) .. (353.03,252.8) ;
\draw [shift={(353.15,255.45)}, rotate = 271.07] [fill={rgb, 255:red, 19; green, 117; blue, 183 }  ,fill opacity=1 ][line width=0.08]  [draw opacity=0] (8.93,-4.29) -- (0,0) -- (8.93,4.29) -- (5.93,0) -- cycle    ;
%Curve Lines [id:da8489441326261478] 
\draw [color={rgb, 255:red, 173; green, 138; blue, 81 }  ,draw opacity=1 ]   (321.42,189.68) .. controls (321.71,215.47) and (365.31,216.11) .. (367.38,252.58) ;
\draw [shift={(367.46,255.45)}, rotate = 270.15999999999997] [fill={rgb, 255:red, 173; green, 138; blue, 81 }  ,fill opacity=1 ][line width=0.08]  [draw opacity=0] (8.93,-4.29) -- (0,0) -- (8.93,4.29) -- (5.93,0) -- cycle    ;
%Curve Lines [id:da5996813313184077] 
\draw [color={rgb, 255:red, 201; green, 49; blue, 53 }  ,draw opacity=1 ]   (360.3,18.67) .. controls (359.93,51.68) and (406.64,51.51) .. (406.34,84.44) ;
%Curve Lines [id:da9674589170490288] 
\draw [color={rgb, 255:red, 173; green, 138; blue, 81 }  ,draw opacity=1 ]   (367.46,18.67) .. controls (367.76,45.12) and (413.61,45.12) .. (413.5,84.44) ;
%Straight Lines [id:da6455715550674487] 
\draw [color={rgb, 255:red, 201; green, 49; blue, 53 }  ,draw opacity=1 ]   (314.26,104.18) -- (314.26,189.68) ;
%Straight Lines [id:da3131353493456247] 
\draw [color={rgb, 255:red, 19; green, 117; blue, 183 }  ,draw opacity=1 ]   (307.11,104.18) -- (307.11,189.68) ;
%Straight Lines [id:da4421271415676088] 
\draw [color={rgb, 255:red, 173; green, 138; blue, 81 }  ,draw opacity=1 ]   (321.42,104.18) -- (321.42,189.68) ;
%Curve Lines [id:da8021957174774711] 
\draw [color={rgb, 255:red, 201; green, 49; blue, 53 }  ,draw opacity=1 ]   (406.34,189.68) .. controls (405.98,222.69) and (360.6,222.52) .. (360.3,255.45) ;
%Curve Lines [id:da5811587347234891] 
\draw [color={rgb, 255:red, 173; green, 138; blue, 81 }  ,draw opacity=1 ]   (413.5,189.68) .. controls (413.61,228.72) and (367.95,229.28) .. (367.46,255.45) ;
%Straight Lines [id:da4246635914360911] 
\draw [color={rgb, 255:red, 201; green, 49; blue, 53 }  ,draw opacity=1 ]   (406.34,84.44) -- (406.34,189.68) ;
%Straight Lines [id:da4264675688569748] 
\draw [color={rgb, 255:red, 173; green, 138; blue, 81 }  ,draw opacity=1 ]   (413.5,84.44) -- (413.5,189.68) ;
%Curve Lines [id:da7780508136844696] 
\draw [color={rgb, 255:red, 189; green, 16; blue, 224 }  ,draw opacity=1 ]   (374.5,20.93) .. controls (375,40.88) and (421,40.88) .. (420.54,86.7) ;
%Curve Lines [id:da5825613609522491] 
\draw [color={rgb, 255:red, 189; green, 16; blue, 224 }  ,draw opacity=1 ]   (420.54,191.94) .. controls (420,232.38) and (374.5,233.88) .. (374.5,257.71) ;
%Straight Lines [id:da5755526441877564] 
\draw [color={rgb, 255:red, 189; green, 16; blue, 224 }  ,draw opacity=1 ]   (420.54,86.7) -- (420.54,191.94) ;
%Curve Lines [id:da7826918366370738] 
\draw [color={rgb, 255:red, 189; green, 16; blue, 224 }  ,draw opacity=1 ]   (328.46,103.43) .. controls (328.6,126.29) and (374.23,128.77) .. (374.56,166.25) ;
\draw [shift={(374.5,169.21)}, rotate = 272.98] [fill={rgb, 255:red, 189; green, 16; blue, 224 }  ,fill opacity=1 ][line width=0.08]  [draw opacity=0] (8.93,-4.29) -- (0,0) -- (8.93,4.29) -- (5.93,0) -- cycle    ;
%Curve Lines [id:da0472314935974687] 
\draw [color={rgb, 255:red, 80; green, 227; blue, 194 }  ,draw opacity=1 ]   (344.5,19.15) .. controls (345.08,37.9) and (301.84,44.55) .. (298.63,81.98) ;
\draw [shift={(298.46,84.92)}, rotate = 271.63] [fill={rgb, 255:red, 80; green, 227; blue, 194 }  ,fill opacity=1 ][line width=0.08]  [draw opacity=0] (8.93,-4.29) -- (0,0) -- (8.93,4.29) -- (5.93,0) -- cycle    ;
%Straight Lines [id:da6545070468659535] 
\draw [color={rgb, 255:red, 189; green, 16; blue, 224 }  ,draw opacity=1 ]   (375.5,188.43) -- (375.5,251.21) ;
\draw [shift={(375.5,254.21)}, rotate = 270] [fill={rgb, 255:red, 189; green, 16; blue, 224 }  ,fill opacity=1 ][line width=0.08]  [draw opacity=0] (8.93,-4.29) -- (0,0) -- (8.93,4.29) -- (5.93,0) -- cycle    ;
%Straight Lines [id:da6139933216147023] 
\draw [color={rgb, 255:red, 80; green, 227; blue, 194 }  ,draw opacity=1 ]   (344.49,189.43) -- (344.49,252.21) ;
\draw [shift={(344.49,255.21)}, rotate = 270] [fill={rgb, 255:red, 80; green, 227; blue, 194 }  ,fill opacity=1 ][line width=0.08]  [draw opacity=0] (8.93,-4.29) -- (0,0) -- (8.93,4.29) -- (5.93,0) -- cycle    ;

% Text Node
\draw  [fill={rgb, 255:red, 238; green, 238; blue, 238 }  ,fill opacity=1 ]  (42.07,-2) -- (83.07,-2) -- (83.07,20) -- (42.07,20) -- cycle  ;
\draw (62.57,9) node   [align=left] {\begin{minipage}[lt]{25.493335767125938pt}\setlength\topsep{0pt}
\begin{center}
0
\end{center}

\end{minipage}};
% Text Node
\draw  [fill={rgb, 255:red, 238; green, 238; blue, 238 }  ,fill opacity=1 ]  (-2,83.5) -- (39,83.5) -- (39,105.5) -- (-2,105.5) -- cycle  ;
\draw (18.5,94.5) node   [align=left] {\begin{minipage}[lt]{25.493335767125938pt}\setlength\topsep{0pt}
\begin{center}
1
\end{center}

\end{minipage}};
% Text Node
\draw  [fill={rgb, 255:red, 238; green, 238; blue, 238 }  ,fill opacity=1 ]  (42.07,167.69) -- (83.07,167.69) -- (83.07,189.69) -- (42.07,189.69) -- cycle  ;
\draw (62.57,178.69) node   [align=left] {\begin{minipage}[lt]{25.493335767125938pt}\setlength\topsep{0pt}
\begin{center}
2
\end{center}

\end{minipage}};
% Text Node
\draw  [fill={rgb, 255:red, 238; green, 238; blue, 238 }  ,fill opacity=1 ]  (42.07,253.2) -- (83.07,253.2) -- (83.07,275.2) -- (42.07,275.2) -- cycle  ;
\draw (62.57,264.2) node   [align=left] {\begin{minipage}[lt]{25.493335767125938pt}\setlength\topsep{0pt}
\begin{center}
3\\
\end{center}

\end{minipage}};
% Text Node
\draw (54.35,291.46) node [anchor=north west][inner sep=0.75pt]   [align=left] {a)};
% Text Node
\draw  [fill={rgb, 255:red, 238; green, 238; blue, 238 }  ,fill opacity=1 ]  (188.82,-1.54) -- (229.82,-1.54) -- (229.82,20.46) -- (188.82,20.46) -- cycle  ;
\draw (209.32,9.46) node   [align=left] {\begin{minipage}[lt]{25.493335767125938pt}\setlength\topsep{0pt}
\begin{center}
0
\end{center}

\end{minipage}};
% Text Node
\draw  [fill={rgb, 255:red, 238; green, 238; blue, 238 }  ,fill opacity=1 ]  (144.75,83.97) -- (185.75,83.97) -- (185.75,105.97) -- (144.75,105.97) -- cycle  ;
\draw (165.25,94.97) node   [align=left] {\begin{minipage}[lt]{25.493335767125938pt}\setlength\topsep{0pt}
\begin{center}
1
\end{center}

\end{minipage}};
% Text Node
\draw  [fill={rgb, 255:red, 238; green, 238; blue, 238 }  ,fill opacity=1 ]  (188.82,168.16) -- (229.82,168.16) -- (229.82,190.16) -- (188.82,190.16) -- cycle  ;
\draw (209.32,179.16) node   [align=left] {\begin{minipage}[lt]{25.493335767125938pt}\setlength\topsep{0pt}
\begin{center}
2
\end{center}

\end{minipage}};
% Text Node
\draw  [fill={rgb, 255:red, 238; green, 238; blue, 238 }  ,fill opacity=1 ]  (188.82,253.66) -- (229.82,253.66) -- (229.82,275.66) -- (188.82,275.66) -- cycle  ;
\draw (209.32,264.66) node   [align=left] {\begin{minipage}[lt]{25.493335767125938pt}\setlength\topsep{0pt}
\begin{center}
3\\
\end{center}

\end{minipage}};
% Text Node
\draw  [fill={rgb, 255:red, 238; green, 238; blue, 238 }  ,fill opacity=1 ]  (338.82,-1.54) -- (379.82,-1.54) -- (379.82,20.46) -- (338.82,20.46) -- cycle  ;
\draw (359.32,9.46) node   [align=left] {\begin{minipage}[lt]{25.493335767125938pt}\setlength\topsep{0pt}
\begin{center}
0
\end{center}

\end{minipage}};
% Text Node
\draw  [fill={rgb, 255:red, 238; green, 238; blue, 238 }  ,fill opacity=1 ]  (294.75,83.97) -- (335.75,83.97) -- (335.75,105.97) -- (294.75,105.97) -- cycle  ;
\draw (315.25,94.97) node   [align=left] {\begin{minipage}[lt]{25.493335767125938pt}\setlength\topsep{0pt}
\begin{center}
1
\end{center}

\end{minipage}};
% Text Node
\draw  [fill={rgb, 255:red, 238; green, 238; blue, 238 }  ,fill opacity=1 ]  (338.82,168.16) -- (379.82,168.16) -- (379.82,190.16) -- (338.82,190.16) -- cycle  ;
\draw (359.32,179.16) node   [align=left] {\begin{minipage}[lt]{25.493335767125938pt}\setlength\topsep{0pt}
\begin{center}
2
\end{center}

\end{minipage}};
% Text Node
\draw  [fill={rgb, 255:red, 238; green, 238; blue, 238 }  ,fill opacity=1 ]  (338.82,253.66) -- (379.82,253.66) -- (379.82,275.66) -- (338.82,275.66) -- cycle  ;
\draw (359.32,264.66) node   [align=left] {\begin{minipage}[lt]{25.493335767125938pt}\setlength\topsep{0pt}
\begin{center}
3\\
\end{center}

\end{minipage}};
% Text Node
\draw (200.75,291.46) node [anchor=north west][inner sep=0.75pt]   [align=left] {b)};
% Text Node
\draw (350.75,291.46) node [anchor=north west][inner sep=0.75pt]   [align=left] {c)};

\end{tikzpicture}

%% file: fig/ws-darts.tex
\tikzset{every picture/.style={line width=0.75pt}} %set default line width to 0.75pt        

\begin{tikzpicture}[x=0.75pt,y=0.75pt,yscale=-0.9,xscale=0.9]
%uncomment if require: \path (0,676); %set diagram left start at 0, and has height of 676

%Rounded Rect [id:dp9215369866337773] 
\draw   (80.91,55.02) .. controls (80.91,41.2) and (92.11,30) .. (105.93,30) -- (180.98,30) .. controls (194.8,30) and (206,41.2) .. (206,55.02) -- (206,254.98) .. controls (206,268.8) and (194.8,280) .. (180.98,280) -- (105.93,280) .. controls (92.11,280) and (80.91,268.8) .. (80.91,254.98) -- cycle ;
%Curve Lines [id:da6422532093821738] 
\draw [color={rgb, 255:red, 19; green, 117; blue, 183 }  ,draw opacity=1 ][line width=2.25]    (408.54,58.94) .. controls (408.26,84.07) and (376.02,85.71) .. (371.55,107.63) ;
\draw [shift={(371.04,112.51)}, rotate = 270.74] [fill={rgb, 255:red, 19; green, 117; blue, 183 }  ,fill opacity=1 ][line width=0.08]  [draw opacity=0] (10.36,-4.98) -- (0,0) -- (10.36,4.98) -- (6.88,0) -- cycle    ;
%Curve Lines [id:da2999416605083536] 
\draw [color={rgb, 255:red, 80; green, 227; blue, 194 }  ,draw opacity=1 ][line width=2.25]    (371.04,128.57) .. controls (370.76,153.71) and (404,155.22) .. (408.12,177.23) ;
\draw [shift={(408.54,182.14)}, rotate = 270.52] [fill={rgb, 255:red, 80; green, 227; blue, 194 }  ,fill opacity=1 ][line width=0.08]  [draw opacity=0] (10.36,-4.98) -- (0,0) -- (10.36,4.98) -- (6.88,0) -- cycle    ;
%Curve Lines [id:da12325882501290741] 
\draw [color={rgb, 255:red, 173; green, 138; blue, 81 }  ,draw opacity=1 ][line width=2.25]    (408.54,58.94) .. controls (408.24,85.82) and (446.28,85.69) .. (446.03,112.51) ;
%Curve Lines [id:da33883226195853133] 
\draw [color={rgb, 255:red, 173; green, 138; blue, 81 }  ,draw opacity=1 ][line width=2.25]    (446.03,198.21) .. controls (445.73,225.08) and (408.78,224.95) .. (408.54,251.77) ;
%Straight Lines [id:da023124849823858096] 
\draw [color={rgb, 255:red, 173; green, 138; blue, 81 }  ,draw opacity=1 ][line width=2.25]    (446.03,112.51) -- (446.03,198.21) ;
%Straight Lines [id:da23194639767243008] 
\draw [color={rgb, 255:red, 19; green, 117; blue, 183 }  ,draw opacity=1 ][line width=2.25]    (408.54,198.21) -- (408.54,246.77) ;
\draw [shift={(408.54,251.77)}, rotate = 270] [fill={rgb, 255:red, 19; green, 117; blue, 183 }  ,fill opacity=1 ][line width=0.08]  [draw opacity=0] (11.79,-5.66) -- (0,0) -- (11.79,5.66) -- (7.83,0) -- cycle    ;
%Rounded Rect [id:dp5422082786779072] 
\draw   (346,55.02) .. controls (346,41.2) and (357.2,30) .. (371.02,30) -- (446.07,30) .. controls (459.89,30) and (471.09,41.2) .. (471.09,55.02) -- (471.09,254.98) .. controls (471.09,268.8) and (459.89,280) .. (446.07,280) -- (371.02,280) .. controls (357.2,280) and (346,268.8) .. (346,254.98) -- cycle ;
%Shape: Rectangle [id:dp876760044073307] 
\draw  [fill={rgb, 255:red, 238; green, 238; blue, 238 }  ,fill opacity=1 ] (236,133) -- (316,133) -- (316,173) -- (236,173) -- cycle ;
%Straight Lines [id:da7734161239420554] 
\draw    (206,153) -- (234,153) ;
\draw [shift={(236,153)}, rotate = 180] [color={rgb, 255:red, 0; green, 0; blue, 0 }  ][line width=0.75]    (10.93,-3.29) .. controls (6.95,-1.4) and (3.31,-0.3) .. (0,0) .. controls (3.31,0.3) and (6.95,1.4) .. (10.93,3.29)   ;
%Straight Lines [id:da43222240133755174] 
\draw [line width=0.75]    (276,100) -- (276,131) ;
\draw [shift={(276,133)}, rotate = 270] [color={rgb, 255:red, 0; green, 0; blue, 0 }  ][line width=0.75]    (10.93,-3.29) .. controls (6.95,-1.4) and (3.31,-0.3) .. (0,0) .. controls (3.31,0.3) and (6.95,1.4) .. (10.93,3.29)   ;
%Shape: Rectangle [id:dp2958656404611426] 
\draw  [fill={rgb, 255:red, 230; green, 241; blue, 255 }  ,fill opacity=1 ] (267.43,60) -- (296,60) -- (296,88.57) -- (267.43,88.57) -- cycle ;
%Shape: Rectangle [id:dp36223059595942586] 
\draw  [fill={rgb, 255:red, 230; green, 241; blue, 255 }  ,fill opacity=1 ] (261.71,65.71) -- (290.29,65.71) -- (290.29,94.29) -- (261.71,94.29) -- cycle ;
%Shape: Rectangle [id:dp5705795907842319] 
\draw  [fill={rgb, 255:red, 230; green, 241; blue, 255 }  ,fill opacity=1 ] (256,71.43) -- (284.57,71.43) -- (284.57,100) -- (256,100) -- cycle ;
%Straight Lines [id:da7523903912923129] 
\draw    (316,152) -- (344,152) ;
\draw [shift={(346,152)}, rotate = 180] [color={rgb, 255:red, 0; green, 0; blue, 0 }  ][line width=0.75]    (10.93,-3.29) .. controls (6.95,-1.4) and (3.31,-0.3) .. (0,0) .. controls (3.31,0.3) and (6.95,1.4) .. (10.93,3.29)   ;
%Straight Lines [id:da7716174468223846] 
\draw [color={rgb, 255:red, 80; green, 227; blue, 194 }  ,draw opacity=1 ]   (133.58,54.56) -- (133.58,162.31) ;
\draw [shift={(133.58,165.31)}, rotate = 270] [fill={rgb, 255:red, 80; green, 227; blue, 194 }  ,fill opacity=1 ][line width=0.08]  [draw opacity=0] (8.93,-4.29) -- (0,0) -- (8.93,4.29) -- (5.93,0) -- cycle    ;
%Curve Lines [id:da05843778990592963] 
\draw [color={rgb, 255:red, 201; green, 49; blue, 53 }  ,draw opacity=1 ]   (145.31,53.83) .. controls (145.04,77.34) and (114.09,78.28) .. (111.34,99.83) ;
\draw [shift={(111.15,102.63)}, rotate = 270.74] [fill={rgb, 255:red, 201; green, 49; blue, 53 }  ,fill opacity=1 ][line width=0.08]  [draw opacity=0] (8.93,-4.29) -- (0,0) -- (8.93,4.29) -- (5.93,0) -- cycle    ;
%Curve Lines [id:da020324445473092423] 
\draw [color={rgb, 255:red, 173; green, 138; blue, 81 }  ,draw opacity=1 ]   (150.61,53.83) .. controls (150.69,81.77) and (120,83.04) .. (116.73,99.84) ;
\draw [shift={(116.46,102.63)}, rotate = 270.65] [fill={rgb, 255:red, 173; green, 138; blue, 81 }  ,fill opacity=1 ][line width=0.08]  [draw opacity=0] (8.93,-4.29) -- (0,0) -- (8.93,4.29) -- (5.93,0) -- cycle    ;
%Straight Lines [id:da9404630637802063] 
\draw [color={rgb, 255:red, 19; green, 117; blue, 183 }  ,draw opacity=1 ]   (140,53.83) -- (140,166.07) ;
%Straight Lines [id:da057776925728556305] 
\draw [color={rgb, 255:red, 173; green, 138; blue, 81 }  ,draw opacity=1 ]   (150.61,53.09) -- (150.61,165.33) ;
%Straight Lines [id:da15258262726911687] 
\draw [color={rgb, 255:red, 173; green, 138; blue, 81 }  ,draw opacity=1 ]   (150.61,180.71) -- (150.61,229.51) ;
%Curve Lines [id:da0957337847353783] 
\draw [color={rgb, 255:red, 19; green, 117; blue, 183 }  ,draw opacity=1 ]   (105.84,117.27) .. controls (105.92,144.79) and (137,146.54) .. (139.8,163.29) ;
\draw [shift={(140,166.07)}, rotate = 271.07] [fill={rgb, 255:red, 19; green, 117; blue, 183 }  ,fill opacity=1 ][line width=0.08]  [draw opacity=0] (8.93,-4.29) -- (0,0) -- (8.93,4.29) -- (5.93,0) -- cycle    ;
%Curve Lines [id:da13368178797328878] 
\draw [color={rgb, 255:red, 173; green, 138; blue, 81 }  ,draw opacity=1 ]   (116.46,117.27) .. controls (116.67,136.21) and (148.36,136.87) .. (150.5,163.11) ;
\draw [shift={(150.61,166.07)}, rotate = 270.15999999999997] [fill={rgb, 255:red, 173; green, 138; blue, 81 }  ,fill opacity=1 ][line width=0.08]  [draw opacity=0] (8.93,-4.29) -- (0,0) -- (8.93,4.29) -- (5.93,0) -- cycle    ;
%Curve Lines [id:da810405726450603] 
\draw [color={rgb, 255:red, 201; green, 49; blue, 53 }  ,draw opacity=1 ]   (111.15,180.71) .. controls (110.88,204.22) and (142.81,205.04) .. (145.17,226.69) ;
\draw [shift={(145.31,229.51)}, rotate = 270.52] [fill={rgb, 255:red, 201; green, 49; blue, 53 }  ,fill opacity=1 ][line width=0.08]  [draw opacity=0] (8.93,-4.29) -- (0,0) -- (8.93,4.29) -- (5.93,0) -- cycle    ;
%Curve Lines [id:da6662750310770948] 
\draw [color={rgb, 255:red, 19; green, 117; blue, 183 }  ,draw opacity=1 ]   (105.84,180.71) .. controls (105.92,208.23) and (137,209.98) .. (139.8,226.73) ;
\draw [shift={(140,229.51)}, rotate = 271.07] [fill={rgb, 255:red, 19; green, 117; blue, 183 }  ,fill opacity=1 ][line width=0.08]  [draw opacity=0] (8.93,-4.29) -- (0,0) -- (8.93,4.29) -- (5.93,0) -- cycle    ;
%Curve Lines [id:da726209927094595] 
\draw [color={rgb, 255:red, 173; green, 138; blue, 81 }  ,draw opacity=1 ]   (116.46,180.71) .. controls (116.67,199.65) and (148.36,200.31) .. (150.5,226.55) ;
\draw [shift={(150.61,229.51)}, rotate = 270.15999999999997] [fill={rgb, 255:red, 173; green, 138; blue, 81 }  ,fill opacity=1 ][line width=0.08]  [draw opacity=0] (8.93,-4.29) -- (0,0) -- (8.93,4.29) -- (5.93,0) -- cycle    ;
%Curve Lines [id:da3006251303234675] 
\draw [color={rgb, 255:red, 201; green, 49; blue, 53 }  ,draw opacity=1 ]   (145.31,53.83) .. controls (145.03,78.32) and (179.69,78.2) .. (179.47,102.63) ;
%Curve Lines [id:da18455312873178786] 
\draw [color={rgb, 255:red, 173; green, 138; blue, 81 }  ,draw opacity=1 ]   (150.61,53.83) .. controls (150.84,73.46) and (184.86,73.46) .. (184.77,102.63) ;
%Straight Lines [id:da0990737055395674] 
\draw [color={rgb, 255:red, 201; green, 49; blue, 53 }  ,draw opacity=1 ]   (111.15,117.27) -- (111.15,180.71) ;
%Straight Lines [id:da13995994148406754] 
\draw [color={rgb, 255:red, 19; green, 117; blue, 183 }  ,draw opacity=1 ]   (105.84,117.27) -- (105.84,180.71) ;
%Straight Lines [id:da15647114185945332] 
\draw [color={rgb, 255:red, 173; green, 138; blue, 81 }  ,draw opacity=1 ]   (116.46,117.27) -- (116.46,180.71) ;
%Curve Lines [id:da7700769672989227] 
\draw [color={rgb, 255:red, 201; green, 49; blue, 53 }  ,draw opacity=1 ]   (179.47,180.71) .. controls (179.19,205.2) and (145.53,205.07) .. (145.31,229.51) ;
%Curve Lines [id:da31467506179479] 
\draw [color={rgb, 255:red, 173; green, 138; blue, 81 }  ,draw opacity=1 ]   (184.77,180.71) .. controls (184.86,209.67) and (150.98,210.09) .. (150.61,229.51) ;
%Straight Lines [id:da729095957727697] 
\draw [color={rgb, 255:red, 201; green, 49; blue, 53 }  ,draw opacity=1 ]   (179.47,102.63) -- (179.47,180.71) ;
%Straight Lines [id:da6220967124269019] 
\draw [color={rgb, 255:red, 173; green, 138; blue, 81 }  ,draw opacity=1 ]   (184.77,102.63) -- (184.77,180.71) ;
%Curve Lines [id:da891022295686823] 
\draw [color={rgb, 255:red, 189; green, 16; blue, 224 }  ,draw opacity=1 ]   (155.84,55.51) .. controls (156.21,70.31) and (190.34,70.31) .. (190,104.3) ;
%Curve Lines [id:da6704305191430773] 
\draw [color={rgb, 255:red, 189; green, 16; blue, 224 }  ,draw opacity=1 ]   (190,182.38) .. controls (189.6,212.39) and (155.84,213.5) .. (155.84,231.18) ;
%Straight Lines [id:da04668942293333589] 
\draw [color={rgb, 255:red, 189; green, 16; blue, 224 }  ,draw opacity=1 ]   (190,104.3) -- (190,182.38) ;
%Curve Lines [id:da09173389857650993] 
\draw [color={rgb, 255:red, 189; green, 16; blue, 224 }  ,draw opacity=1 ]   (121.68,116.72) .. controls (121.78,133.59) and (155.29,135.49) .. (155.88,162.9) ;
\draw [shift={(155.84,165.52)}, rotate = 272.98] [fill={rgb, 255:red, 189; green, 16; blue, 224 }  ,fill opacity=1 ][line width=0.08]  [draw opacity=0] (8.93,-4.29) -- (0,0) -- (8.93,4.29) -- (5.93,0) -- cycle    ;
%Curve Lines [id:da2601654967624192] 
\draw [color={rgb, 255:red, 80; green, 227; blue, 194 }  ,draw opacity=1 ]   (133.58,54.19) .. controls (134.01,68.02) and (102.26,72.98) .. (99.58,100.38) ;
\draw [shift={(99.42,102.98)}, rotate = 271.63] [fill={rgb, 255:red, 80; green, 227; blue, 194 }  ,fill opacity=1 ][line width=0.08]  [draw opacity=0] (8.93,-4.29) -- (0,0) -- (8.93,4.29) -- (5.93,0) -- cycle    ;
%Straight Lines [id:da48038031864409025] 
\draw [color={rgb, 255:red, 189; green, 16; blue, 224 }  ,draw opacity=1 ]   (156.58,179.78) -- (156.58,225.58) ;
\draw [shift={(156.58,228.58)}, rotate = 270] [fill={rgb, 255:red, 189; green, 16; blue, 224 }  ,fill opacity=1 ][line width=0.08]  [draw opacity=0] (8.93,-4.29) -- (0,0) -- (8.93,4.29) -- (5.93,0) -- cycle    ;
%Straight Lines [id:da7036275324661845] 
\draw [color={rgb, 255:red, 80; green, 227; blue, 194 }  ,draw opacity=1 ]   (133.58,180.53) -- (133.58,226.32) ;
\draw [shift={(133.58,229.32)}, rotate = 270] [fill={rgb, 255:red, 80; green, 227; blue, 194 }  ,fill opacity=1 ][line width=0.08]  [draw opacity=0] (8.93,-4.29) -- (0,0) -- (8.93,4.29) -- (5.93,0) -- cycle    ;

% Text Node
\draw (71,5) node [anchor=north west][inner sep=0.75pt]   [align=left] {transfer architecture $ \lambda_i$};
% Text Node
\draw (310,6) node [anchor=north west][inner sep=0.75pt]   [align=left] {task-dependent architecture};
% Text Node
\draw  [fill={rgb, 255:red, 238; green, 238; blue, 238 }  ,fill opacity=1 ]  (390.24,249.77) -- (425.24,249.77) -- (425.24,268.77) -- (390.24,268.77) -- cycle  ;
\draw (407.74,259.27) node   [align=left] {\begin{minipage}[lt]{20.760834759008493pt}\setlength\topsep{0pt}
\begin{center}
3\\
\end{center}

\end{minipage}};
% Text Node
\draw  [fill={rgb, 255:red, 238; green, 238; blue, 238 }  ,fill opacity=1 ]  (390.24,180.14) -- (425.24,180.14) -- (425.24,199.14) -- (390.24,199.14) -- cycle  ;
\draw (407.74,189.64) node   [align=left] {\begin{minipage}[lt]{20.760834759008493pt}\setlength\topsep{0pt}
\begin{center}
2\\
\end{center}

\end{minipage}};
% Text Node
\draw  [fill={rgb, 255:red, 238; green, 238; blue, 238 }  ,fill opacity=1 ]  (354.35,111.58) -- (389.35,111.58) -- (389.35,130.58) -- (354.35,130.58) -- cycle  ;
\draw (371.85,121.08) node   [align=left] {\begin{minipage}[lt]{20.76083475900849pt}\setlength\topsep{0pt}
\begin{center}
1
\end{center}

\end{minipage}};
% Text Node
\draw  [fill={rgb, 255:red, 238; green, 238; blue, 238 }  ,fill opacity=1 ]  (390.24,41.94) -- (425.24,41.94) -- (425.24,60.94) -- (390.24,60.94) -- cycle  ;
\draw (407.74,51.44) node   [align=left] {\begin{minipage}[lt]{20.760834759008493pt}\setlength\topsep{0pt}
\begin{center}
0
\end{center}

\end{minipage}};
% Text Node
\draw (250,144) node [anchor=north west][inner sep=0.75pt]   [align=left] {DARTS};
% Text Node
\draw (257,77) node [anchor=north west][inner sep=0.75pt]   [align=left] {$\displaystyle t_{\text{new}}$};
% Text Node
\draw (191,290) node [anchor=north west][inner sep=0.75pt]   [align=left] {a) Warm-starting DARTS};
% Text Node
\draw  [fill={rgb, 255:red, 238; green, 238; blue, 238 }  ,fill opacity=1 ]  (128.57,38) -- (160.57,38) -- (160.57,56) -- (128.57,56) -- cycle  ;
\draw (144.57,47) node   [align=left] {\begin{minipage}[lt]{18.914410407867624pt}\setlength\topsep{0pt}
\begin{center}
0
\end{center}

\end{minipage}};
% Text Node
\draw  [fill={rgb, 255:red, 238; green, 238; blue, 238 }  ,fill opacity=1 ]  (95.88,101.44) -- (127.88,101.44) -- (127.88,119.44) -- (95.88,119.44) -- cycle  ;
\draw (111.88,110.44) node   [align=left] {\begin{minipage}[lt]{18.914410407867635pt}\setlength\topsep{0pt}
\begin{center}
1
\end{center}

\end{minipage}};
% Text Node
\draw  [fill={rgb, 255:red, 238; green, 238; blue, 238 }  ,fill opacity=1 ]  (128.57,163.9) -- (160.57,163.9) -- (160.57,181.9) -- (128.57,181.9) -- cycle  ;
\draw (144.57,172.9) node   [align=left] {\begin{minipage}[lt]{18.914410407867624pt}\setlength\topsep{0pt}
\begin{center}
2
\end{center}

\end{minipage}};
% Text Node
\draw  [fill={rgb, 255:red, 238; green, 238; blue, 238 }  ,fill opacity=1 ]  (128.57,227.34) -- (160.57,227.34) -- (160.57,245.34) -- (128.57,245.34) -- cycle  ;
\draw (144.57,236.34) node   [align=left] {\begin{minipage}[lt]{18.914410407867624pt}\setlength\topsep{0pt}
\begin{center}
3\\
\end{center}

\end{minipage}};
% Text Node
\draw (136.28,254.03) node [anchor=north west][inner sep=0.75pt]   [align=left] {c)};

\end{tikzpicture}

%% file: chapters/03_experiments.tex
Our experiments consist of two stages: collecting a set of transfer architectures using transfer architecture search and evaluation of the discovered architectures by warm-starting DARTS. We conduct two experiments to evaluate our approach on two different scenarios: (1) single task TAS and (2) multi-tasks TAS.

We have opted for sensible default configurations inspired by the authors of P-DARTS \citep{chen2019pdarts} across all tasks, with the exception of the operation level Dropout parameter. The alternative would require enormous amounts of hyper-parameter tuning to find the best-performing combination of configurations. While the configurations are likely not optimal for every single task, we observed that they are still performant enough to demonstrate our approach.

\subsection{Datasets}
\begin{figure}[h]
    \centering
        \includegraphics[width=0.75\linewidth]{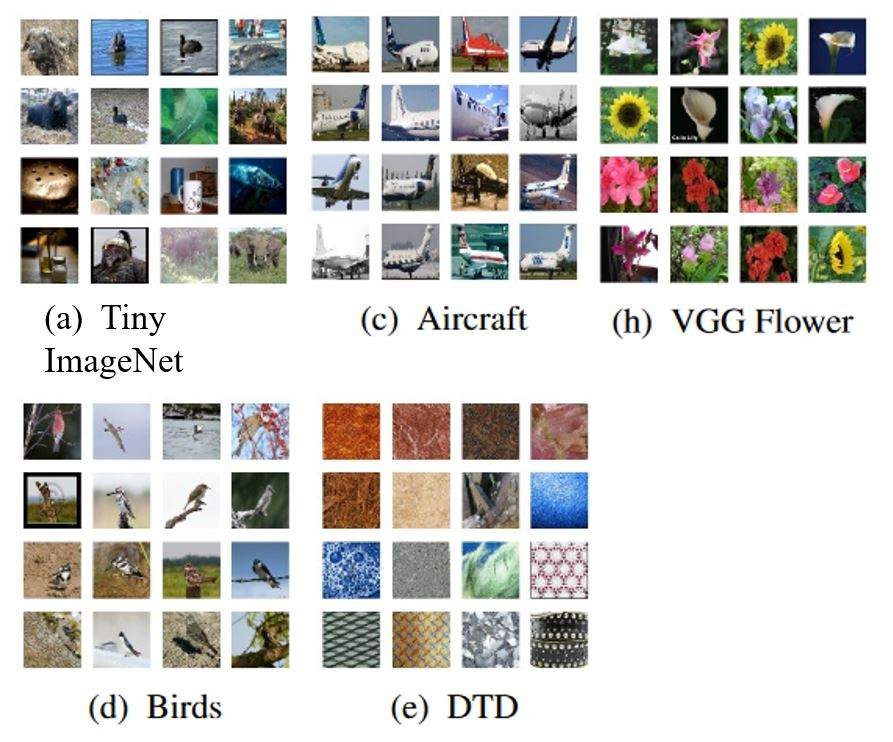}
    \caption{Example images from selected datasets.}
    \label{fig:img-datasets}
\end{figure}

We conduct experiments on the five datasets defined in Table \ref{table:p-darts} that are sampled from the meta-dataset collection by \cite{triantafillou2019metadataset}. Example images from the selected datasets can be observed in Figure \ref{fig:img-datasets}. Datasets were sampled based on dataset size and task difficulty, where \textit{birds} and \textit{dtd} are regarded as hard datasets. Because of the size of the ImageNet dataset, we replaced it with a down-sampled version called Tiny ImageNet\footnote{https://tiny-imagenet.herokuapp.com/}. Pre-processing of data follows the steps described by the authors of the meta-dataset \citep{triantafillou2019metadataset}, where images are resized to the shape $ 84 \times 84 \times 3 $ using bilinear interpolation.

We deterministically split datasets to create 2/3 training and 1/3 validation sets, where validation sets are only used in the final evaluations of our searched models. All sets are stratified. During architecture search, we further split the training set into two equal subsets following the P-DARTS \citep{chen2019pdarts}: one for fine-tuning network parameters and the other for tuning the architecture. 

\begin{table}[h]
\caption{List of datasets that were sampled from the meta-dataset \citep{triantafillou2019metadataset}.}
\centering
    \input{table/datasets}
\label{table:p-darts}
\centering
\end{table}

\subsection{Search Space} \label{search-space}
\begin{figure}[h]
    \centering
    \begin{subfigure}[b]{0.3\textwidth}
        \centering
        \input{fig/arch}
        \caption{}
    \label{fig:Ng1} 
    \end{subfigure}
    \begin{subfigure}[b]{0.69\textwidth}
        \includegraphics[width=1\linewidth]{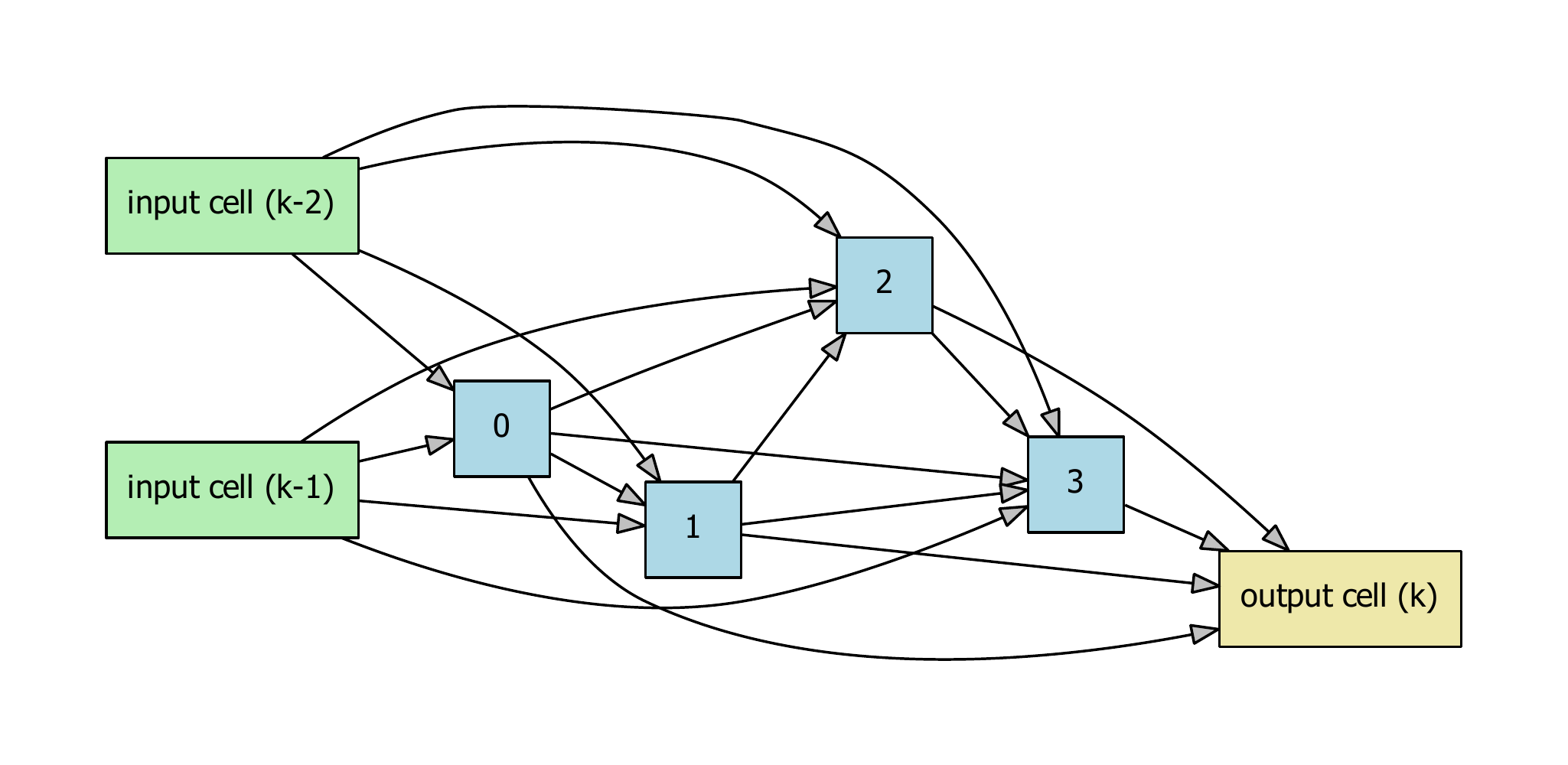}
        \caption{}
   \label{fig:Ng2} 
    \end{subfigure}
    \caption{Illustration of the search space: a) Overview of the whole architecture, where we manually stack \textit{normal} and \textit{reduction} cells together. This approach follows the common practice \citep{zoph2017learning, liu2019darts, chen2019pdarts} in the field of gradient descent based NAS. b) Cell architecture where we put a mixed operation $\bar{o}^{(i,j)}$ on every edge running into four intermediate nodes (nodes 0, 1, 2, 3 on the figure).}
    \label{fig:final-architecture}
\end{figure}

A well-designed search space plays a central role in finding well-performing architectures by NAS. We follow the standard practice of gradient descent based NAS \citep{liu2019darts, liu2017hierarchical, chen2019pdarts, real2019regularized}, where we only search for \textit{normal} and \textit{reduction} cells, while the outer shell architecture is manually set by stacking cells together, as can be observed in Figure \ref{fig:final-architecture}. Normal cells are the basic building blocks that compute the feature map of an image and have a stride of 1. Reduction cell, as can be deduced from the name, reduce the feature map dimensions and have a stride of 2. Although the decision on how to arrange normal and reduction cells can be viewed as another hyper-parameter, we follow the practice \citep{zoph2017learning, liu2019darts, chen2019pdarts} to put one reduction cell at $1/3$ and $2/3$ depth of the network.

A cell consists of 7 nodes, where we have two input nodes (input nodes are defined as the cell outputs in the previous two layers), a single output node, and four intermediate nodes where each intermediate node has 2 parents. Cell architecture and the initial set of candidate operations described below follows that of DARTS (\citep{liu2019darts}) and P-DARTS (\citep{chen2019pdarts})

Our initial set of candidate operations, $\mathcal{O}$, is as follows: 
\begin{AutoMultiColItemize}
  \item 3x3 max pooling
  \item 3x3 average pooling
  \item 3x3 separable conv
  \item 5x5 separable conv
  \item 3x3 dilated separable conv
  \item 5x5 dilated separable conv
  \item \textit{skip-connect}
  \item \textit{zero\footnote{Special operator that indicates a lack of connection between two nodes.}}
\end{AutoMultiColItemize}

\subsection{Task similarity measure}

In this work, we use Task2Vec \citep{achille2019task2vec} as our task similarity method. We processed images through a pre-trained ResNet34 probe network that was trained on the ImageNet dataset to compute embeddings based on estimates of the Fisher information matrix associated with the probe network parameters. Since the probe network is pre-trained, we have to resize our images to the shape $ 255 \times 255 \times 3 $ using bilinear interpolation before running Task2Vec algorithm. The cosine distance between normalized embeddings is used to measure the distance between tasks. The transfer architecture, $\lambda_i^*$, discovered on task $t_i^*$ with the shortest distance to $t_\text{new}$ is used to warm-start DARTS.

\begin{table}[h]
\caption{Task similarity matrix for selected tasks. Tasks with lower number are closer to each other and thus more similar.}
\centering
    \vrule\pgfplotstabletypeset[%
     color cells={min=0,max=0.025,textcolor=black},
     /pgfplots/colormap={blackwhite}{rgb255=(255,170,0) color=(white)},
     %/pgfplots/colormap={blackwhite}{gray(0cm)=(1); gray(1cm)=(0)},
    /pgf/number format/fixed,
    /pgf/number format/precision=3,
    col sep=comma,
    columns/Distance/.style={reset styles,string type}%
]{%%%%%%%
%Launch,0,1.36,1.20,1.46
%Space,1.36,1.1422,1.06,1.29
%Lunar,1.20,1.06,0, 1.05
%Moon,1.46,1.29,1.05,0.54
Distance,aircraft,dtd,birds,flower,imagenet
aircraft,0,,,,
dtd,0.025,0,,,
birds,0.020,0.023,0,,
flower,0.019,0.018,0.017,0,
imagenet,0.020,0.014,0.019,0.015,0
}\vrule
\label{table:task-similar}
\end{table}

\subsection{Setup of the transfer architecture search}
TAS is split into 3 stages of increasing depth: initial, intermediate, and final. In the initial stage, 5 cells are stacked in the search network and the full operation space is used, where we have all 8 candidate operations on each edge. Next, in the intermediate stage, we increase the number of stacked cells from 5 to 11 and reduce the number of operations preserved on each edge to 5 by removing the operations with the lowest $\alpha$. In the final stage, the search network consists of 17 stacked cells and each cell keeps only the 3 best performing candidate operations on each edge. 

TAS hyper-parameters are guided by the GPU memory limitations, where for each stage we train a network for 25 epochs with a batch size of 96. The setup closely follows one described by the authors of P-DARTS \citep{chen2019pdarts}. In the first 10 epochs, only network parameters are tuned while in the remaining 15 epochs we jointly learn network and architectural parameters. Furthermore, for additional acceleration, we use the first-order DARTS optimization scheme. For architecture parameters, an Adam optimizer with learning rate $\eta=0.0006$, momentum ($\beta=(0.5,0.999)$ and weight decay $0.001$ is used. To simplify the experiment architecture parameters are not tuned per task. 

\subsection{Setup of warm-starting DARTS}
Warm-started DARTS (WS-DARTS) follows the configuration of the final stage of TAS. The search network consists of 17 stacked cells and a transfer architecture with three candidate operations on every edge. We use the same architecture parameters as for the transfer architecture search described above. While it is possible to have an independent set of settings for each WS-DARTS, we want discovered architectures to be comparable between different experiments. 

\subsection{Evaluation}
To select the architecture for the final evaluation, we run WS-DARTS 10 times with different settings of the operation level Dropout (from 0 to 0.9 in steps of 0.1). In essence, we are performing a hyper-parameter search on the training set to find the optimal value of \textit{skip-connect} dropout. In our preliminary experiments, we found that the value of \textit{skip-connect} dropout plays a decisive role in finding a well-performing architecture or a poorly performing one. Found architectures are then trained from scratch for a short period of 75 epochs and the best performing architectures are selected for the final evaluation. 

The final evaluation of best performing discovered networks follows that of P-DARTS \cite{chen2019pdarts}. An evaluation network of 20 cells and 16 initial channels is trained from scratch (weights learned during the search are discarded) for 300 epochs with batch size 96. This network is learned on the training set while the performance is evaluated on the validation set. The validation set is never used before the evaluation phase of the final architecture. Additionally, cutout regularization of length 16, drop-path of probability 0.3 and auxiliary towers \citep{szegedy2014going} of weight 0.4 are applied to further improve performance. Auxiliary towers are used to help reduce the vanishing gradient problem in our deep networks. A standard SGD optimizer with a weight decay of 0.0003 and a momentum of 0.9 is used. The initial learning rate is 0.025. 

\subsection{Experiment 1 } \label{sec:e1}

We perform TAS on every task, $t_i \in \mathcal{T}$, to compute a transfer architecture, $\lambda_i$, and learned transfer architecture, $\hat{\lambda}_i$. $\lambda_i$ and $\hat{\lambda}_i$ are saved for later (re-)use. Based on the similarity measure (Table \ref{table:task-similar}), we perform WS-DARTS for every task $t_i \in \mathcal{T}$, where $\lambda_i$ and $\hat{\lambda}_i$ are selected from the most similar task. The Dropout probability is determined for every task separately by fine-tuning the parameter, where we keep the values that resulted in the best performing architecture for the same task.

\subsection{Experiment 2 } \label{sec:e2}

We devised a simple definition of a meta-dataset to demonstrate an alternative approach to WS-DARTS. Task are split into meta-train and meta-test groups as follows: $\{dtd, flower, \allowbreak tiny\_imagenet\} \in  \mathcal{T}_\text{meta-train}$ and $\{ aircraf, birds \} \in \mathcal{T}_\text{meta-test}$.  $\mathcal{T}_\text{meta-train}$ is used to compute \textit{meta-transfer architecture} $\lambda_\text{meta}$. $\lambda_\text{meta}$ is discovered by TAS algorithm, where tasks $\mathcal{T}_\text{meta}$ are jointly learned in order from the smallest to the largest task (as written). The Dropout probability on skip-connect is determined by   fine-tuning the parameter on $\mathcal{T}_\text{meta-test}$. The baseline architecture was determined by performing P-DARTS on our meta-dataset. 

%% file: table/datasets.tex
\resizebox{\textwidth}{!}{%
\begin{tabular}{@{}cccc@{}}
\toprule
ID        & Dataset name         & $N$ classes & $N$ observations \\ \midrule
aircraft       & FGVC-AIRCRAF \citep{maji13fine-grained}         & $100$       & $10000$          \\
flower         & VGG Flower \citep{Nilsback08}           & $102$       & $8189$           \\
birds          & CUB-200-2011 \citep{WahCUB_200_2011}         & $200$       & $11788$          \\
dtd            & Describable Textures \citep{cimpoi14describing} & $47$        & $5640$           \\
tiny\_imagenet & Tiny ImageNet        & $200$       & $100000$         \\ \bottomrule
\end{tabular}%
}

%% file: fig/arch.tex
\tikzset{every picture/.style={line width=0.75pt}} %set default line width to 0.75pt        

\begin{tikzpicture}[x=0.75pt,y=0.6pt,yscale=-1,xscale=1]
%uncomment if require: \path (0,415); %set diagram left start at 0, and has height of 415

% Text Node
\draw    (38.82,306.22) -- (147.82,306.22) -- (147.82,336.22) -- (38.82,336.22) -- cycle  ;
\draw (93.32,321.22) node   [align=left] {\begin{minipage}[lt]{71.52444000000001pt}\setlength\topsep{0pt}
\begin{center}
Image
\end{center}

\end{minipage}};
% Text Node
\draw  [fill={rgb, 255:red, 240; green, 240; blue, 240 }  ,fill opacity=1 ]  (38.82,43.26) -- (147.82,43.26) -- (147.82,73.26) -- (38.82,73.26) -- cycle  ;
\draw (93.32,58.26) node   [align=left] {\begin{minipage}[lt]{71.52444000000001pt}\setlength\topsep{0pt}
\begin{center}
Normal Cell
\end{center}

\end{minipage}};
% Text Node
\draw  [fill={rgb, 255:red, 221; green, 221; blue, 221 }  ,fill opacity=1 ]  (38.82,95.85) -- (147.82,95.85) -- (147.82,125.85) -- (38.82,125.85) -- cycle  ;
\draw (93.32,110.85) node   [align=left] {\begin{minipage}[lt]{71.52444000000001pt}\setlength\topsep{0pt}
\begin{center}
Reduction Cell
\end{center}

\end{minipage}};
% Text Node
\draw (64.38,2) node [anchor=north west][inner sep=0.75pt]   [align=left] {Softmax};
% Text Node
\draw  [fill={rgb, 255:red, 240; green, 240; blue, 240 }  ,fill opacity=1 ]  (38.82,148.45) -- (147.82,148.45) -- (147.82,178.45) -- (38.82,178.45) -- cycle  ;
\draw (93.32,163.45) node   [align=left] {\begin{minipage}[lt]{71.52444000000001pt}\setlength\topsep{0pt}
\begin{center}
Normal Cell
\end{center}

\end{minipage}};
% Text Node
\draw  [fill={rgb, 255:red, 221; green, 221; blue, 221 }  ,fill opacity=1 ]  (38.82,201.04) -- (147.82,201.04) -- (147.82,231.04) -- (38.82,231.04) -- cycle  ;
\draw (93.32,216.04) node   [align=left] {\begin{minipage}[lt]{71.52444000000001pt}\setlength\topsep{0pt}
\begin{center}
Reduction Cell
\end{center}

\end{minipage}};
% Text Node
\draw  [fill={rgb, 255:red, 240; green, 240; blue, 240 }  ,fill opacity=1 ]  (38.82,253.63) -- (147.82,253.63) -- (147.82,283.63) -- (38.82,283.63) -- cycle  ;
\draw (93.32,268.63) node   [align=left] {\begin{minipage}[lt]{71.52444000000001pt}\setlength\topsep{0pt}
\begin{center}
Normal Cell
\end{center}

\end{minipage}};
% Text Node
\draw (1,49.27) node [anchor=north west][inner sep=0.75pt]   [align=left] {$\displaystyle N\times $};
% Text Node
\draw (1,154.45) node [anchor=north west][inner sep=0.75pt]   [align=left] {$\displaystyle N\times $};
% Text Node
\draw (1,259.63) node [anchor=north west][inner sep=0.75pt]   [align=left] {$\displaystyle N\times $};
% Connection
\draw    (93.32,306.22) -- (93.32,285.63) ;
\draw [shift={(93.32,283.63)}, rotate = 450] [color={rgb, 255:red, 0; green, 0; blue, 0 }  ][line width=0.75]    (10.93,-3.29) .. controls (6.95,-1.4) and (3.31,-0.3) .. (0,0) .. controls (3.31,0.3) and (6.95,1.4) .. (10.93,3.29)   ;
% Connection
\draw    (93.32,253.63) -- (93.32,233.04) ;
\draw [shift={(93.32,231.04)}, rotate = 450] [color={rgb, 255:red, 0; green, 0; blue, 0 }  ][line width=0.75]    (10.93,-3.29) .. controls (6.95,-1.4) and (3.31,-0.3) .. (0,0) .. controls (3.31,0.3) and (6.95,1.4) .. (10.93,3.29)   ;
% Connection
\draw    (93.32,201.04) -- (93.32,180.45) ;
\draw [shift={(93.32,178.45)}, rotate = 450] [color={rgb, 255:red, 0; green, 0; blue, 0 }  ][line width=0.75]    (10.93,-3.29) .. controls (6.95,-1.4) and (3.31,-0.3) .. (0,0) .. controls (3.31,0.3) and (6.95,1.4) .. (10.93,3.29)   ;
% Connection
\draw    (93.32,148.45) -- (93.32,127.85) ;
\draw [shift={(93.32,125.85)}, rotate = 450] [color={rgb, 255:red, 0; green, 0; blue, 0 }  ][line width=0.75]    (10.93,-3.29) .. controls (6.95,-1.4) and (3.31,-0.3) .. (0,0) .. controls (3.31,0.3) and (6.95,1.4) .. (10.93,3.29)   ;
% Connection
\draw    (93.32,95.85) -- (93.32,75.26) ;
\draw [shift={(93.32,73.26)}, rotate = 450] [color={rgb, 255:red, 0; green, 0; blue, 0 }  ][line width=0.75]    (10.93,-3.29) .. controls (6.95,-1.4) and (3.31,-0.3) .. (0,0) .. controls (3.31,0.3) and (6.95,1.4) .. (10.93,3.29)   ;
% Connection
\draw    (93.18,43.26) -- (93.01,25) ;
\draw [shift={(92.99,23)}, rotate = 449.47] [color={rgb, 255:red, 0; green, 0; blue, 0 }  ][line width=0.75]    (10.93,-3.29) .. controls (6.95,-1.4) and (3.31,-0.3) .. (0,0) .. controls (3.31,0.3) and (6.95,1.4) .. (10.93,3.29)   ;

\end{tikzpicture}

%% file: chapters/04_results.tex
In this section, we analyze our approach and present the results of our experiments. We evaluate our approach by two criteria: (1) required computational resources to run warm-started DARTS (measured in seconds) and (2) performance of the discovered architectures. All our experiments were conducted on $2\times$ NVIDIA Tesla K80 GPUs. While our experiment setup closely mimics that of P-DARTS, it is hard to make direct comparisons due to the differences in hardware, namely GPU memory size. We used inferior GPUs in our experiments compared to DARTS \citep{liu2019darts} or P-DARTS \citep{chen2019pdarts}.

\subsection{Experiment 1} \label{e1}

\begin{figure}[h]
    \centering
    \begin{subfigure}[b]{1.1\textwidth}
        \includegraphics[width=1.1\linewidth,center]{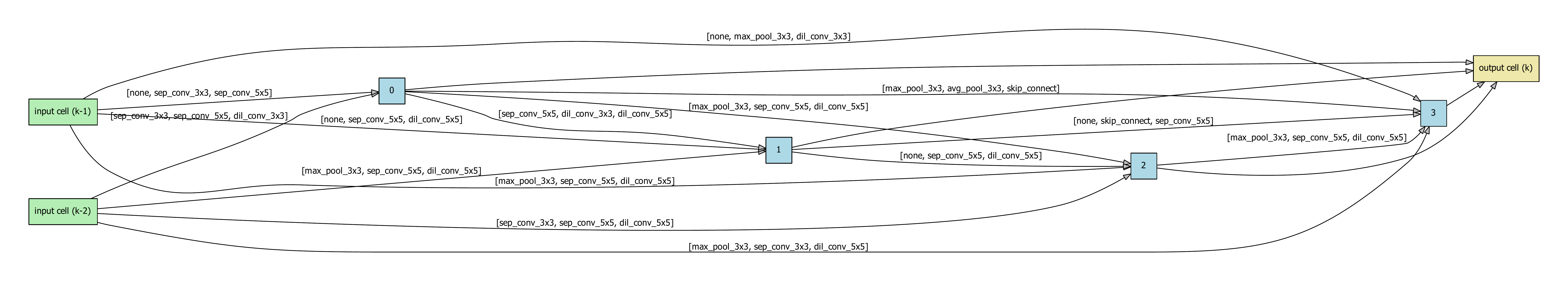}
        \caption{FLOWER\_V3 transfer architecture for normal cell discovered on the \textit{flower} dataset. On each edge, a combination of three candidate operations can be observed.}
   \label{fig:TA} 
    \end{subfigure}
    \begin{subfigure}[b]{0.49\textwidth}
        \includegraphics[width=1\linewidth]{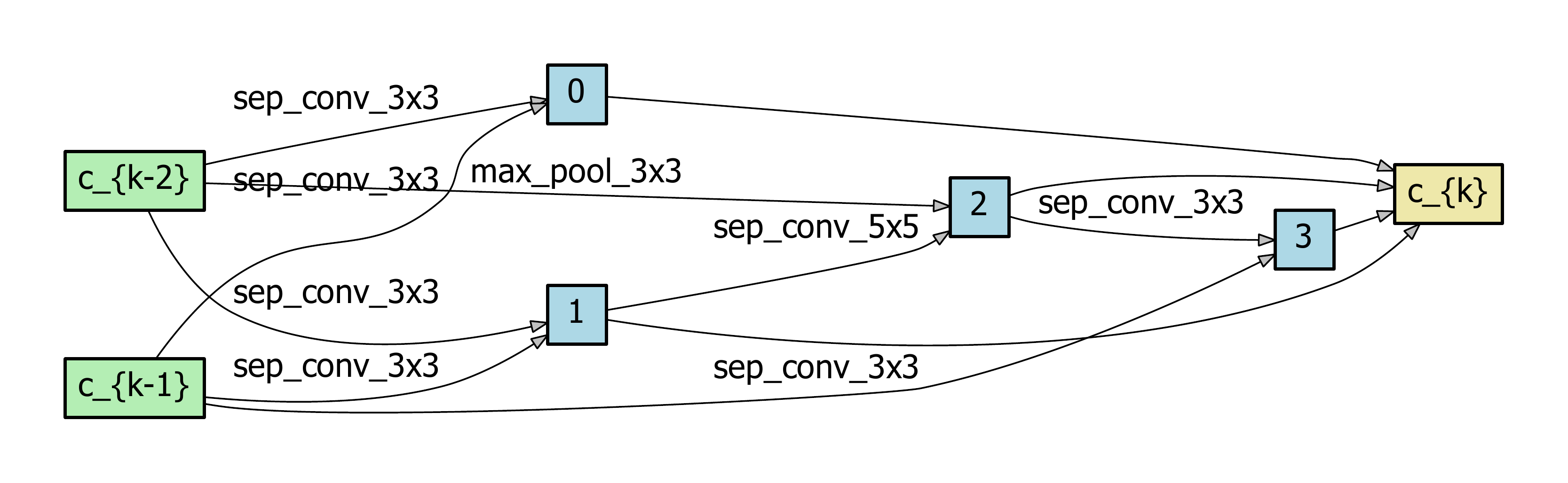}
        \caption{Normal cell found on the \textit{aircraft} task by warm-stating DARTS using the transfer architecture shown in (a).}
   \label{fig:normal_cell_a} 
    \end{subfigure}
    \begin{subfigure}[b]{0.49\textwidth}
\includegraphics[width=1\linewidth]{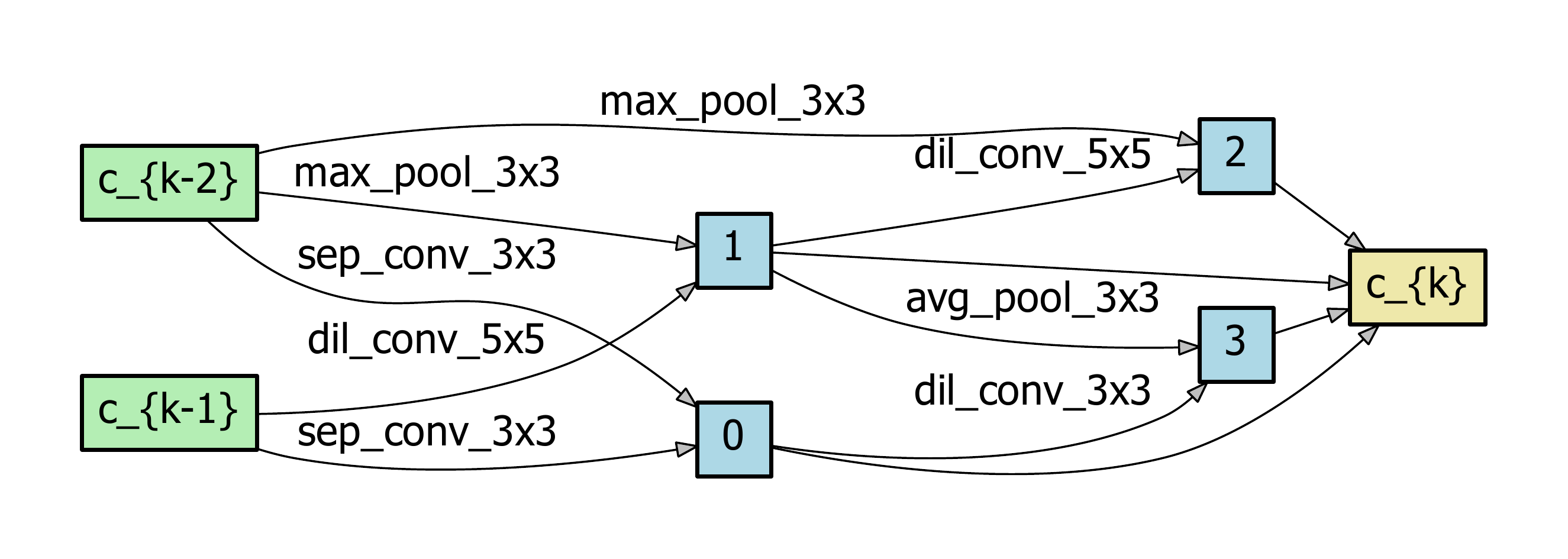}
        \caption{Normal cell found on the \textit{birds} task by warm-stating DARTS using the transfer architecture shown in (a).}
   \label{fig:normal_cell_b}
    \end{subfigure}
    \caption{Visualization of (a) transfer architecture for normal cell, $\lambda_\text{flower}$, and normal cells discovered by warm-starting DARTS using $\lambda_\text{flower}$ on (b) \textit{aircraft} task and (c) \textit{birds} task.}
    \label{fig:e1}
\end{figure}

Figure \ref{fig:TA} visualizes a transfer architecture, $\lambda_\text{flower}$, discovered by TAS on the \textit{flower} dataset from the initial search space, $\mathcal{O}$, described in section \ref{search-space}. Following the task similarity matrix shown in Table \ref{table:task-similar}, we used $\lambda_\text{flower}$ to warm-start DARTS on the \textit{aircraft} and \textit{birds} datasets as $\lambda_\text{flower}$ is the transfer architecture found on the most similar dataset \textit{flower}. The final architectures discovered on the \textit{aircraft} and \textit{birds} datasets are shown in Figure \ref{fig:normal_cell_a} and \ref{fig:normal_cell_b}, respectively. 

Our proposed approach finds the final architectures significantly faster than running P-DARTS (or slower DARTS) from scratch regardless of the task. This behavior was expected and anticipated as reducing the search space will lead to faster search times. In Table \ref{table:time}, we can observe that, on average, warm-started DARTS resulted in finding final architecture 60\% faster compared to P-DARTS. Comparison between the two warm-starting approaches, transfer architecture $\lambda_i$ and learned transfer architecture $\hat{\lambda}_i$, shows that learned transfer architecture, $\hat{\lambda}_i$, requires more computational resources. $\hat{\lambda}_i$ is on average faster than P-DARTS by only $58.5\%$ compared to $\lambda_i$'s $64.7\%$. This difference is especially visible on smaller datasets (\textit{aircraft}, \textit{flower}, \textit{birds}, and \textit{dtd}), whereas on \textit{tiny\_imagenet} there was no significant statistical difference in means at $p < 0.05$.

\begin{table}[h]
\caption{Search times for WS-DARTS performed on selected tasks. The P-DARTS column represents search times of P-DARTS algorithm from scratch as a baseline for comparison.}
\centering
    \input{table/time}
\label{table:time}
\end{table}

For the second evaluation criteria, we looked at the performance results of the discovered architectures. Our goal is to find better-performing architectures in comparison to the \textit{baseline} architecture, where the \textit{baseline} architecture is obtained by running P-DARTS on the most similar dataset. Since we can already obtain the final P-DARTS architecture from our learned transfer architecture, $\hat{\lambda}_i$, by replaced each mixed operation $\bar{o}^{(i,j)}$ with the most likely operation. This served two purposes: (1) it saved resources that could be used on other experiments and (2) more importantly it provided that the final architecture is a sub-graph of the transfer architecture, $\hat{\lambda}_i$. Later allows for a more equal comparison of results.

In Table \ref{table:performance}, accuracy results of our approach are shown and compared with baseline results and results from the same tasks from the meta-dataset authors \citep{triantafillou2019metadataset}. Warm-started DARTS was able to find better-performing architectures on all 5 tasks compared to the baseline architectures. We discovered that finding competitive architectures using WS-DARTS($\lambda$) is much harder compared to WS-DARTS($\hat{\lambda}_i$) and only successful in 3 out of 5 experiments. For harder tasks \textit{dtd} and \textit{birds} we were unable to find competitive architectures. This can be attributed to multiple reasons: 1) the most similar task is not similar enough, 2) the initial search space is too restrictive, and 3) the search or evaluation setup are not suitable. Further research would be needed to determent why we were unable to find better-performing architectures.

Since any discovered architecture by WS-DARTS($\lambda_i$) can only be a sub-graph of $\lambda_i$, we know that it is possible to find at least as good architectures as is our best one for the specific task. Looking more closely at the results of the \textit{tiny\_imagenet} task, we observe, that $\lambda_i$ performed slightly better than $\hat{\lambda}_i$. Our intuition would suggest that $\hat{\lambda}_i$ does not give any additional boost to tasks with large datasets. However, this is not proven and additional research is needed as the simple explanation could also be that we were unsuccessful in finding better performing architecture for $\hat{\lambda}_i$. 

In comparison to the meta-dataset \citep{triantafillou2019metadataset}, we were able to find better performing architectures for three datasets: \textit{airplane}, \textit{flower} and \textit{tiny\_imagenet}. We are comparing our results with the best results found by the authors of the meta-dataset by algorithms that are not based on DARTS. \textit{Tiny\_imagenet} task is a smaller version of a much larger \textit{ImageNet}\citep{imagenet_cvpr09} dataset and should be taken into consideration when directly comparing the two results. However, we could not find competitive architectures for harder tasks \textit{birds} and \textit{dtd} by either with our approach or by searching with P-DARTS from scratch. 

\begin{table}[h]
\caption{Performance results showing validation accuracy of evaluated architectures. Meta-dataset \citep{triantafillou2019metadataset} results are the best results obtained by the authors and are used for the external performance comparison between different approaches.}
\centering
    \input{table/acc}
\label{table:performance}
\end{table}

\subsection{Experiment 2}

\begin{figure}[h]
    \centering
    \begin{subfigure}[b]{1.1\textwidth}
        \includegraphics[width=1.1\linewidth, center]{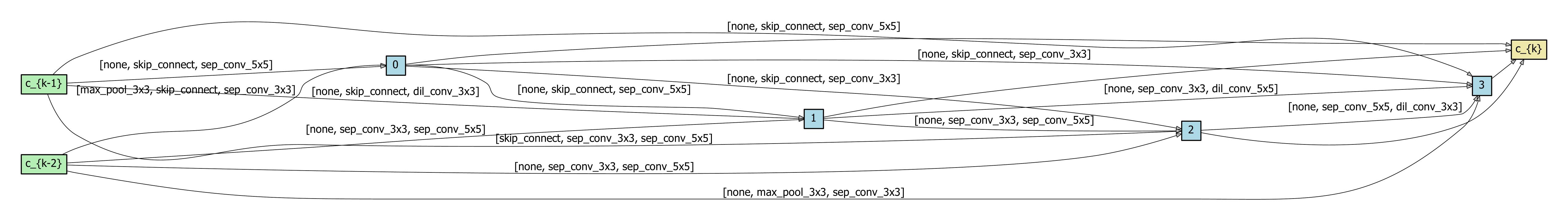}
        \caption{Meta-transfer architecture for normal cell. On each edge a combination of three candidate operations can be observed.}
   \label{fig:meta-ta} 
    \end{subfigure}
    \begin{subfigure}[b]{0.49\textwidth}
        \includegraphics[width=1\linewidth]{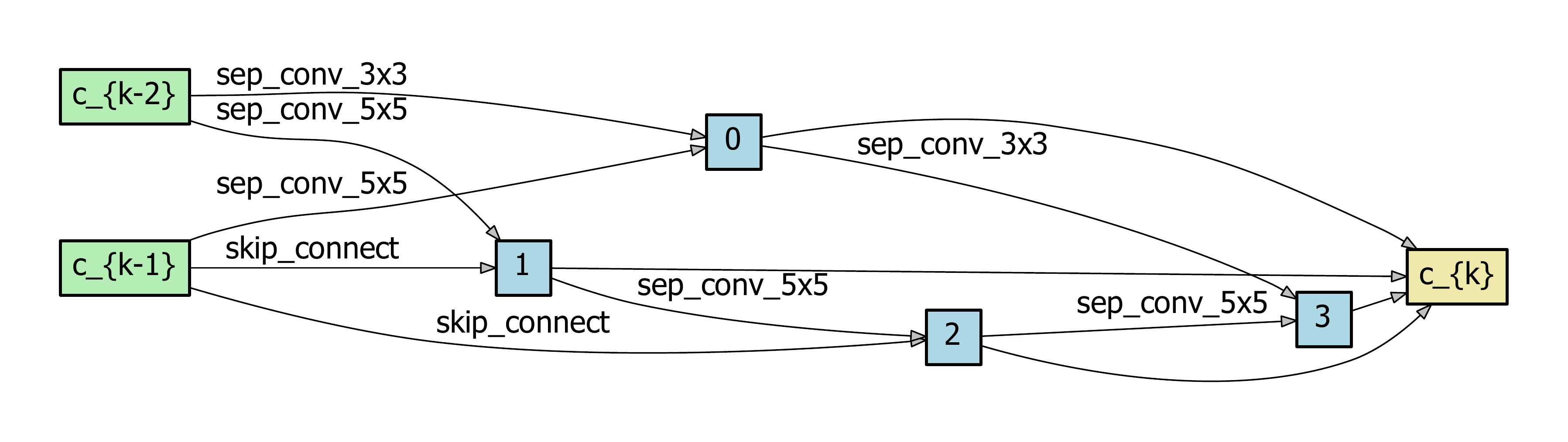}
        \caption{Normal cell found on \textit{aircraft} task by WS-DARTS($\lambda_\text{meta}$)}
   \label{fig:a2} 
    \end{subfigure}
    \begin{subfigure}[b]{0.49\textwidth}
        \includegraphics[width=1\linewidth]{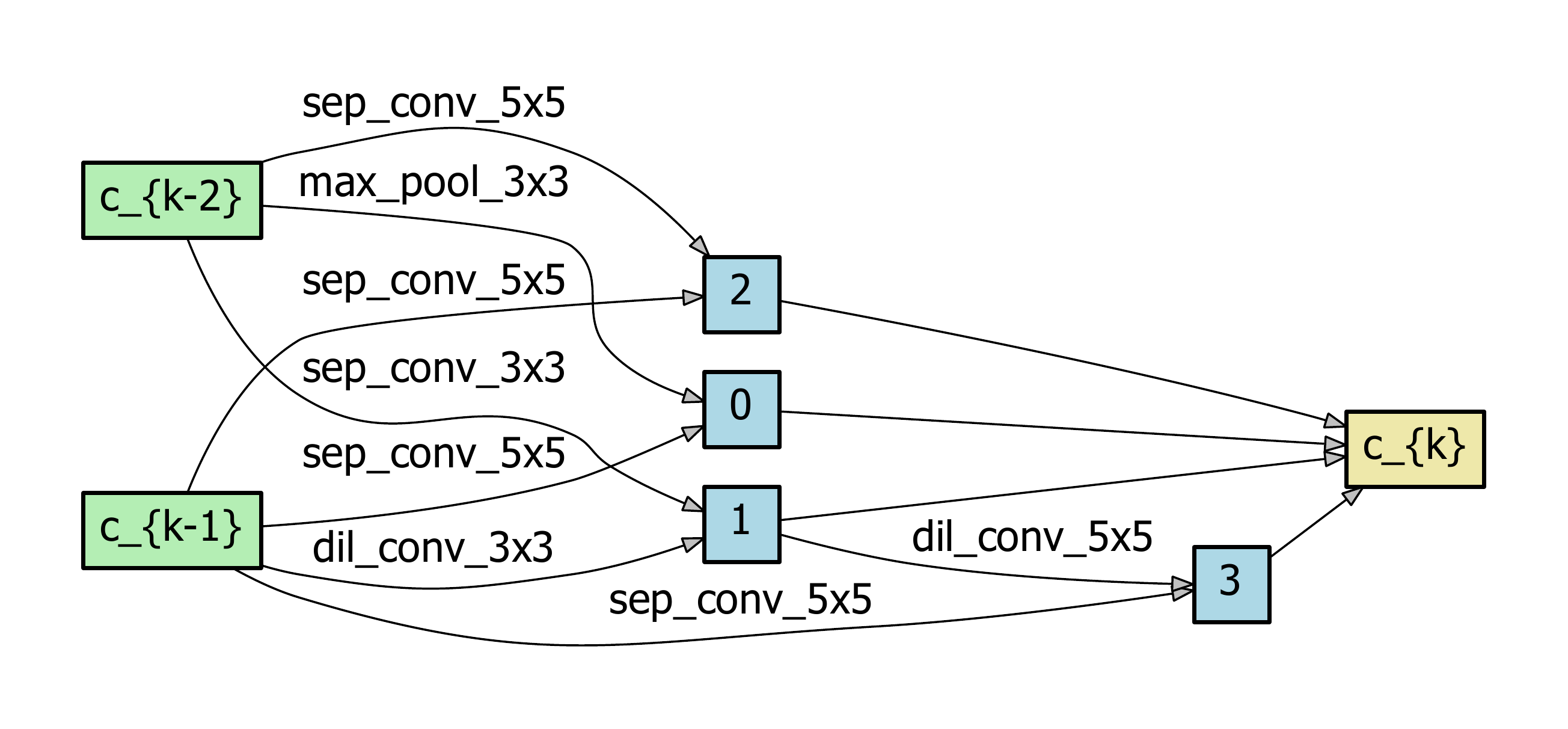}
        \caption{Normal cell found on \textit{birds} task by WS-DARTS($\lambda_\text{meta}$)}
   \label{fig:b2}
    \end{subfigure}
    \caption{Visualization of (a) meta-transfer architecture for normal cell, $\lambda_\text{meta}$, and normal cells discovered by warm-starting DARTS using $\lambda_\text{meta}$ on (b) \textit{aircraft} task and  (c) \textit{birds} task.}
    \label{fig:e2}
\end{figure}

In experiment 2, we were searching for the meta-transfer architecture over the meta-train dataset --- a collection of three datasets described in the section \ref{sec:e2}. It took $50.2$ hours to find the meta-transfer architecture shown in Figure 10a. Interestingly, four candidate operations (\textit{none}, \textit{skip\_connect}, sep\_conv\_3x3 and sep\_conv\_5x5) represent $88\%$ of the search space by our meta-transfer architecture. If we disregard the \textit{none} operator, as it is a special operator that indicates the lack of connection, and \textit{skip-connect}, we are left with only 3x3 and 5x5 separable convolution. This presents a challenge for the \textit{birds} task as it sub-optimally restricts the search space and prevents DARTS from finding well performant architectures as will be discussed later. Comparing the architectures for the \textit{birds} task found in Experiment 1 (Figure \ref{fig:normal_cell_b}) to the one found in Experiment 2 (Figure \ref{fig:b2}) shows that more diverse operations are needed in the transfer architecture to find better performing architectures. Otherwise, the transfer architecture is too restrictive and negatively affect the performance

Looking at our first evaluation criteria, we found similar results to Experiment \ref{e1}. The time needed to find an architecture by warm-starting DARTS using our meta-transfer architecture, $\lambda_\text{meta}$, was drastically reduced as is presented in Table \ref{table:time-meta}. Warm-started DARTS using $\lambda_\text{meta}$ shows the fastest performance compared to transfer architecture, $\lambda_i$, and learned transfer architecture, $\hat{\lambda}_i$, from the Experiment \ref{e1}. The difference in the mean is statistically significant at $p < 0.05$, where the \textit{p}-value is $0.016938$ and $0.002933$ for the aircraft and birds task, respectively. 

\begin{table}[h]
\caption{Search times for warm started DARTS using meta-transfer architecture, $\lambda_\text{meta}$, performed on selected tasks. P-DARTS column represents search times of the P-DARTS algorithm for comparisons.}

\centering
    \input{table/time2}
\label{table:time-meta}
\end{table}
Table \ref{table:performance2} shows accuracy results of the task-specific architectures obtained from warm-starting DARTS using meta-transfer architecture and compares it to the results obtained by the authors of the meta-dataset \citep{triantafillou2019metadataset}. We got mixed results with the meta-transfer architecture, which could be attributed to the selection of the meta-dataset. More precisely, if a the the meta-dataset consists of tasks that are not similar enough to our target task, we will sub-optimally restrict the search space. However, the further tests would be needed to give a definitive answer. While we quickly found good architecture for \textit{aircraft} task, we were unable to find one for \textit{birds} task. This is consistent with the findings in Experiment \ref{e1}. However, accuracy for the \textit{birds} task deteriorated even further. Low accuracy results for the \textit{birds} task are also confirmed by comparing them to the results obtained by the authors of meta-dataset \citep{triantafillou2019metadataset}. 

\begin{table}[h]
\caption{Performance results showing validation accuracy of evaluated architectures. Meta-dataset \citep{triantafillou2019metadataset} results are the best results obtained by the authors and are used for the external performance comparison between different approaches.}
\centering
    \input{table/acc2}
\label{table:performance2}
\end{table}

%% file: table/time.tex
\resizebox{\textwidth}{!}{%
\begin{tabular}{@{}lccccc@{}}
\toprule
\multicolumn{1}{c}{}                                & \textbf{P-DARTS}                                  & \multicolumn{2}{c}{\textbf{WS-DARTS($\lambda_i$)}} & \multicolumn{2}{c}{\textbf{WS-DARTS($\hat{\lambda}_i$)}} \\ 
\multicolumn{1}{c}{\multirow{-2}{*}{\textbf{Task}}} & Time (hours)                                      & Time (hours)   & Improvement                    & Time (hours)            & Improvement           \\ \cmidrule(r){1-6}
\multicolumn{1}{l|}{aircraft}                       & \multicolumn{1}{c|}{\cellcolor[HTML]{EFEFEF}4.7}  & 1.6            & \multicolumn{1}{c|}{66.3 \%}   & 2.2                     & 52.9 \%               \\
\multicolumn{1}{l|}{flower}                         & \multicolumn{1}{c|}{\cellcolor[HTML]{EFEFEF}4.0}  & 1.4            & \multicolumn{1}{c|}{65.8 \%}   & 1.6                     & 59.2 \%               \\
\multicolumn{1}{l|}{birds}                          & \multicolumn{1}{c|}{\cellcolor[HTML]{EFEFEF}5.8}  & 1.9            & \multicolumn{1}{c|}{67.2 \%}   & 2.3                     & 61.0 \%               \\
\multicolumn{1}{l|}{dtd}                            & \multicolumn{1}{c|}{\cellcolor[HTML]{EFEFEF}2.7}  & 0.9            & \multicolumn{1}{c|}{66.4 \%}   & 1.1                     & 60.8 \%               \\
\multicolumn{1}{l|}{tiny\_imagenet}                 & \multicolumn{1}{c|}{\cellcolor[HTML]{EFEFEF}42.6} & 17.7           & \multicolumn{1}{c|}{58.5 \%}   & 17.7                    & 58.4 \%               \\ \bottomrule
\end{tabular}%
}

%% file: table/acc.tex
\resizebox{\textwidth}{!}{%
\begin{tabular}{@{}ccccc@{}}
\toprule
\textbf{Task} & Meta-dataset & \textbf{Baseline} & \textbf{WS-DARTS($\lambda_i$)} & \multicolumn{1}{l}{\textbf{WS-DARTS($\hat{\lambda}_i$)}} \\ \midrule
aircraft       & 76.41   & $78.937891$ & \cellcolor[HTML]{C6E0B4}$79.447942$ & \cellcolor[HTML]{A9D08E}\textbf{79.747973} \\
flower         & 88.72 & $90.476188$ & \cellcolor[HTML]{C6E0B4}$90.879119$ & \cellcolor[HTML]{A9D08E}\textbf{91.501829} \\
birds          & \textbf{69.88}   & $61.246817$ & \cellcolor[HTML]{FCE4D6}$60.63613$  & \cellcolor[HTML]{A9D08E}$61.857504$ \\
dtd            & \textbf{68.25} & $49,25531$ & \cellcolor[HTML]{FCE4D6}$48.297871$ & \cellcolor[HTML]{A9D08E}$49,574466$ \\
tiny\_imagenet & 49.53      & $63.34073$  & \cellcolor[HTML]{C6E0B4}\textbf{63.943718} & \cellcolor[HTML]{C6E0B4}63.764182 \\ \bottomrule
\end{tabular}%
}

%% file: table/time2.tex
\begin{tabular}{@{}lccccc@{}}
\toprule
\multicolumn{1}{c}{}                                & \textbf{P-DARTS}                                  & \multicolumn{2}{c}{\textbf{WS-DARTS($\lambda_\text{meta}$)}} \\ 
\multicolumn{1}{c}{\multirow{-2}{*}{\textbf{Task}}} & Time (hours)                                      & Time (hours)   & Improvement \\ \cmidrule(r){1-4}
\multicolumn{1}{l|}{aircraft}                       & \multicolumn{1}{c|}{\cellcolor[HTML]{EFEFEF}4.7}  & 1.5            & \multicolumn{1}{c}{67.5 \%} \\
\multicolumn{1}{l|}{birds}                          & \multicolumn{1}{c|}{\cellcolor[HTML]{EFEFEF}5.8}  & 1.8            & \multicolumn{1}{c}{69 \%} \\ \bottomrule
\end{tabular}

%% file: table/acc2.tex
\begin{tabular}{@{}ccccc@{}}
\toprule
\textbf{Task} & Meta-dataset & \textbf{Baseline} & \textbf{WS-DARTS($\lambda_\text{meta}$)} \\ \midrule
aircraft       & 76.14 & 78.257461 & \cellcolor[HTML]{C6E0B4}\textbf{79,71797} \\
birds         & \textbf{69.88} & 59.149377 & \cellcolor[HTML]{FCE4D6}$57,83714$ \\ \bottomrule

\end{tabular}

%% file: chapters/05_conclusions.tex
This work focuses on speeding up the neural architecture search over multiple tasks. Naïve solutions, like transferring task-specific architectures or searching from scratch, are either not optimal or inefficient and thus not suitable for multi-task environments \citep{Lian2020TowardsFA}. We presented a meta-learning framework to warm-start DARTS on novel tasks. This was achieved by introducing a transfer architecture that can be quickly adapted to some new task. On average, it yields $2.5\times$ faster architecture search times compared to the searching from scratch. Not only does this reduce the needed computational resources, but it can also lead to better-performing architectures in comparison to the simple cell transfer from a proxy task. 

We first introduced a transfer architecture search, that is able to find a transfer architecture for a single task or a meta-transfer architecture on multiple tasks. We restricted the size of transfer architecture to only three candidate operations on every edge. While this approach greatly reduces the search time, it also severely limits the number of all possible architectures that can be found. We addressed this limitation by adding a task similarity measure to find transfer architecture found on the most similar task. The idea behind this is that NAS will find similar architectures on similar tasks and by significantly reducing the search space we do not negatively affect performance. In this work, we were not fully successful to mitigate the risk of negative performance due to the restricted search space. Similarly, a meta-transfer architecture was obtained by searching on a meta-dataset --- a collection of three datasets described in Section \ref{sec:e2}. 

Finally, we demonstrated a method to warm-start DARTS for novel tasks. We conducted experiments by warm-starting DARTS with transfer and meta-transfer architectures. We obtained better results compared to the simple cell transfer in the majority of cases, while greatly reducing search time. Our approach can be used to warm-start other NAS methods based on DARTS.

We believe that we only scratched the surface in this work and future research using more datasets, leave-one-out evaluation, and hyperparameter tuning (\eg{number of epochs}) is needed. Additionally, an investigation into the possibilities regarding learned transfer architectures would give a better understanding of this approach. Especially, the effect on tasks containing a small number of available images (\eg{few-shot}) and the possibility of lowering the number of epochs of warm-started DARTS when initialized with a learned transfer architecture.

Observing nature, we can see that living things learn from past experiences. We believe that this should also be the case in neural architecture search to reduce computational overhead and optimize the process. We hope that this work may provide additional insight or inspiration on meta-learning NAS.

%% file: main.bbl
\begin{thebibliography}{54}
\providecommand{\natexlab}[1]{#1}
\providecommand{\url}[1]{\texttt{#1}}
\expandafter\ifx\csname urlstyle\endcsname\relax
  \providecommand{\doi}[1]{doi: #1}\else
  \providecommand{\doi}{doi: \begingroup \urlstyle{rm}\Url}\fi

\bibitem[Achille et~al.(2019)Achille, Lam, Tewari, Ravichandran, Maji, Fowlkes,
  Soatto, and Perona]{achille2019task2vec}
Alessandro Achille, Michael Lam, Rahul Tewari, Avinash Ravichandran, Subhransu
  Maji, Charless~C. Fowlkes, Stefano Soatto, and Pietro Perona.
\newblock {Task2Vec}: {Task} embedding for meta-learning.
\newblock \emph{CoRR}, abs/1902.03545, 2019.
\newblock URL \url{http://arxiv.org/abs/1902.03545}.

\bibitem[Alvarez-Melis and Fusi(2020)]{alvarezmelis2020geometric}
David Alvarez-Melis and Nicolo Fusi.
\newblock Geometric dataset distances via optimal transport.
\newblock In H.~Larochelle, M.~Ranzato, R.~Hadsell, M.~F. Balcan, and H.~Lin,
  editors, \emph{Advances in Neural Information Processing Systems}, volume~33,
  pages 21428--21439. Curran Associates, Inc., 2020.
\newblock URL
  \url{https://proceedings.neurips.cc/paper/2020/file/f52a7b2610fb4d3f74b4106fb80b233d-Paper.pdf}.

\bibitem[Amari and Nagaoka(2000)]{book}
Shun-ichi Amari and Hiroshi Nagaoka.
\newblock \emph{Methods of Information Geometry}, volume 191.
\newblock 01 2000.

\bibitem[Baker et~al.(2016)Baker, Gupta, Naik, and Raskar]{baker2017designing}
Bowen Baker, Otkrist Gupta, Nikhil Naik, and Ramesh Raskar.
\newblock Designing neural network architectures using reinforcement learning.
\newblock \emph{CoRR}, abs/1611.02167, 2016.
\newblock URL \url{http://arxiv.org/abs/1611.02167}.

\bibitem[Bender et~al.(2018)Bender, Kindermans, Zoph, Vasudevan, and
  Le]{pmlr-v80-bender18a}
Gabriel Bender, Pieter-Jan Kindermans, Barret Zoph, Vijay Vasudevan, and Quoc
  Le.
\newblock Understanding and simplifying one-shot architecture search.
\newblock In Jennifer Dy and Andreas Krause, editors, \emph{Proceedings of the
  35th International Conference on Machine Learning}, volume~80 of
  \emph{Proceedings of Machine Learning Research}, pages 550--559,
  Stockholmsmässan, Stockholm Sweden, 10--15 Jul 2018. PMLR.
\newblock URL \url{http://proceedings.mlr.press/v80/bender18a.html}.

\bibitem[Bergstra and Bengio(2012)]{Bergstra2012random}
James Bergstra and Yoshua Bengio.
\newblock Random search for hyper-parameter optimization.
\newblock \emph{J. Mach. Learn. Res.}, 13:\penalty0 281--305, 2012.
\newblock URL
  \url{http://dblp.uni-trier.de/db/journals/jmlr/jmlr13.html#BergstraB12}.

\bibitem[Bergstra et~al.(2013)Bergstra, Yamins, and Cox]{pmlr-v28-bergstra13}
James Bergstra, Daniel Yamins, and David Cox.
\newblock Making a science of model search: Hyperparameter optimization in
  hundreds of dimensions for vision architectures.
\newblock \emph{Proceedings of Machine Learning Research}, 28\penalty0
  (1):\penalty0 115--123, 17--19 Jun 2013.
\newblock URL \url{http://proceedings.mlr.press/v28/bergstra13.html}.

\bibitem[Brazdil et~al.(2003)Brazdil, Soares, and da~Costa]{Brazdil2003}
Pavel~B. Brazdil, Carlos Soares, and Joaquim~Pinto da~Costa.
\newblock Ranking learning algorithms: Using {IBL} and meta-learning on
  accuracy and time results.
\newblock \emph{Machine Learning}, 50\penalty0 (3):\penalty0 251--277, Mar
  2003.
\newblock ISSN 1573-0565.
\newblock \doi{10.1023/A:1021713901879}.
\newblock URL \url{https://doi.org/10.1023/A:1021713901879}.

\bibitem[Brock et~al.(2017)Brock, Lim, Ritchie, and Weston]{brock2017smash}
Andrew Brock, Theodore Lim, James~M. Ritchie, and Nick Weston.
\newblock {SMASH:} one-shot model architecture search through hypernetworks.
\newblock \emph{CoRR}, abs/1708.05344, 2017.
\newblock URL \url{http://arxiv.org/abs/1708.05344}.

\bibitem[Cai et~al.(2018)Cai, Zhu, and Han]{cai2019proxylessnas}
Han Cai, Ligeng Zhu, and Song Han.
\newblock Proxylessnas: Direct neural architecture search on target task and
  hardware.
\newblock \emph{CoRR}, abs/1812.00332, 2018.
\newblock URL \url{http://arxiv.org/abs/1812.00332}.

\bibitem[Chen and Hsieh(2020)]{pmlr-v119-chen20f}
Xiangning Chen and Cho-Jui Hsieh.
\newblock Stabilizing differentiable architecture search via perturbation-based
  regularization.
\newblock In Hal~Daumé III and Aarti Singh, editors, \emph{Proceedings of the
  37th International Conference on Machine Learning}, volume 119 of
  \emph{Proceedings of Machine Learning Research}, pages 1554--1565. PMLR,
  13--18 Jul 2020.
\newblock URL \url{http://proceedings.mlr.press/v119/chen20f.html}.

\bibitem[Chen et~al.(2019)Chen, Xie, Wu, and Tian]{chen2019pdarts}
Xin Chen, Lingxi Xie, Jun Wu, and Qi~Tian.
\newblock Progressive differentiable architecture search: Bridging the depth
  gap between search and evaluation.
\newblock \emph{CoRR}, abs/1904.12760, 2019.
\newblock URL \url{http://arxiv.org/abs/1904.12760}.

\bibitem[Chu et~al.(2019)Chu, Zhou, Zhang, and Li]{Xiangxiang2019FairDARTS}
Xiangxiang Chu, Tianbao Zhou, Bo~Zhang, and Jixiang Li.
\newblock Fair {DARTS:} eliminating unfair advantages in differentiable
  architecture search.
\newblock \emph{CoRR}, abs/1911.12126, 2019.
\newblock URL \url{http://arxiv.org/abs/1911.12126}.

\bibitem[Cimpoi et~al.(2014)Cimpoi, Maji, Kokkinos, Mohamed, and
  Vedaldi]{cimpoi14describing}
M.~Cimpoi, S.~Maji, I.~Kokkinos, S.~Mohamed, and A.~Vedaldi.
\newblock Describing textures in the wild.
\newblock In \emph{Proceedings of the {IEEE} Conf. on Computer Vision and
  Pattern Recognition ({CVPR})}, 2014.

\bibitem[{Deng} et~al.(2009){Deng}, {Dong}, {Socher}, {Li}, {Kai Li}, and {Li
  Fei-Fei}]{imagenet_cvpr09}
J.~{Deng}, W.~{Dong}, R.~{Socher}, L.~{Li}, {Kai Li}, and {Li Fei-Fei}.
\newblock {ImageNet}: {A} large-scale hierarchical image database.
\newblock In \emph{2009 IEEE Conference on Computer Vision and Pattern
  Recognition}, pages 248--255, 2009.
\newblock \doi{10.1109/CVPR.2009.5206848}.

\bibitem[{Elsken} et~al.(2018){Elsken}, {Hendrik Metzen}, and
  {Hutter}]{elsken2019efficient}
Thomas {Elsken}, Jan {Hendrik Metzen}, and Frank {Hutter}.
\newblock {Efficient Multi-objective Neural Architecture Search via Lamarckian
  Evolution}.
\newblock \emph{arXiv e-prints}, art. arXiv:1804.09081, April 2018.

\bibitem[Elsken et~al.(2019{\natexlab{a}})Elsken, Metzen, and
  Hutter]{elsken2019survey}
Thomas Elsken, Jan~Hendrik Metzen, and Frank Hutter.
\newblock Neural architecture search: A survey.
\newblock \emph{Journal of Machine Learning Research}, 20\penalty0
  (55):\penalty0 1--21, 2019{\natexlab{a}}.
\newblock URL \url{http://jmlr.org/papers/v20/18-598.html}.

\bibitem[Elsken et~al.(2019{\natexlab{b}})Elsken, Staffler, Metzen, and
  Hutter]{elsken2020metalearning}
Thomas Elsken, Benedikt Staffler, Jan~Hendrik Metzen, and Frank Hutter.
\newblock Meta-learning of neural architectures for few-shot learning.
\newblock \emph{CoRR}, abs/1911.11090, 2019{\natexlab{b}}.
\newblock URL \url{http://arxiv.org/abs/1911.11090}.

\bibitem[Fang et~al.(2019)Fang, Chen, Zhang, Zhang, Huang, Meng, Liu, and
  Wang]{fang2019eatnas}
Jiemin Fang, Yukang Chen, Xinbang Zhang, Qian Zhang, Chang Huang, Gaofeng Meng,
  Wenyu Liu, and Xinggang Wang.
\newblock {EAT-NAS:} {Elastic} architecture transfer for accelerating
  large-scale neural architecture search.
\newblock \emph{CoRR}, abs/1901.05884, 2019.
\newblock URL \url{http://arxiv.org/abs/1901.05884}.

\bibitem[Finn et~al.(2017)Finn, Abbeel, and Levine]{maml}
Chelsea Finn, Pieter Abbeel, and Sergey Levine.
\newblock Model-agnostic meta-learning for fast adaptation of deep networks.
\newblock \emph{CoRR}, abs/1703.03400, 2017.
\newblock URL \url{http://arxiv.org/abs/1703.03400}.

\bibitem[Golovin et~al.(2017)Golovin, Solnik, Moitra, Kochanski, Karro, and
  Sculley]{Golovin2017vizier}
Daniel Golovin, Benjamin Solnik, Subhodeep Moitra, Greg Kochanski, John~Elliot
  Karro, and D.~Sculley, editors.
\newblock \emph{Google Vizier: A Service for Black-Box Optimization}, 2017.
\newblock URL
  \url{http://www.kdd.org/kdd2017/papers/view/google-vizier-a-service-for-black-box-optimization}.

\bibitem[Houben(2019)]{houben2019msc}
Maurice Houben.
\newblock Towards warm-starting darts.
\newblock Master's thesis, {Eindhoven University of Technology}, 2019.

\bibitem[Hundt et~al.(2019)Hundt, Jain, and Hager]{Hundt2019sharpDARTS}
Andrew Hundt, Varun Jain, and Gregory~D. Hager.
\newblock sharpdarts: Faster and more accurate differentiable architecture
  search.
\newblock \emph{CoRR}, abs/1903.09900, 2019.
\newblock URL \url{http://arxiv.org/abs/1903.09900}.

\bibitem[Kim et~al.(2016)Kim, Alletto, and Rigazio]{kim2017similarity}
Minyoung Kim, Stefano Alletto, and Luca Rigazio.
\newblock Similarity mapping with enhanced siamese network for multi-object
  tracking.
\newblock \emph{CoRR}, abs/1609.09156, 2016.
\newblock URL \url{http://arxiv.org/abs/1609.09156}.

\bibitem[Krizhevsky(2009)]{Krizhevsky09learningmultiple}
Alex Krizhevsky.
\newblock Learning multiple layers of features from tiny images.
\newblock Technical report, University of Toronto, 2009.
\newblock URL
  \url{https://www.cs.toronto.edu/~kriz/learning-features-2009-TR.pdf}.

\bibitem[Leite and Brazdil(2007)]{Leite2007}
Rui Leite and Pavel Brazdil.
\newblock An iterative process for building learning curves and predicting
  relative performance of classifiers.
\newblock In \emph{Proceedings of the Aritficial Intelligence 13th Portuguese
  Conference on Progress in Artificial Intelligence}, EPIA'07, page 87–98,
  Berlin, Heidelberg, 2007. Springer-Verlag.
\newblock ISBN 3540770003.

\bibitem[Lemke et~al.(2013)Lemke, Budka, and Gabrys]{Lemke2013}
Christiane Lemke, Marcin Budka, and Bogdan Gabrys.
\newblock Metalearning: a survey of trends and technologies.
\newblock \emph{Artificial Intelligence Review}, DOI:
  10.1007/s10462-013-9406-y, 06 2013.
\newblock \doi{10.1007/s10462-013-9406-y}.

\bibitem[Lian et~al.(2020)Lian, Zheng, Xu, Lu, Lin, Zhao, Huang, and
  Gao]{Lian2020TowardsFA}
Dongze Lian, Yin Zheng, Yintao Xu, Yanxiong Lu, Leyu Lin, Peilin Zhao, Junzhou
  Huang, and Shenghua Gao.
\newblock Towards fast adaptation of neural architectures with meta learning.
\newblock In \emph{8th International Conference on Learning Representations,
  {ICLR} 2020, Addis Ababa, Ethiopia, April 26-30, 2020}. OpenReview.net, 2020.
\newblock URL \url{https://openreview.net/forum?id=r1eowANFvr}.

\bibitem[Liang et~al.(2019)Liang, Zhang, Sun, He, Huang, Zhuang, and
  Li]{Hanwen2019DARTS+}
Hanwen Liang, Shifeng Zhang, Jiacheng Sun, Xingqiu He, Weiran Huang, Kechen
  Zhuang, and Zhenguo Li.
\newblock {DARTS+:} improved differentiable architecture search with early
  stopping.
\newblock \emph{CoRR}, abs/1909.06035, 2019.
\newblock URL \url{http://arxiv.org/abs/1909.06035}.

\bibitem[Liu et~al.(2017)Liu, Simonyan, Vinyals, Fernando, and
  Kavukcuoglu]{liu2017hierarchical}
Hanxiao Liu, Karen Simonyan, Oriol Vinyals, Chrisantha Fernando, and Koray
  Kavukcuoglu.
\newblock Hierarchical representations for efficient architecture search.
\newblock \emph{CoRR}, abs/1711.00436, 2017.
\newblock URL \url{http://arxiv.org/abs/1711.00436}.

\bibitem[Liu et~al.(2018)Liu, Simonyan, and Yang]{liu2019darts}
Hanxiao Liu, Karen Simonyan, and Yiming Yang.
\newblock {DARTS:} differentiable architecture search.
\newblock \emph{CoRR}, abs/1806.09055, 2018.
\newblock URL \url{http://arxiv.org/abs/1806.09055}.

\bibitem[Maji et~al.(2013)Maji, Rahtu, Kannala, Blaschko, and
  Vedaldi]{maji13fine-grained}
Subhransu Maji, Esa Rahtu, Juho Kannala, Matthew~B. Blaschko, and Andrea
  Vedaldi.
\newblock Fine-grained visual classification of aircraft.
\newblock \emph{CoRR}, abs/1306.5151, 2013.
\newblock URL \url{http://arxiv.org/abs/1306.5151}.

\bibitem[Miller et~al.(1989)Miller, Todd, and Hegde]{miller1989}
Geoffrey~F. Miller, Peter~M. Todd, and Shailesh~U. Hegde.
\newblock Designing neural networks using genetic algorithms.
\newblock In \emph{Proceedings of the Third International Conference on Genetic
  Algorithms}, page 379–384, San Francisco, CA, USA, 1989. Morgan Kaufmann
  Publishers Inc.
\newblock ISBN 1558600063.

\bibitem[Nichol et~al.(2018)Nichol, Achiam, and Schulman]{nichol2018firstorder}
Alex Nichol, Joshua Achiam, and John Schulman.
\newblock On first-order meta-learning algorithms.
\newblock \emph{CoRR}, abs/1803.02999, 2018.
\newblock URL \url{http://arxiv.org/abs/1803.02999}.

\bibitem[Nilsback and Zisserman(2008)]{Nilsback08}
Maria-Elena Nilsback and Andrew Zisserman.
\newblock Automated flower classification over a large number of classes.
\newblock In \emph{Indian Conference on Computer Vision, Graphics and Image
  Processing}, Dec 2008.

\bibitem[Real et~al.(2017)Real, Moore, Selle, Saxena, Suematsu, Le, and
  Kurakin]{real2017largescale}
Esteban Real, Sherry Moore, Andrew Selle, Saurabh Saxena, Yutaka~Leon Suematsu,
  Quoc~V. Le, and Alex Kurakin.
\newblock Large-scale evolution of image classifiers.
\newblock \emph{CoRR}, abs/1703.01041, 2017.
\newblock URL \url{http://arxiv.org/abs/1703.01041}.

\bibitem[Real et~al.(2018)Real, Aggarwal, Huang, and Le]{real2019regularized}
Esteban Real, Alok Aggarwal, Yanping Huang, and Quoc~V. Le.
\newblock Regularized evolution for image classifier architecture search.
\newblock \emph{CoRR}, abs/1802.01548, 2018.
\newblock URL \url{http://arxiv.org/abs/1802.01548}.

\bibitem[Ren et~al.(2021)Ren, Xiao, Chang, Huang, Li, Chen, and
  Wang]{ren2021comprehensive}
Pengzhen Ren, Yun Xiao, Xiaojun Chang, Po-Yao Huang, Zhihui Li, Xiaojiang Chen,
  and Xin Wang.
\newblock A comprehensive survey of neural architecture search: Challenges and
  solutions, 2021.

\bibitem[Rosenstein et~al.(2005)Rosenstein, Marx, Kaelbling, and
  Dietterich]{Rosenstein2005}
Michael~T. Rosenstein, Zvika Marx, Leslie~Pack Kaelbling, and Thomas~G.
  Dietterich.
\newblock To transfer or not to transfer.
\newblock In \emph{In NIPS’05 Workshop, Inductive Transfer: 10 Years Later},
  2005.

\bibitem[Runge et~al.(2018)Runge, Stoll, Falkner, and
  Hutter]{runge2019learning}
Frederic Runge, Danny Stoll, Stefan Falkner, and Frank Hutter.
\newblock Learning to design {RNA}.
\newblock \emph{CoRR}, abs/1812.11951, 2018.
\newblock URL \url{http://arxiv.org/abs/1812.11951}.

\bibitem[Schroff et~al.(2015)Schroff, Kalenichenko, and
  Philbin]{schroff2015facenet}
Florian Schroff, Dmitry Kalenichenko, and James Philbin.
\newblock Facenet: A unified embedding for face recognition and clustering.
\newblock \emph{2015 IEEE Conference on Computer Vision and Pattern Recognition
  (CVPR)}, Jun 2015.
\newblock \doi{10.1109/cvpr.2015.7298682}.
\newblock URL \url{http://dx.doi.org/10.1109/CVPR.2015.7298682}.

\bibitem[Simonyan and Zisserman(2015)]{simonyan2014deep}
Karen Simonyan and Andrew Zisserman.
\newblock Very deep convolutional networks for large-scale image recognition.
\newblock In Yoshua Bengio and Yann LeCun, editors, \emph{3rd International
  Conference on Learning Representations, {ICLR} 2015, San Diego, CA, USA, May
  7-9, 2015, Conference Track Proceedings}, 2015.
\newblock URL \url{http://arxiv.org/abs/1409.1556}.

\bibitem[Srivastava et~al.(2014)Srivastava, Hinton, Krizhevsky, Sutskever, and
  Salakhutdinov]{Srivastava2014dropout}
Nitish Srivastava, Geoffrey Hinton, Alex Krizhevsky, Ilya Sutskever, and Ruslan
  Salakhutdinov.
\newblock Dropout: A simple way to prevent neural networks from overfitting.
\newblock \emph{J. Mach. Learn. Res.}, 15\penalty0 (1):\penalty0 1929–1958,
  January 2014.
\newblock ISSN 1532-4435.

\bibitem[Szegedy et~al.(2014)Szegedy, Liu, Jia, Sermanet, Reed, Anguelov,
  Erhan, Vanhoucke, and Rabinovich]{szegedy2014going}
Christian Szegedy, Wei Liu, Yangqing Jia, Pierre Sermanet, Scott~E. Reed,
  Dragomir Anguelov, Dumitru Erhan, Vincent Vanhoucke, and Andrew Rabinovich.
\newblock Going deeper with convolutions.
\newblock \emph{CoRR}, abs/1409.4842, 2014.
\newblock URL \url{http://arxiv.org/abs/1409.4842}.

\bibitem[Triantafillou et~al.(2019)Triantafillou, Zhu, Dumoulin, Lamblin, Xu,
  Goroshin, Gelada, Swersky, Manzagol, and
  Larochelle]{triantafillou2019metadataset}
Eleni Triantafillou, Tyler Zhu, Vincent Dumoulin, Pascal Lamblin, Kelvin Xu,
  Ross Goroshin, Carles Gelada, Kevin Swersky, Pierre{-}Antoine Manzagol, and
  Hugo Larochelle.
\newblock Meta-dataset: {A} dataset of datasets for learning to learn from few
  examples.
\newblock \emph{CoRR}, abs/1903.03096, 2019.
\newblock URL \url{http://arxiv.org/abs/1903.03096}.

\bibitem[Vanschoren(2018)]{vanschoren2018mlsurvey}
Joaquin Vanschoren.
\newblock Meta-learning: {A} survey.
\newblock \emph{CoRR}, abs/1810.03548, 2018.
\newblock URL \url{http://arxiv.org/abs/1810.03548}.

\bibitem[Villani(2008)]{Villani2008}
Cédric Villani.
\newblock \emph{Optimal transport - Old and new}, volume 338, pages xxii+973.
\newblock Springer-Verlag Berlin Heidelberg, 01 2008.
\newblock \doi{10.1007/978-3-540-71050-9}.
\newblock URL
  \url{https://cedricvillani.org/sites/dev/files/old_images/2012/08/preprint-1.pdf}.

\bibitem[Wah et~al.(2011)Wah, Branson, Welinder, Perona, and
  Belongie]{WahCUB_200_2011}
C.~Wah, S.~Branson, P.~Welinder, P.~Perona, and S.~Belongie.
\newblock {The Caltech-UCSD Birds-200-2011 Dataset}.
\newblock Technical Report CNS-TR-2011-001, California Institute of Technology,
  2011.

\bibitem[Wong et~al.(2018)Wong, Houlsby, Lu, and Gesmundo]{wong2019transfer}
Catherine Wong, Neil Houlsby, Yifeng Lu, and Andrea Gesmundo.
\newblock Transfer automatic machine learning.
\newblock \emph{CoRR}, abs/1803.02780, 2018.
\newblock URL \url{http://arxiv.org/abs/1803.02780}.

\bibitem[Zamir et~al.(2018)Zamir, Sax, Shen, Guibas, Malik, and
  Savarese]{zamir2018taskonomy}
Amir~Roshan Zamir, Alexander Sax, William~B. Shen, Leonidas~J. Guibas, Jitendra
  Malik, and Silvio Savarese.
\newblock Taskonomy: Disentangling task transfer learning.
\newblock \emph{CoRR}, abs/1804.08328, 2018.
\newblock URL \url{http://arxiv.org/abs/1804.08328}.

\bibitem[Zela et~al.(2019)Zela, Elsken, Saikia, Marrakchi, Brox, and
  Hutter]{Zela2019Understanding}
Arber Zela, Thomas Elsken, Tonmoy Saikia, Yassine Marrakchi, Thomas Brox, and
  Frank Hutter.
\newblock Understanding and robustifying differentiable architecture search.
\newblock \emph{CoRR}, abs/1909.09656, 2019.
\newblock URL \url{http://arxiv.org/abs/1909.09656}.

\bibitem[Zhong et~al.(2017)Zhong, Yan, and Liu]{zhong2017practical}
Zhao Zhong, Junjie Yan, and Cheng{-}Lin Liu.
\newblock Practical network blocks design with q-learning.
\newblock \emph{CoRR}, abs/1708.05552, 2017.
\newblock URL \url{http://arxiv.org/abs/1708.05552}.

\bibitem[Zoph and Le(2016)]{zoph2017neural}
Barret Zoph and Quoc~V. Le.
\newblock Neural architecture search with reinforcement learning.
\newblock \emph{CoRR}, abs/1611.01578, 2016.
\newblock URL \url{http://arxiv.org/abs/1611.01578}.

\bibitem[Zoph et~al.(2017)Zoph, Vasudevan, Shlens, and Le]{zoph2017learning}
Barret Zoph, Vijay Vasudevan, Jonathon Shlens, and Quoc~V. Le.
\newblock Learning transferable architectures for scalable image recognition.
\newblock \emph{CoRR}, abs/1707.07012, 2017.
\newblock URL \url{http://arxiv.org/abs/1707.07012}.

\end{thebibliography}
